%% file: main.tex
\input{analysis/final_review_v54/caption_copy/CAPTION_COPY_FINAL_PATCH.tex}
\documentclass[a4paper]{medreflect}
\usepackage{fix-cm,array,tabularx,colortbl,enumitem,makecell,needspace,longtable,float}
\usepackage[utf8]{inputenc}
\input{glyphtounicode}
\pdfglyphtounicode{ij}{0069 006A}
\pdfglyphtounicode{IJ}{0049 004A}
\input{analysis/glyph_fix_v38/ADDITIONAL_GLYPH_MAP.tex}
\pdfglyphtounicode{summationtext}{2211}
\pdfglyphtounicode{summationdisplay}{2211}
\pdfglyphtounicode{parenlefttpA}{239B}
\pdfglyphtounicode{parenleftexA}{239C}
\pdfglyphtounicode{parenleftbtA}{239D}
\pdfglyphtounicode{parenrighttpA}{239E}
\pdfglyphtounicode{parenrightexA}{239F}
\pdfglyphtounicode{parenrightbtA}{23A0}
\DisableLigatures[-]{encoding=*,family=zlmtt}
\usepackage{accsupp}
\newcommand{\PaperCopyMath}[2]{%
  \BeginAccSupp{method=hex,unicode,ActualText=#1}%
  #2%
  \EndAccSupp{}%
}
\newcommand{\PaperNormHistory}{\PaperCopyMath{006E006F0072006D00282211005F007B00702208210B005F0061007D00200028D835DC51005F0070002F007CD835DC9C005F0070007C00290020D835DC1E005F00700029}{%
  \operatorname{norm}\!\left(\sum_{p\in\mathcal H_a}
  \frac{d_p}{|\mathcal A_p|}\,\mathbf e_p\right)%
}}
\newcolumntype{L}[1]{>{\raggedright\arraybackslash}p{#1}}
\newcolumntype{Y}{>{\raggedright\arraybackslash}X}
\definecolor{primary}{HTML}{285598}
\definecolor{tablehead}{HTML}{1B2A4A}
\definecolor{tablepale}{HTML}{F3F6FA}
\definecolor{knowteal}{HTML}{3566A2}
\definecolor{knowgreen}{HTML}{317A69}
\definecolor{knowlight}{HTML}{EDF4FC}
\definecolor{knowgold}{HTML}{A66F22}
\definecolor{knowpale}{HTML}{EFF3FA}
\newcommand{\knowme}{\textsc{KnowMe}}
\newcommand{\clip}{\operatorname{clip}}

\newcommand{\cival}[3]{\ensuremath{#1\,{\color{black!48}\scriptstyle\pm\,\ifdim#2pt>#3pt\relax#2\else#3\fi}}}
\newcommand{\pmval}[2]{\ensuremath{#1\,{\color{black!48}\scriptstyle\pm\,#2}}}
\newcommand{\ci}[2]{{\color{black!48}\footnotesize #1--#2}}
\newcommand{\nopm}[1]{\ensuremath{#1\phantom{\,{\scriptstyle\pm\,0.00}}}}
\newcolumntype{C}{>{\centering\arraybackslash}X}
\newcommand{\keypoint}[1]{\textcolor{knowteal}{\textbf{#1}}}
\newcommand{\grouphead}[2]{\rowcolor{tablepale}\multicolumn{#1}{l}{\textbf{#2}}\\}
\setlist{nosep,leftmargin=*}
\setcitestyle{numbers,square,comma,sort&compress}
\hypersetup{pdftitle={KnowMe Paper Radar: Lightweight Discovery of Promising Papers in the Age of Autoresearch},pdfauthor={Zhuo Chen, Jian Wang, Jinjie Gu, Lijun Wu},linkcolor=knowteal,citecolor=knowteal,urlcolor=knowteal}
\renewcommand{\titlefont}{\fontsize{19}{23}\selectfont}
\title{\raisebox{-1pt}{\includegraphics[height=15pt]{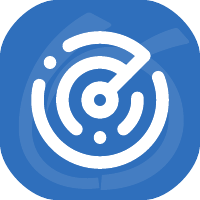}}\hspace{5pt}KnowMe Paper Radar:\\{\fontsize{14.2}{19}\selectfont Lightweight Discovery of Promising Papers in the Age of Autoresearch}}
\renewcommand{\authorlist}{{\fontsize{11}{14}\selectfont\textbf{Zhuo Chen}\quad \textbf{Jian Wang}\quad \textbf{Jinjie Gu}\quad \textbf{Lijun Wu}}}
\renewcommand{\affiliationlist}{{\fontsize{10.5}{13}\selectfont Ant Group\par}\vspace{1pt}{\fontsize{10}{12.5}\selectfont\href{mailto:chenzhuo.zc@antgroup.com,bobblair.wj@antgroup.com,jinjie.gujj@antgroup.com,wlj554215@antgroup.com}{\{chenzhuo.zc, bobblair.wj, jinjie.gujj, wlj554215\}@antgroup.com}\par}}
\renewcommand{\abstractinfobox}{\centering\small\sffamily
\href{https://hackerchenzhuo.github.io/knowme-research-ai/en/}{\textbf{Website \& Research Atlas}}\enspace\textbar\enspace
\href{https://chromewebstore.google.com/detail/aneomkhamcijmfmeeegocbknklflilof}{\textbf{Install Paper Radar}}}
\renewcommand{\mymaketitle}{
 {\centering\titlelist\par\vspace{7pt}\authorlist\par\vspace{3pt}\affiliationlist\par}
 \vspace{9pt}
 \begin{tcolorbox}[colback=knowpale,colframe=knowpale,boxrule=0pt,arc=2mm,left=4mm,right=4mm,top=3mm,bottom=3mm]
 \looseness=-1\relax\textbf{\textcolor{primary}{Abstract}}\quad\abstractlist\par\vspace{7pt}\abstractinfobox
 \end{tcolorbox}\thispagestyle{firststyle}}
\fancypagestyle{mainpages}{\fancyhf{}\fancyhead[L]{\footnotesize\color{textmid}KnowMe Paper Radar}\fancyfoot[C]{\footnotesize\thepage}}
\renewcommand{\maketitle}{\mymaketitle\vspace{3pt}}
\abstract{Large language models (LLMs) are pushing research toward semi-automation: ideas turn into preprints ever faster, and autoresearch pipelines add papers of uneven rigor. When a technique can be overtaken within months, citations arrive too late to tell which new papers deserve reading, so strong work goes unnoticed and newcomers lose their bearings. Yet from the day it appears, a preprint carries the track records of the people and institutions behind it. Established researchers guard their reputations and choose problems and collaborators with \emph{research taste}, so their involvement tends to raise a paper's quality floor. Building on this signal, we present \textbf{KnowMe Paper Radar}, a lightweight arXiv discovery tool that weights each paper's topical relevance by the reputation its authors and institutions stake on it. Author credit also diffuses by message passing along collaborations to surface rising stars, and institutions are profiled through the researchers they hold in each direction rather than by headcount. In a sealed test on 10,386 new papers, those with an identity-bound author were twice as likely to reach the top tenth of early citations. As its library grows into a scholarly knowledge graph that research agents can query through a command-line interface (CLI), KnowMe also offers a promising literature entry point for future autoresearch and recursive self-improvement (RSI).}

\newtcblisting{commandpanel}[2][]{listing only,enhanced,breakable,
  colback=blue!3,colframe=primarylight!85!black,colbacktitle=primarylight!85!black,
  coltitle=white,title={#2},fonttitle=\bfseries,boxrule=.6pt,arc=1mm,
  left=2mm,right=2mm,top=1.2mm,bottom=1.2mm,before skip=5pt,after skip=5pt,
  listing options={basicstyle=\ttfamily\small,breaklines=true,columns=fullflexible,
    keepspaces=true,showstringspaces=false,upquote=true},#1}

\input{analysis/revision_v59/links/FIGURE_ARXIV_LINKS.tex}
\newcommand{\radarpreprint}[1]{\href{https://arxiv.org/abs/#1}{abs/#1}}

\begin{document}
\pagestyle{mainpages}
\raggedbottom
\clubpenalty=10000\widowpenalty=10000
\interlinepenalty=0
\setlength{\parskip}{2pt plus .5pt minus .5pt}
\setlength{\textfloatsep}{9pt plus 2pt minus 2pt}\setlength{\floatsep}{8pt}\setlength{\intextsep}{9pt}
\setlength{\abovedisplayskip}{6pt}\setlength{\belowdisplayskip}{6pt}
\titlespacing*{\section}{0pt}{10pt plus 2pt minus 1pt}{5pt}
\titlespacing*{\subsection}{0pt}{8pt plus 1pt minus 1pt}{4pt}
\titlespacing*{\paragraph}{0pt}{5pt}{.6em}
\emergencystretch=1em\lefthyphenmin=3\righthyphenmin=4
\maketitle
\par\vspace{1pt}
\begin{tcolorbox}[colback=blue!2,colframe=blue!2,boxrule=0pt,arc=1mm,left=3mm,right=3mm,top=1.3mm,bottom=1.3mm,before skip=0pt,after skip=2pt]
{\fontsize{8}{9.4}\selectfont\color{black!65}\emph{``A day in AI is a year in the human world.''}\hfill\mbox{--- Zhilin Yang, Founder of Moonshot AI (Kimi)}}
\end{tcolorbox}
\par\vspace{3pt}
\noindent\begin{minipage}{\linewidth}\centering
\includegraphics[width=6.5in]{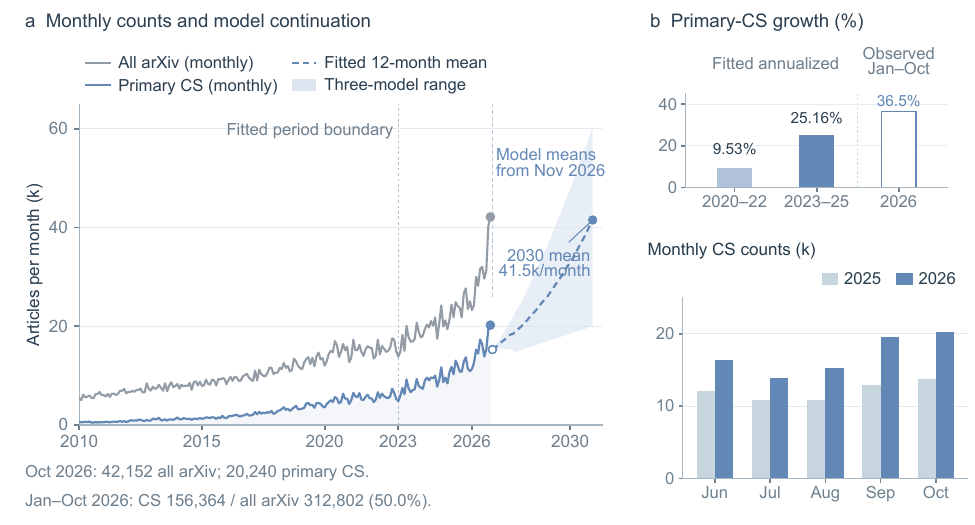}
\captionof{figure}{\textbf{Primary computer-science (CS) submissions to arXiv rose 36.5\% in January--October 2026 over the same months of 2025.} (a) Raw monthly all-arXiv and primary CS submissions through October 2026~\citep{arxivMonthlySubmissions,arxivCSArchiveStats}. From November 2026, the dashed segment shows the saved model's trailing twelve-month mean; shading spans three fitted models. January 2023 marks the fitted period boundary. (b) Fitted annualized rates and a separate observed January--October comparison. Paired bars compare raw June--October CS counts in 2025 and 2026, following the same left-to-right time order. The cumulative CS share uses the matched January--October totals. Assumptions, 95\% intervals and validation appear in App.~\ref{app:growth}.}\label{fig:arxiv-growth-global}
\end{minipage}

\clearpage
\noindent\begin{minipage}{\linewidth}\centering
\includegraphics[width=\linewidth]{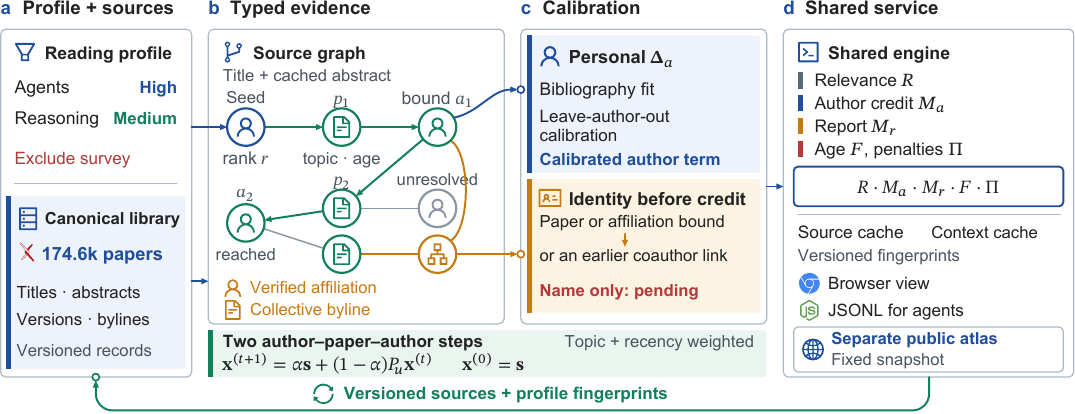}
\captionof{figure}{\textbf{KnowMe weights topical relevance by the credit of identity-bound authors, adds a capped share of collaboration credit and shows institution context beside the score.} (a) The reader sets topic priorities and exclusions over a versioned library. (b) Two restarted propagation steps follow author--paper--author paths weighted by topic and recency, and collaboration credit flows from catalogue authors to their coauthors (Sec.~\ref{sec:rising}). Unresolved byline positions remain in the denominator. Gold links identify verified affiliations and collective bylines. (c) The upper branch calibrates personal author fit against the bibliography. The lower branch binds catalogue names to people: a name earns credit when the paper's own source binding or affiliation evidence confirms it, when the catalogue places the person at the institution of a version-verified collective byline, or when a shared coauthor links it to an earlier confirmed paper; otherwise it stays pending (Sec.~\ref{sec:identity}). (d) The engine multiplies relevance by the strongest bound author's credit, or a capped share of collaboration credit, an institutional-report factor, freshness and penalties (Eq.~\eqref{eq:personal}); institution context is shown beside the score. Browser and JSON Lines (JSONL) clients receive every factor and a reading percentile. The public atlas uses a separate fixed snapshot.}\label{fig:teaser}
\end{minipage}

\section{Introduction}
\label{sec:intro}
Large language models (LLMs) are carrying research toward semi-automation. Automated research systems already propose hypotheses, run experiments and report their findings~\citep{km260520025,wang2026mechanist}, although evaluations of these systems still find missing analyses and conclusions with weak empirical support~\citep{km261000531,wang2026autoscirub}. Meanwhile, the stream of preprints keeps growing: primary computer-science submissions to arXiv rose by 36.5\% in January--October 2026 over the same months of 2025 (Fig.~\ref{fig:arxiv-growth-global})~\citep{arxivMonthlySubmissions,arxivCSArchiveStats}. A topical search now returns hundreds or thousands of recent preprints in the right vocabulary, and topical similarity can tell a researcher which of them concern the problem at hand, but not which deserve close reading.

Speed makes this choice harder. In LLM research a technique can be overtaken within a few months, whereas citations accumulate over months and years, so by the time a preprint is widely cited its idea has often been absorbed, extended or superseded. Foundational work remains indispensable, but much of it already lives in textbooks and model weights; what a researcher at the frontier needs is new work from active groups and emerging researchers before it is widely recognized. Some of that work is never recognized at all, and researchers entering a new area, who have yet to learn whose work to follow, are the most likely to lose their bearings.

\begin{figure}[t!]\centering
\includegraphics[width=\linewidth]{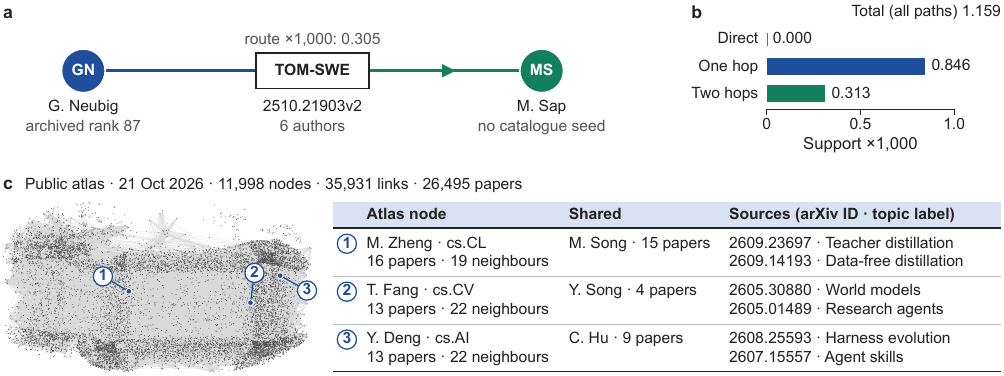}
\caption{\textbf{Collaboration paths bring researchers outside the catalogue into view.} (a,b) The Code-LLM readout matches titles and cached abstracts at the 4 November graph clock.
TOM-SWE~\citep{km251021903} connects G.~Neubig to non-catalogue author M.~Sap through a documented coauthorship path.
(a) The illustrated paper contributes $0.305\times10^{-3}$ through that one-hop route. (b) The bars sum all admitted direct, one-hop and two-hop walks into the target, giving $0$, $0.846\times10^{-3}$ and $0.313\times10^{-3}$, respectively.
The total is $1.159\times10^{-3}$.
(c) The public atlas of 21 October 2026 contains 11,998 nodes, 35,931 links and 26,495 supporting papers; the inset redraws its original layout in grey, without category colours. Markers 1--3 locate three original atlas nodes; the table repeats their cards: papers, neighbours, a shared-paper link and two source identifiers with title-derived topic labels~\citep{km260923697,km260914193,km260530880,km260501489,km260825593,km260715557}.
Readers can explore nodes and sources in the \href{https://hackerchenzhuo.github.io/knowme-research-ai/en/}{interactive website atlas}.}\label{fig:atlas}\label{fig:method-trace}
\end{figure}

Long before citations arrive, however, a preprint already carries the track records of its authors and institutions, and such records predict later impact~\citep{acuna2012future}. Researchers and institutions stake their reputations on the work they publish, and retraction studies show that a damaged reputation also changes how an author's other work is received~\citep{azoulay2017,lu2013}. Established researchers therefore have reason to be selective about what they sign, and their choices of problems and collaborators express \emph{research taste}: a judgment, formed before any results exist, about which questions deserve sustained effort. Across many papers, their involvement tends to raise the floor on quality. This judgment also travels through collaboration: when established researchers repeatedly choose to work with the same junior colleague, that choice itself signals ability~\citep{li2019coauthorship,sekara2018chaperone}, and the colleague's next paper deserves attention even when no familiar name appears on it. Authors, their collaborations and their institutions thus form a credit network: credit flows from trusted researchers to rising ones, and institutions matter through the researchers they hold.

Turning this intuition into a ranking raises two difficulties. First, a name is not an identity: two researchers can share one, and a byline can borrow a famous one. Researchers, however, tend to move among similar fields, nearby collaborators and related institutions, so a shared coauthor or affiliation can tie a new byline to its author's earlier work. Second, prominent names can attract disproportionate attention~\citep{merton1968matthew,huber2022,authority2026}. We therefore treat track records as priors over many papers, not as verdicts on any single one. Every contribution is bounded and linked to sources the reader can inspect, and signals are tested on held-out papers where enough exist, including whether they reward institution size.

KnowMe Paper Radar turns these histories into a daily reading order (Fig.~\ref{fig:teaser}). A reader chooses research directions, priorities and exclusions, and KnowMe weights each recent paper's topical fit and recency by the reputation staked on it. On the author side, this is the credit of the strongest author whose name is bound to a person, seeded from AMiner rankings~\citep{wan2019aminer,aminerai2000}, including the 2026 AI 2000 lists, and from the most productive faculty of leading CSRankings AI departments~\citep{csrankings2026}. Credit follows identities rather than names and rises nonlinearly, so top researchers stand out while no single byline dominates. On the institution side, a technical report released under an institution's name, such as ``Kimi Team'', or by a large company-led team receives a bonus and fades more slowly, because engineering reports at that scale tend to remain references for longer. Research-paper recommenders~\citep{beel2016survey} such as Scholar Inbox and PaperFlow adapt recommendations to readers' interests and histories~\citep{scholarinbox2025,paperflow2026}, and AI literature assistants such as PaperQA and OpenScholar search and synthesize the literature on request~\citep{lala2023paperqa,asai2024openscholar}. KnowMe also starts from the reader's directions, but it weights each topical match by the credit of identity-bound authors, a prior that exists on the day a paper appears.

Two further signals extend this credit network. Collaboration credit propagates along author--paper--author paths and brings researchers outside the starting catalogue into view. In a Code-LLM profile, for example, the paper TOM-SWE links the catalogue author G.~Neubig to M.~Sap, who is outside the catalogue, so a reader can follow that joint paper before deciding whether to explore the collaborator's other work (Fig.~\ref{fig:atlas}a). The same mechanism lets credit carry forward: a newcomer who leads one paper with two top researchers keeps that collaboration credit on later papers with entirely new coauthors, and coauthors share it by byline role (Sec.~\ref{sec:rising}). Each institution is profiled through a fixed panel of its researchers active in the reader's direction, so the profile adapts to the question instead of counting heads. Collaboration credit enters the score as a small capped share, and the institution profile explains each paper's context on its card (Secs.~\ref{sec:rising} and~\ref{sec:institution-rule}).

The radar also stays lightweight: daily ranking runs on a CPU without per-query LLM or embedding inference. A browser extension and a command-line interface (CLI) share one cached, versioned library on the reader's machine, so daily use needs no hosted backend or GPU, and its records and graph context are reused as interests change. Beyond daily reading, the library grows into a research resource. Each acquisition and reading session adds versioned papers, authors, affiliations and their relations to a local scholarly knowledge graph, which can be exported as a dated snapshot for graph-learning research (Sec.~\ref{sec:kg}). The CLI also returns ranked papers with their evidence as structured records, so a research agent can draw its literature input from the same source as a human reader. As autoresearch makes experiments cheaper to run, choosing which ideas to pursue, and hence which papers to read, matters more. We therefore see this traceable input as a promising entry point for future autoresearch and recursive self-improvement (RSI) workflows (Sec.~\ref{sec:plan-interface}).

We evaluate KnowMe on 174,640 acquired records and make four contributions:
\begin{itemize}
\item \textbf{A credit-centred approach to early paper discovery.} Author track records exist on the day a paper appears, long before citations. A bounded score multiplies topical relevance by the credit of the strongest identity-bound author, with a bonus for institutional technical reports; a name earns credit only after affiliations or coauthors bind it to a person (Sec.~\ref{sec:system}). In a sealed test on September--October 2026 papers, those with an identity-bound author were twice as likely to reach the top early-citation decile, and each month's credit-weighted top tenth gained 3.8 points over relevance alone in May--August and 1.1 in September--October (Sec.~\ref{sec:heldout}). A pre-declared check on those opened months found that the full score, which adds the 2026 AI 2000 and CSRankings entries, mention weights and a capped collaboration share to the score of the sealed test, did not measurably change the hit rate of each month's top tenth (Sec.~\ref{sec:full-check}).
\item \textbf{A lightweight radar.} A browser extension and a CLI share one versioned library without per-query LLM inference. Multi-level reuse merges 44,512 repeated record occurrences and makes direction switches more than three times faster (Tab.~\ref{tab:reuse-levels}; Secs.~\ref{sec:source-reuse}--\ref{sec:context-reuse}).
\item \textbf{A growing scholarly graph as a resource.} Extending the history from one month to thirteen raises the median number of authors reached beyond the seed catalogue about tenfold, and cached abstracts raise the normalized discounted cumulative gain at rank 10 (NDCG@10) of expert-judged retrieval from 0.412 to 0.598 (Sec.~\ref{sec:experiments}).
\item \textbf{A literature entry point for research agents.} In a controlled handoff with evidence packets frozen from a score variant that also weighted institution context, 79.2\% of a planner's citations carried author-history, institution or two-hop context under the full-context packets, versus 56.4\% under a lexical baseline (Sec.~\ref{sec:plan-interface}).
\end{itemize}

\begin{table}[tbp]\centering\small
\caption{\textbf{Reuse operates at several distinct levels.} Each row reports a separate reuse operation with its own unit and denominator. The source replay and complete traversal reuse cached responses from the canonical acquisition pipeline.}\label{tab:reuse-levels}
\begin{tabularx}{\linewidth}{L{2.6cm}Yr}\toprule
\rowcolor{tablehead}\textcolor{white}{\textbf{Level}}&\textcolor{white}{\textbf{Measured operation}}&\textcolor{white}{\textbf{Observed result}}\\\midrule
\rowcolor{tablepale}Paper / version&Collapse repeated eligible ID/version occurrences&44,512 repeats merged\\
Source response&Re-filter 142 cached pages under an expanded scope&1,533 IDs recovered\\
\rowcolor{tablepale}Source traversal&Reuse raw and structured pages from 508 processed pages&282 read locally\\
Metadata repair&Consolidate daily repair windows into a bounded plan&945 $\rightarrow$ 108 windows\\
\rowcolor{tablepale}Canonical library&Merge first submissions and earlier updated papers&174,640 in-scope IDs\\\bottomrule
\end{tabularx}\end{table}

\Needspace{8\baselineskip}
\section{A person-centred credit network for daily reading}
\label{sec:system}
Fig.~\ref{fig:teaser} summarizes how KnowMe turns a reader's directions into a daily reading order. The reader sets directions, priorities and exclusions over a versioned library; catalogue names earn credit only once they are bound to people (Sec.~\ref{sec:identity}); the score multiplies topical relevance by the credit of the strongest bound author, an institutional-report factor, freshness and penalties (Eq.~\eqref{eq:personal}); collaboration credit enters as a capped share, and institution profiles add context on the card (Secs.~\ref{sec:rising} and~\ref{sec:institution-rule}). The last subsections reuse the same library and graph as interests change (Secs.~\ref{sec:source-reuse} and~\ref{sec:context-reuse}).

\enlargethispage{1pt}
\subsection{The researcher-facing workflow}
A reading session begins with the researcher's question. The user selects or customizes topics, assigns priorities and a date range, and names adjacent themes to exclude; because this scope is explicit, every paper in the resulting order can be checked against the direction that admitted it. The bundled vocabulary emphasizes artificial intelligence (AI), and custom queries extend to any valid arXiv category, with visible topic names kept separate from the English retrieval terms.

The extension either maintains a standalone local library or connects to an existing one, and the two storage modes stay explicit. In both, a user can revisit cached queries, inspect excluded candidates together with the reasons for their exclusion, save papers and follow source links. Relevance, each credit factor and the optional display percentile are shown separately, so the reader can tell a content match from the credit contributed by research histories or personal preferences. App.~\ref{app:features} summarizes these controls and the agent interface, and App.~\ref{app:runtime} works through reading examples.

The research atlas turns this document-centred view into a relationship-centred one. Selecting a node leads to the papers that support an author or institutional connection, and the source and identity status explain how that connection was obtained. The public atlas presents a fixed snapshot alongside the researcher's locally maintained graph; its three examples in Fig.~\ref{fig:atlas}c show how distinct research topics lead back to compact, inspectable source records.

\subsection{Author and institutional track records}
\label{sec:track-records}
Two kinds of history help a reader assess a new byline. An author's publications show where that person has contributed over time, while work attributed to a laboratory or institution places the paper within a research programme that may predate any individual's record. AMiner's academic social network and AI 2000 catalogues provide the starting identities~\citep{wan2019aminer,aminerai2000}, together with the most productive faculty of leading CSRankings AI departments~\citep{csrankings2026}; KnowMe combines them with topic-specific author and institutional evidence from the acquired papers; repeated collaborations can then draw attention to researchers outside the initial catalogue.

The archived AMiner lists cover 2024--2025, 2022/2023--2025 and older windows. Their recorded ranks supply the cold starts, and new publications update the graph context. The catalogue also includes the 2026 AI 2000 lists (2,661 researchers after removing 83 ambiguous or risky entries): names already on an archived list keep their level, and new names enter at level B with at least 75,000 citations at top AI venues, C with 20,000 and D otherwise. It also includes the 531 faculty with at least six author-weighted papers at CCF A (full weight) and B (half weight) AI venues in 2021--2025 at the sixty CSRankings departments with the most such output, the 350 not already listed at level D; the CSRankings table (CC BY-NC-ND 4.0) is downloaded on the reader's device rather than shipped with the extension. A catalogue name without identity evidence stays a pending match: it is displayed but earns no credit (Sec.~\ref{sec:identity}).

\subsection{Identity before credit}
\label{sec:identity}
\keypoint{A catalogue name earns credit only after it is bound to a person.} A name matches a catalogue entry through its normalized form and aliases, but common names are shared, and a byline can borrow a famous one. The scorer therefore separates three states. A match is \emph{verified} when the archived source binds the paper itself, when the author's own affiliation on this paper matches an anchor of the catalogue entry, or when the catalogue places the person at the institution of a version-verified collective byline such as ``Kimi Team'' (Sec.~\ref{sec:score}). It is \emph{linked} when it shares a coauthor with an earlier paper on which the same catalogue author is verified; both bylines must have at most thirty authors, links look only backwards in time, and a linked paper never becomes an anchor itself, so credit cannot chain through namesakes. Any other match stays \emph{pending}: the browser shows it, but it earns no credit and seeds no collaboration credit. Listed affiliations that name only other institutions exclude the match altogether, whereas a collective byline excludes no one: names the catalogue places elsewhere stay pending.

The rule is a deliberately conservative form of author name disambiguation and record linkage~\citep{ferreira2012survey,subramanian2021s2and,fellegi1969linkage}. Like coauthor-based and collective resolution~\citep{kang2009coauthorship,bhattacharya2007collective}, it follows the locality of research careers, but as precision-first entity resolution~\citep{christophides2021er,konda2016magellan} it prefers an unresolved name to a wrong merge. On a pseudo-homonym benchmark built from rare names, a shared coauthor or institution recovers 76.2\% of same-person pairs while linking 0.6\% of different people in the same direction; a shared coauthor alone, the basis of the linked state, recovers 56.9\% and links 0.2\%; requiring coauthor, institution and topic together recovers only 11.5\%, and 23.8\% of true pairs share neither, so a missing link means unknown rather than different. Before the rule, 46\% of catalogue credit in the library rested on names alone, and a fixture byline that hid its affiliation gave 151 of 1,077 catalogue names a median of 2.9 points; under the rule none earns credit. Linking matters most for new papers, whose affiliations are verified later: in October 2026, 0.17\% of papers had a verified catalogue author but 14.9\% a linked one (27 November snapshot, full score with the report rule). Linked papers reach the top citation decile as often as verified ones, whereas pending names carry a weaker and exploitable signal (Fig.~\ref{fig:heldout}a).

\subsection{Credit-weighted relevance}\label{sec:score}
For paper $p$, reading profile $u$ and evidence time $t$, the reading score multiplies topical relevance by bounded credit and freshness factors:
\begin{equation}
 s(p,u,t)=R(p,u)\cdot M_a(p)\cdot M_r(p)\cdot F(p,t)\cdot \Pi(p).
 \label{eq:personal}
\end{equation}
Relevance $R$ adds the title-and-abstract topic match, a model-cue term of at most three points and the personal author fit $\Delta_a\in[-6,6]$ of Sec.~\ref{sec:author-fit}. A topic term counts fully in the title, the opening sentence, a stated contribution or more than one sentence, three quarters as a single passing mention and not at all when negated. The remaining factors act as a prior on whether a relevant paper deserves reading time, much as a document prior multiplies query relevance in retrieval~\citep{kraaij2002prior,brin1998anatomy}: a strong prior lifts a relevant paper, but eligibility still requires a selected topic, so credit cannot admit an off-topic paper. Tab.~\ref{tab:score-semantics} summarizes the role, bound and origin of each term.

\emph{Author credit.} For an identity-bound author $i$ (Sec.~\ref{sec:identity}), the scorer clips the catalogue input $b_i$ to $[0,b_{\max}]$ and assigns the credit $c_i=c_{\max}(b_i/b_{\max})^{\gamma}\ell_i$, with $b_{\max}=5$, $c_{\max}=12.5$ and convexity $\gamma=1.6$, where the byline factor $\ell_i$ weights the author's position; an identity-bound author in the highest catalogue class receives $\ell_i=1$ in any position, and any bound author on a collective byline, which lists a roster rather than roles, receives at least a small team's last-author weight, $\ell_i\geq0.68$. At full byline weight, input 5 thus contributes 12.5 points and input 2 contributes 2.9, and an explicit follow of grade $g\leq6$ counts as a bound author with credit $2.5g$. Any other author $j$ counts through a share $\phi=0.15$ of its collaboration credit $\rho_j$ (Sec.~\ref{sec:rising}), converted like a catalogue input to $\phi\,c_{\max}(\rho_j/b_{\max})^{\gamma}\ell_j$ and capped at 1.5 points. The paper's credit $C$ is the largest of these credits, plus one point when a top author is on a team of at most four, and the author factor is
\begin{equation}
 M_a=1+\frac{a}{k}\bigl(1-e^{-C/\tau}\bigr),
 \label{eq:author-credit}
\end{equation}
with strength $a=1.25$ and credit scale $\tau=8$, where $k\geq1$ is the first-author burst size defined below ($k=1$ for nearly every paper). A top author in a lead role thus gives $M_a=1.99$, input 2 gives 1.38, collaboration credit alone at most 1.21, and the factor never reaches 2.25. The convex strength curve separates the strongest catalogue levels nonlinearly, while the saturation, like term-frequency saturation in BM25~\citep{robertson2009bm25}, keeps a single byline from dominating a list. Only the strongest author counts: with the strongest credit held fixed, a second bound author raised the chance of an above-median citation rank only slightly (risk ratio 1.09) and left list-level hit rates unchanged (App.~\ref{app:tuning}).

Credit is also shared across first-author bursts, in which one researcher posts several closely related papers in quick succession. If a paper's first author has $k-1$ other first-author papers submitted within 30 days of it, before or after, each sharing a coauthor with it and having a title--abstract TF--IDF cosine~\citep{salton1988termweighting} of at least 0.3 to it, the paper keeps $1/k$ of its credit lift in Eq.~\eqref{eq:author-credit}, whether a bound author or collaboration credit supplies it; an earlier paper's credit is therefore shared once a later sibling appears. Only first authorship defines a burst, because later positions often reflect courtesy authorship, and an explicit follow is never shared. This post hoc rule, which the sealed test did not include, was prompted by two closely related papers posted six days apart and is reported separately (App.~\ref{app:burst}); all other results use $k=1$, except the runs on the 27 November 2026 snapshot (Sec.~\ref{sec:full-check}, App.~\ref{app:full-score}), which apply the rule.

\emph{Technical reports.} An institutional report is a version-verified collective byline such as ``Kimi Team'', or a team of at least thirty authors led by a company affiliation. Because such reports document engineering at a scale that usually remains a reference for longer, they receive $M_r=1.25$ and a 240-day half-life, while other papers have $M_r=1$. A collective byline lists the institution's roster in an order that carries no roles, often without author-level affiliations, so the report rule binds a uniquely resolved catalogue name on it when the catalogue places the person at the same parent institution as the byline (Sec.~\ref{sec:identity}) and counts every bound author on it at least at a small team's last-author weight; names the catalogue places elsewhere stay pending. This post hoc rule was prompted by the Kimi K3 card (Fig.~\ref{fig:cases}); Sec.~\ref{sec:full-check} and App.~\ref{app:full-score} evaluate it. Cached LLM assessments of an abstract's method and evaluation passages stay out of Eq.~\eqref{eq:personal}: they are unavailable on the day a paper appears and came from the label pool later used for evaluation, so the browser shows them only as reading aids.

Freshness decays exponentially, as in literature half-lives~\citep{burton1960halflife}: $F=1$ for the first $t_0=30$ days and $F=F_{\min}+(1-F_{\min})\,2^{-(\mathrm{age}-t_0)/H}$ afterwards, counted from the original first-submission date. The half-life $H$ is 60 days for ordinary records, up to 365 days for verified strong authors and 240 days for institutional reports, and the longest applicable half-life is used rather than a sum; the floor $F_{\min}=0.35$ keeps a still-relevant older paper visible. Finally, $\Pi$ applies the unverified single-author and venue preferences as factors of at least 0.65.

For display, the browser and CLI convert the reading score into a 0--100 percentile within the complete eligible list, before pagination. For $N$ papers and descending midrank $r$, the percentile is $100(N-r)/(N-1)$, so a unique first-place paper scores 100, ties share a position, and a singleton or all-tied list has no comparative percentile. The raw score and every factor remain available for inspection; because the percentile is relative, changing the reading question changes the comparison population as well as the evidence weights.

\emph{Where the constants come from.} Only $a$, $\tau$ and $M_r$ were fitted to outcomes before the sealed test, on 2026 papers before the sealed months and under constraints declared before the search (Sec.~\ref{sec:heldout}, App.~\ref{app:tuning}). The levels of the 2026 AI 2000 and CSRankings entries, the mention weight and the collaboration rule with its share $\phi$ were set afterwards on January--August outcomes and LLM relevance labels under the same constraints and checked once on the opened September--October months (Sec.~\ref{sec:full-check}, App.~\ref{app:full-score}). The other constants were left untouched by both tunings, and many cap a factor rather than set it (Tab.~\ref{tab:score-semantics}); the burst and report rules are post hoc, and the reach constants of Sec.~\ref{sec:rising} affect only the displayed context. Cited work supports the functional forms, not these values. The sealed comparison of papers with and without a bound author involves no constant of Eq.~\eqref{eq:personal}, so the constants decide how far credit moves a list, not whether it carries signal.

\begin{table}[tbp]\centering\small
\caption{\textbf{Each kind of evidence enters as a bounded term or factor; most constants were not set on outcomes.} Relevance decides whether a paper concerns the reader's question; the factors weigh whether it deserves reading time. Constants marked Fixed stayed unchanged while $a$, $\tau$ and $M_r$ were chosen (App.~\ref{app:tuning}) and while the levels of the 2026 AI 2000 and CSRankings entries, the mention weight and the collaboration share were chosen on January--August papers (App.~\ref{app:full-score}).}\label{tab:score-semantics}
\begin{tabularx}{\linewidth}{L{3.1cm}YL{3.3cm}L{2.3cm}}\toprule
\rowcolor{tablehead}\textcolor{white}{\textbf{Evidence}}&\textcolor{white}{\textbf{Role in selection}}&\textcolor{white}{\textbf{Constants and bound}}&\textcolor{white}{\textbf{Set by}}\\\midrule
\rowcolor{tablepale}\includegraphics[height=9pt]{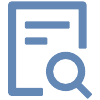}\enspace Content&Topic match and model phrases in relevance $R$&Model cue $\leq3$&Fixed\\
\rowcolor{tablepale}\quad mentions&A single passing mention counts three quarters, a negated term zero&Weight $\frac34$&Chosen, Jan--Aug\\
\includegraphics[height=9pt]{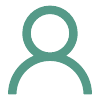}\enspace Personal author fit&Bibliography fit to the reader's directions, added to $R$&$n_0=10$, $\beta=0.5$, $\lambda=12$; $|\Delta_a|\leq6$&Fixed\\
\rowcolor{tablepale}\includegraphics[height=9pt]{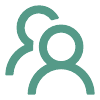}\enspace Author credit&Strongest identity-bound catalogue author, explicit follow or capped collaboration share&$b_{\max}=5$, $c_{\max}=12.5$, $\gamma=1.6$&Fixed\\
\rowcolor{tablepale}\quad factor $M_a$&Saturating lift from that credit&$a=1.25$, $\tau=8$; $1\leq M_a<2.25$&\textbf{Fitted}\\
\rowcolor{tablepale}\quad bursts&Credit shared across a first author's related papers&30 days, cosine $\geq0.3$&Post hoc\\
\rowcolor{tablepale}\quad catalogue&AI 2000 (2026) and a CSRankings slice join the archived lists&AI 2000: B/C/D; CSRankings: D&Chosen, Jan--Aug\\
\rowcolor{tablepale}\quad collaboration&Share of the strongest collaborator's credit (Sec.~\ref{sec:rising})&$\phi=0.15$, cap 1.5; $\kappa_0=2.5$, $\delta=0.5$, $E_{\max}=8$&Chosen, Jan--Aug\\
\rowcolor{tablepale}\quad\enspace graph&Window, decay and ceilings of collaboration credit&$h_c=365$, $\rho_{\max}=4.5$, $\rho^*=1.5$&Fixed\\
\includegraphics[height=9pt]{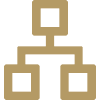}\enspace Identity&Name-only catalogue matches stay pending and earn no credit&Bylines $\leq30$ for links&Rule\\
\rowcolor{tablepale}\includegraphics[height=9pt]{assets/ant-icons/file-search-blue.pdf}\enspace Technical report&Collective or company-led large bylines; factor $M_r$&$M_r\in\{1,1.25\}$&\textbf{Fitted}\\
\rowcolor{tablepale}\quad roster&A collective byline binds members placed at its institution&$\ell_i\geq0.68$&Post hoc\\
\includegraphics[height=9pt]{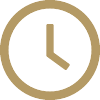}\enspace Recency&First-submission clock with type-aware half-lives; factor $F$&$t_0=30$, $H=60$--$365$ days; $F\geq0.35$&Fixed\\
\rowcolor{tablepale}\includegraphics[height=9pt]{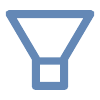}\enspace Penalties&Unverified single-author and venue preferences; factor $\Pi$&$\Pi\geq0.65$&Fixed\\
\includegraphics[height=9pt]{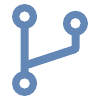}\enspace Shown, not scored&Institution context, cached reading evidence&$h_d$, $\alpha$&Unscored\\\bottomrule
\end{tabularx}\end{table}

\subsection{Author weights conditioned on a research direction}
\label{sec:author-fit}
Let $T_u$ be the user's selected topics. A priority $p_k$ maps to normalized weight $\pi_k=(5-p_k)/\max_j(5-p_j)$. For a usable bibliography title $d$, let
\begin{equation}
 m(d,u)=\max_{k\in T_u:\ d\text{ matches }k}\pi_k,\qquad
 f_a(u)=\frac{1}{n_a}\sum_{d\in D_a}m(d,u),
 \label{eq:authorfit}
\end{equation}
where an empty maximum is zero and titles shorter than eight characters are ignored. A fixed neutral fit of 50\% would penalize authors who concentrate on a narrow topic that is rare in the stored bibliography, so the reference-centred rule instead compares $f_a$ with the other usable titles in the same snapshot. Let $\mathcal B$ be the union of distinct identity-bound titles, $\mathcal B_{-a}=\mathcal B\setminus D_a$, and $N_{-a}=|\mathcal B_{-a}|$. The reference fit and shrinkage factor are
\begin{equation}
 r_a(u)=\frac{1}{N_{-a}}\sum_{d\in\mathcal B_{-a}}m(d,u),
 \qquad \eta_a=\frac{n_a}{n_a+n_0}.
 \label{eq:authorreference}
\end{equation}
\textbf{Reference-centred shrinkage.} The factor $\eta_a$, a credibility weight~\citep{buhlmann1967credibility}, attenuates the deviation from the leave-author-out reference; its prior mass $n_0=10$ regularizes sparse bibliographies. With bound $\beta=0.5$ and minimum bibliography size $n_{\min}=10$, the personal author weight is
\begin{equation}
 w_a(u)=\begin{cases}\clip\!\left(1+\eta_a[f_a(u)-r_a(u)],\,1-\beta,\,1+\beta\right),&n_a\geq n_{\min},\ N_{-a}\geq n_{\min},\\1,&\text{otherwise}.\end{cases}
 \label{eq:authorweight}
\end{equation}
An author at the snapshot's reference fit is therefore neutral, small bibliographies shrink toward that point, and the comparison excludes the author's own distinct titles. The weight stays at 1 when either bibliography has fewer than ten usable titles, and manually followed authors keep a floor of 1. App.~\ref{app:author-calibration} compares this calibration with a fixed 50\% reference on the same source records.

For a paper with $n$ authors, position shares $q_i$ are normalized over the \emph{whole} byline, including authors without a matched profile. With gain $\lambda=12$ and bound $\Delta_{\max}=6$, the personal author adjustment is
\begin{equation}
 \Delta_a(p,u)=\clip\!\left(\sum_{i=1}^{n}\operatorname{round}_{0.1}\!\left[\lambda\bigl(w_{a_i}(u)-1\bigr)q_i\right],-\Delta_{\max},\Delta_{\max}\right).
 \label{eq:authordelta}
\end{equation}
Only the departure $w_{a_i}-1$ enters the author term. In the archived ten-topic profile of 9 November, for example, the first author of GenMem has a priority-weighted fit of 17.2\% against a 5.797\% reference; with sixteen titles, shrinkage gives $w_{a_i}=1.070$, and with a byline share of about 0.278, $12(1.070-1)\cdot0.278$ contributes $+0.2$ after rounding. Under the same profile, the first author of RRSI has a 31.0\% fit and contributes $+0.6$ at $w_{a_i}=1.170$. These increments measure how well the authors' earlier work matches the chosen directions and enter relevance, whereas author credit acts separately as a factor in Eq.~\eqref{eq:personal}.

Because rounding precedes summation, some small individual increments vanish, and normalizing over the entire byline keeps a few identifiable authors from inheriting the credit of unresolved contributors. Bibliography profiles share one snapshot, and context fingerprints keep the evidence stable across result pages.

All of these weights apply only after identity binding: a compatible name must be supported by the paper's byline or by uniquely resolving repeated-coauthor evidence, and unresolved alternatives remain separate. The same principle governs large-team reports. A long byline does not create a new identity cluster, but an occurrence may reconnect to an existing high-confidence identity when at least two repeatedly observed small-team collaborators uniquely support that connection; the attachment retains those source papers and leaves ambiguous or duplicate occurrences unresolved.

\subsection{Institutions through their researchers}
\label{sec:institution-rule}
\keypoint{Institutions enter through the researchers they hold: affiliations bind identities, and a direction-specific profile explains a paper's context, but neither adds a factor to the score.} We first consolidate explicitly mapped departments and research brands into parents. Qualified IIIS, AIR and BNRist attributions share Tsinghua's parent; Google includes DeepMind; Alibaba includes Qwen, Tongyi and DAMO; ByteDance includes Seed and TikTok; Tencent includes Hunyuan; and Meta includes its named AI research groups. Affiliations come from verified author-specific evidence or an explicit, version-verified collective byline such as ``Kimi Team''; collective attribution credits the paper's group without asserting each contributor's employment, and it binds a catalogue member only where the catalogue itself places the person at that group's parent institution (Sec.~\ref{sec:identity}). An unversioned index record that repeats the same multi-institution list across several authors is held for source review.

For each parent and reading direction, the institution card summarizes the people behind a paper's context. At most fifteen direction-active researchers with the strongest attributed support fill fixed slots, so people beyond this panel do not raise the profile, and an archived field standing contributes a fixed 20\% only where the profile matches its frozen scope. The card lists the supporting papers and the researchers reached through collaboration paths, and an institution sort orders papers by this context. The resulting profiles adapt to the reader's question, so that one institution can lead one direction and trail another (Sec.~\ref{sec:institution-profiles}).

Institution context nevertheless stays out of Eq.~\eqref{eq:personal}, by design as much as by evidence. For a candidate institution factor, a documented field standing explained 98\% of its variance, so documentation rather than evidence set its level; Eq.~\eqref{eq:personal} has no institution factor. The people core does predict hits beyond author credit, adding 1.5 points to the top tenth of the check months of Sec.~\ref{sec:heldout}, but it rises with an institution's paper count and leaves frontier laboratories that publish few papers near the bottom, so it rewards size, which an institution signal should not. Size-free variants mostly measure how fully the seed catalogue covers an institution's roster, leaving Google 65th of 113, and added at most 0.8 points on the fitting months. In the sealed test, an institution's track record of early citations to its 2025 papers predicted hits beyond its people (risk ratio 1.46, and 1.31 under finer size control), but it correlated 0.58 with publication volume and reversed on declared top-venue acceptance (App.~\ref{app:institution-audit}). Until an institution signal is both size-free and useful on held-out lists, affiliations bind identities and the card explains.

\subsection{Discovering rising researchers beyond the catalogue}
\label{sec:rising}
Credit diffuses from catalogue authors to their collaborators by message passing, in the spirit of trust propagated from a seed set~\citep{gyongyi2004trustrank}. For author $a$ and a target paper $p$, each earlier joint paper $q$ within 730 days carries an endorsement from its givers $K(q)$, and the authors outside $K(q)$ share it by byline role:
\begin{equation}
 \begin{aligned}
 \kappa_a(p)&=\sum_{q}E\bigl(S(q)\bigr)\,\frac{\omega_a(q)}{\sum_{j\notin K(q)}\omega_j(q)}\,2^{-\mathrm{age}(q)/h_c},\\
 E(S)&=E_{\max}\bigl(1-e^{-S/E_{\max}}\bigr),\qquad \rho_a(p)=\rho_{\max}\left(1-e^{-\kappa_a(p)/\kappa_0}\right),
 \end{aligned}
 \label{eq:collaboration-credit}
\end{equation}
where $S(q)=\sum_{g\in K(q)}\ell_g(q)\tilde b_g$ adds the givers' inputs $\tilde b_g$ weighted by their byline factors, so two strong coauthors endorse a paper more than one but less than twice as much. The role weight $\omega_j$ halves with each byline position from 1 for the first author, while the last author of three or more, usually the senior one, counts $\frac12$; a byline of five or more authors in alphabetical order carries no role information and counts everyone equally (in the library, about 2\% of five-author bylines are alphabetical beyond chance). Because the receivers' shares sum to one, a large team dilutes the endorsement and several strong coauthors leave more of it to the others. Two rounds of message passing run. In the first, the givers are identity-bound catalogue authors, verified on the paper or linked to an earlier verified one, with $\tilde b_g=b_g$. In the second, a collaborator whose credit before $q$ reached $\rho^*=1.5$ gives $\tilde b_g=\delta\rho_g$ with $\delta=0.5$ to coauthors not yet credited, on bylines of at most twenty authors, so credit moves outward and forward in time and never returns to its source. The target paper and later records are excluded, bylines of more than thirty authors pass nothing, and names held by four or more resolved identities keep the stricter evidence rule and take no part in the second round. With half-life $h_c=365$ days, $E_{\max}=8$, ceiling $\rho_{\max}=4.5$ and scale $\kappa_0=2.5$, a newcomer who leads a recent four-author paper with two top researchers (input 5) in third and last position receives two thirds of its endorsement and $\rho_a\approx3.5$, and the second author receives one third ($\rho\approx2.4$); in a twelve-author paper with one top researcher last, the first author keeps about half and the sixth less than 2\%. Further collaborations accumulate toward the ceiling of 4.5, below the strongest catalogue level of 5, while age decay keeps old papers from inflating the credit. Identity binding requires a shared coauthor or affiliation with the earlier joint paper, and ambiguous names remain unresolved. Profiles also retain the layered evidence of repeated first- or second-author work and of collaboration with known authors, so a reader can inspect why an author was promoted. Credited authors include rising researchers but also established researchers who are missing from the catalogue or not yet bound to it.

\emph{Collaboration credit in the score.} Collaboration credit enters Eq.~\eqref{eq:personal} only as a small capped share, because in the sealed test, papers whose authors had worked with bound researchers reached the top citation decile no more often than papers whose authors had worked with activity-matched random researchers (Sec.~\ref{sec:heldout}). Author $j$ counts with $\phi\,c_{\max}(\rho_j/b_{\max})^{\gamma}\ell_j$, $\phi=0.15$, capped at 1.5 points (Eq.~\eqref{eq:author-credit}), so collaboration credit alone gives at most $M_a=1.21$: less than a lead author of input 2 (1.38), more than one at level D (1.09). The browser shades each credited author in the byline, in four levels by the factor that author's credit alone would give, and names in the score details who passed the credit on. On January--August 2026 papers, credited papers in the top two relevance fifths reached the top decile of early citations about twice as often as papers of the same month, relevance fifth and team size without a known author; sharing by role raised this ratio from 1.92 to 2.18, and that of the second round from 1.32 to 2.23, while a third round added 0.6 points of coverage with the weakest papers (App.~\ref{app:full-score}). On the opened September--October months, however, credit propagated from activity-matched random seeds predicted hits as well (Sec.~\ref{sec:full-check}). Early citations may be a poor yardstick for rising researchers; the reader study of Sec.~\ref{sec:limitations} can test the signal with readers' choices instead.

The same author--paper--author graph maps how far a profile reaches beyond the catalogue. A source paper has recency mass $d_p=2^{-\mathrm{age}(p)/h_d}$, with half-life $h_d=180$ days, within a 730-day window, and every original byline position receives $d_p/|A_p|$, including unresolved positions. Uniquely bound AMiner authors initialize a seed strength $\sigma_a=r_a^{-1/2}$ from their newest archived rank. For direction $u$, $q_a(u)=\mu_a(u)/(\mu_a+1)$, where $\mu_a$ sums an author's byline masses and $\mu_a(u)$ weights each by the priority of its matching direction; the seed vector $\mathbf{s}$ has entries $s_a=\sigma_aq_a(u)$, and the operator $P_u$ averages other positions on matching source papers, weighted by recency and topic fit; unresolved peers carry zero. Restarted propagation as in personalized PageRank~\citep{haveliwala2002topic,gasteiger2019appnp}, $\mathbf{x}^{(t+1)}=\alpha\mathbf{s}+(1-\alpha)P_u\mathbf{x}^{(t)}$ from $\mathbf{x}^{(0)}=\mathbf{s}$, stopped after two steps, gives
\begin{equation}
 \mathbf{x}^{(2)}=\underbrace{\alpha\mathbf{s}}_{\mathrm{direct}}+
 \underbrace{\alpha(1-\alpha)P_u\mathbf{s}}_{\mathrm{one\ hop}}+
 \underbrace{(1-\alpha)^2P_u^2\mathbf{s}}_{\mathrm{two\ hops}},
 \label{eq:graph-primary}
\end{equation}
with restart weight $\alpha=0.65$.
Fig.~\ref{fig:method-trace} instantiates this decomposition on an observed source path. An archived AMiner profile initializes at most one compatible resolved identity, and if the available anchors leave several possible owners, the profile remains unresolved rather than being copied across them. The browser, extension and CLI share source preparation, identity anchors and score factors, so a new profile rebuilds only the affected context, and new source records enrich later decisions without repeating the entire preparation.

\subsection{A graph that evolves with reading}
Reading can also extend this evidence base. With the reader's opt-in, the extension records local observations of public Scholar profiles; it bounds how often it observes, records provenance and gives interaction-derived attention a 30-day half-life, so user activity stays distinct from publication evidence. Pausing invalidates in-flight observations before writeback, which leaves researchers in control of how their browsing shapes future context.

A complementary author-profile view summarizes continuity and change in an author's research direction by comparing the current and previous two calendar years with weights 1, 0.5 and 0.25. It compares trajectories only over complete windows with adequate samples and stays separate from the bibliography-fit rule of Eq.~\eqref{eq:authorweight}, so a reader can explore a new direction without penalizing authors who have moved away from their earlier topics.

\Needspace{6\baselineskip}

\subsection{Automatic source reuse when a reading scope expands}
\label{sec:source-reuse}

\keypoint{A wider paper-date scope re-filters cached source responses.} KnowMe stores paper records by arXiv identifier and merges versioned observations, so cross-listings and updates never create extra papers. Ordinary online discovery still checks the source's metadata-harvesting interface (OAI-PMH)~\citep{lagoze2001oai} for new identifiers and versions, while for date-range acquisition the canonical pipeline stores complete raw response chains together with their source windows, observation dates and checksums.

When the requested paper-date range expands, the planner finds compatible, fully traversed source windows and replays their cached responses through the new selection rule, recovering papers that an old response contained but the old paper-date boundary excluded. Cached continuation tokens are consumed locally and never sent back to the source, and every response checksum and the recorded end of pagination must match. The planner then refreshes the source-datestamp tail, overlapping the earliest reused observation by one day, so paper-date selection and source-update discovery keep separate clocks. Because the process reuses existing canonical records and the original compressed source files, it creates neither a second paper library nor duplicate payload copies.

Metadata repair follows the same reuse principle: a complete containing source window can be reused when its version is compatible, while nearby repair days are consolidated into a bounded span. Retained responses cannot overwrite newer verified records, and incomplete or corrupted pagination is not treated as reusable source evidence. App.~\ref{app:source-reuse} provides semantic-equivalence controls alongside the original acquisition counts.

In a larger analysis of 142 real response pages from 40 completed windows, the expanded paper range retains 153,797 record occurrences, representing \textbf{109,285 unique papers}, while \textbf{44,512 repeated ID/version occurrences} collapse during merging. Re-filtering these responses recovers \textbf{1,533 identifiers absent from the starting library} without any additional source request, and the 104.3 MiB of compressed responses are referenced in place (App.~\ref{app:source-reuse} reports the distinct denominators). The completed traversal of 1 October 2025--3 November 2026 retains \textbf{174,640 in-scope papers}, a net addition of \textbf{46,638} over its 128,002-record in-scope baseline. Of its 508 processed raw and structured response pages, \textbf{282 are read locally}, and metadata verification passes for 174,631 records while flagging nine for field conflicts.

\subsection{Reusing the graph when a research direction changes}
\label{sec:context-reuse}
\input{experiments/context_reuse_v41}
Personal-context preparation offers a second opportunity for reuse, separate from source acquisition: like recycled intermediate results in a database~\citep{ivanova2010recycling}, a prepared source graph can be shared across reading directions. Both cost studies use titles and cached abstracts, and the preparation comparison covers 174,640 in-scope records: six directions each request exactly 1,000 eligible papers from the thirteen-month pool, and three matched repetitions compare fresh source preparation with a source graph retained after a request in a different direction, leaving the target direction uncached in both conditions (Tab.~\ref{tab:context-reuse}).

Retaining the source graph cuts mean response preparation from 151.04--168.27 seconds to 41.77--55.47 seconds, a 3.03--3.64-fold ratio of means, and all eighteen matched pairs return identical scores and source bindings; fresh preparation reads 350 compact source pages, whereas graph reuse reads none. These timings predate the identity index of Sec.~\ref{sec:identity}, which a retained service builds once per library revision and date (about 7 seconds for 159,349 records on the development laptop) and then reuses across directions. Successive reading questions can thus share source preparation while computing their own contextual weights.

Prepared responses also hold up over loopback HTTP. With one, four or eight CPU workers and one to sixteen concurrent requests, all 5,760 measured requests across 48 conditions and three repetitions return the requested 500 or 1,000 records. At sixteen concurrent requests, eight workers deliver 79.61--103.33 requests/s for 1,000-record responses and 141.79--184.56 requests/s for 500 records (Tab.~\ref{tab:response-detail}), and with one worker and one request, batch-median latency averages 64.72--85.80 milliseconds at 1,000 records. These responses bind supplied eligible records to prepared author and parent-institution contexts; retrieval and final ranking are separate steps.

Both studies run Node.js v26.4.0 on a Linux host with Intel Xeon Platinum 8369B CPUs and 800 GiB of memory, without GPU computation. The paired study times the native response call through serialization and the HTTP study adds transport of the complete response body; both exclude initial source loading and worker setup, and each worker has an identical local database copy and its own retained graph state.

Scoring itself is cheap. On two Apple M4 Pro workstations with Node.js v26.4.0, each paper costs 0.14--0.24~ms to score and sort in pools of 10,000, 50,000 and 100,000 papers, with a mean resident memory of 0.91--1.45~GiB (Tab.~\ref{tab:scale}). Per-paper cost thus stays flat as pools grow tenfold. Larger sources remain untested, but acquisition proceeds by date window, identity links stay within bylines of at most thirty authors, and daily scoring grows only with the eligible pool.

\section{Reusable graph evidence for researchers and agents}
\label{sec:kg}
\subsection{A dataset-producing workflow}
As it grows, the local library accumulates stable identifiers, versions, bylines, identity bindings, affiliations and source-specific links. KnowMe exports this evidence as a dated, scoped snapshot that keeps selection rules and unresolved relations alongside the observed edges, so a researcher can reconstruct the evidence available for a subfield, compare snapshots or reuse their relations in another analysis. Revision identifiers, observation times and SHA-256 content fingerprints bind each export to its inputs; a downstream controller can therefore save the manifest with an experiment decision and later use it to check shared papers, authors or edges when constructing evaluation splits.

\input{experiments/graph_export_table_v40}
\input{experiments/graph_export_scale}

\subsection{A tool interface for research agents}
The storage and scoring engine is separate from its consumers: a standalone local interface serves individual reading, and a persistent local service retains graph context across requests. The CLI exposes the same source preparation, identifiers and a structured output schema to automated research workflows, so an agent can invoke it as a subprocess-backed tool, supply a research profile and frozen metadata, and receive JSON or JSONL without parsing the browser interface. Ranked records keep source identifiers, abstracts, score components and missing-evidence states, and graph snapshots keep typed relations and provenance; a literature-retrieval agent, hypothesis planner or benchmark-comparison agent can therefore request evidence through one output contract. Independent CLI processes prepare their own contexts, whereas the persistent-service timings of Tab.~\ref{tab:context-reuse} apply to retained service state. App.~\ref{app:cli} gives concrete commands for profile discovery, JSONL results, RAG document extraction and scoped graph export.

The CLI specifies only process invocation and output. A Model Context Protocol (MCP) server would add a separate tool-protocol adapter, and a Skill would supply workflow instructions for selecting inputs and checking results; the product registers no KnowMe MCP tool, and its local maintenance Skills are operating guidance rather than a distributable agent integration. An RSI controller may consume the returned evidence, but the subsequent intervention, evaluation and outcome-driven update remain its own.

For retrieval-augmented generation, an agent can use the selected abstracts and source links as a first-stage evidence set before fetching full text. For training-data work, the outputs identify source candidates and preserve their lineage, while permissions, factual checks and contamination controls decide whether candidates become supervision.

\subsection{Typed citations and scientifically reusable snapshots}\label{sec:citation-snapshots}
Citation edges come from the existing OpenAlex route~\citep{priem2022openalex}. A binding must agree in arXiv identifier, title and author order, ambiguous matches are rejected, and only a uniquely bound work's \texttt{referenced\_works} field supplies directed citation edges. These edges therefore describe the observed index snapshot rather than a version-confirmed reference section, and source failures, unresolved bindings and missing lists stay distinct from returned empty lists.

A small integration check refreshed three public identifiers and exported 75 nodes and 78 relations, including 57 citation edges. Re-exporting locally reused the observation, and replacing an earlier snapshot with 70 citation edges removed obsolete assertions rather than appending them indefinitely, which establishes a working update path on this sample. Tab.~\ref{tab:graph-export-scale} reports reference-list coverage for the larger 14,461-paper pool, including unobserved lists.

Three properties make the snapshots reusable: reconstructible selection rules, source-traceable relations and explicit observation times. Together they let a consumer find shared authors or repeated source edges across splits and separate missing observations from absent relations. The schema and custom retrieval queries can accommodate further research fields, although extending beyond the LLM-related scope evaluated here requires domain-specific identity binding and a fresh assessment of the resulting reading workflow.

\section{Evaluation on an expanding research library}
\label{sec:experiments}
\noindent\textbf{Evaluation roadmap.} A reader who trusts a group's past work needs to know whether a new paper fits the present question, whose earlier contributions support its priority, and which unfamiliar authors the shortlist brings into view. After fixing the corpus and reading task, we first test against held-out early citations whether identity-bound credit picks out the papers that later stand out (Sec.~\ref{sec:heldout}). We then measure how credit changes the lists a reader inspects first and what it costs in topical focus (Sec.~\ref{sec:current-reading-response}), and whether institution profiles adapt to the reader's direction without rewarding size (Sec.~\ref{sec:institution-profiles}). The remaining studies treat the library as a resource: how far longer histories reach beyond the catalogue (Fig.~\ref{fig:year-history}), what cached abstracts add to expert-judged retrieval (Sec.~\ref{sec:abstract-evidence}) and what graph learning recovers (Sec.~\ref{sec:learned-graph}). Reuse measurements establish the cost of revisiting these judgments as the library grows (Secs.~\ref{sec:source-reuse}--\ref{sec:context-reuse}), and Sec.~\ref{sec:plan-interface} traces selected papers into a research agent's proposed experiment. Fig.~\ref{fig:evaluation-map} maps these studies onto their shared source evidence.

\input{sections/evaluation_roadmap_v34}

\input{experiments/discovery_v15_main}

\section{Related work}
\label{sec:related}
\paragraph{Personal discovery and scholarly evidence.}
Research-paper recommenders learn from content, citations and reader histories~\citep{beel2016survey}. Personalized discovery systems such as Scholar Inbox and PaperFlow adapt paper recommendations to readers' interests~\citep{scholarinbox2025,paperflow2026}, while Westlake Scholar supports institutional scholarly workflows~\citep{westlake2026}. At a broader scale, SciAtlas organizes scholarly context and InnoEval combines multiple sources for research-idea assessment~\citep{sciatlas2026,innoeval2026}. KnowMe addresses the earlier decision of which newly released arXiv papers to inspect, using explicit reading scope and source-traceable historical priors in an economical daily workflow, so it works for new readers and new papers alike, the cold-start case of recommendation~\citep{schein2002coldstart}; Tab.~\ref{tab:related} positions these systems by their principal task.
\paragraph{AI literature assistants.}
Retrieval-augmented assistants help researchers work through the literature once they know what to ask. PaperQA answers a researcher's question with an agent that searches full-text papers and cites its sources~\citep{lala2023paperqa}; OpenScholar synthesizes citation-backed answers from a large corpus of open-access papers~\citep{asai2024openscholar}; and scite classifies citation statements as supporting, contrasting or mentioning~\citep{nicholson2021scite}. Incoming citations and their intent show how a paper has been received, which a preprint cannot show on its first day, and question-driven assistants run model inference for every request. KnowMe works one step earlier: on the day a preprint appears, it orders the new papers in a reader's directions without per-query inference, and the papers it selects can then seed the deeper search and synthesis these assistants provide.

\paragraph{Research agents and evaluation.}
Research agents increasingly connect literature to actions and evaluation. AutoSciRub constructs task-specific evaluation rubrics~\citep{wang2026autoscirub}, Mechanist links scientific graphs and method skills to hypothesis testing~\citep{wang2026mechanist}, and DataCOPE develops reusable analytic skills through downstream tasks~\citep{qiu2026datacope}. Their different objectives share a need for inspectable literature inputs. KnowMe supplies that earlier selection stage: scoped records and their supporting sources can accompany a plan into these broader research workflows.
\begin{table}[tbp]\centering\fontsize{9}{11}\selectfont
\caption{\textbf{Roles in a research workflow.} The comparison organizes cited systems by their principal task and their relationship to KnowMe.}\label{tab:related}
\setlength{\tabcolsep}{4pt}\renewcommand{\arraystretch}{1.14}
\begin{tabularx}{\linewidth}{L{3.3cm}L{5.05cm}Y}\toprule
\rowcolor{tablehead}\textcolor{white}{\textbf{System}} & \textcolor{white}{\textbf{Primary task}} & \textcolor{white}{\textbf{Role relative to KnowMe}}\\\midrule
\grouphead{3}{I.\enspace Personal literature discovery}
Scholar Inbox~\citep{scholarinbox2025} & Recommendations, feedback, maps & Personalized discovery comparison\\
PaperFlow~\citep{paperflow2026} & Daily profiles and adaptation & Sequential recommendation comparison\\
Westlake Scholar~\citep{westlake2026} & Institutional scholarly workflows & Repository-centered discovery\\\midrule
\grouphead{3}{II.\enspace AI literature assistants}
PaperQA~\citep{lala2023paperqa} & Question answering over full-text papers & Question-driven search after scoping\\
OpenScholar~\citep{asai2024openscholar} & Citation-backed literature synthesis & Synthesis of selected work\\
scite~\citep{nicholson2021scite} & Citation-intent classification & Reception after citation\\\midrule
\grouphead{3}{III.\enspace Grounded scientific workflows}
InnoEval~\citep{innoeval2026} & Research-idea assessment & Evidence-intensive evaluation\\
SciAtlas~\citep{sciatlas2026} & Scholarly graph and context assembly & Shared infrastructure and retrieval\\
Mechanist~\citep{wang2026mechanist} & Mechanism discovery and experiments & Scientific orchestration\\
AutoSciRub~\citep{wang2026autoscirub} & Task-specific rubrics and revision & Evidence-consuming agent layer\\
DataCOPE~\citep{qiu2026datacope} & Reusable analytic skills & Downstream improvement setting\\\midrule
\rowcolor{knowlight}\textbf{KnowMe} & \textbf{Timely personal reading selection} & \textbf{Browser triage and reusable evidence}\\\bottomrule
\end{tabularx}
\end{table}

\begin{figure}[t!]\centering
\includegraphics[width=\linewidth]{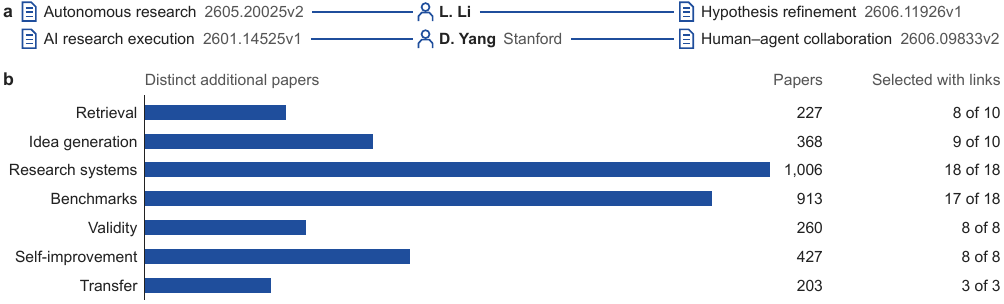}
\caption{\textbf{From selected work to a source-supported neighborhood.}
(a) Two shared-author routes retain abbreviated names, short research descriptors and exact source versions; their links denote shared authorship.
(b) Distinct additional papers for each of the seven research functions (bars and the Papers column) and, on the right, the selected papers with additional neighbors out of all selected papers in that function.
Seventy-one of 75 selected works have graph support; 3,693 selected--neighbor links connect 3,105 distinct additional papers.
Function-level neighborhoods may overlap. These are source-bound associations, not citation or mentorship links.}\label{fig:landscape}
\end{figure}
\paragraph{Research workflows and their connections.}
Recent research connects literature to retrieval, ideation, experimentation, evaluation and iterative improvement. App.~\ref{app:landscape} organizes 75 relevant works along these functions, providing a broader view of the workflows that could consume reusable evidence. Fig.~\ref{fig:landscape} adds source-supported connections between selected works: AutoResearchClaw and hypothesis-tree refinement~\citep{km260520025,km260611926} share L.~Li, while execution-grounded AI research~\citep{km260114525} and CollabSkill~\citep{collabskill2026} share D.~Yang. Both endpoints of the latter route include Stanford affiliation evidence. Such links help a reader move from one relevant system to related work through an identifiable publication record.

\paragraph{Graphs, representations, and resource access.}
ToolSpace derives typed tool relations from observed execution traces and evaluates their downstream use in agent environments~\citep{dong2026toolspace}. It provides a useful example of keeping graph construction, source evidence and downstream validation connected. AMiner, OpenAlex and similar global graphs such as the Microsoft Academic Graph and OAG provide established foundations for scholarly entities and relations~\citep{wan2019aminer,priem2022openalex,wang2020mag,zhang2019oag}, and the Open Graph Benchmark (OGB) adds learning tasks, splits, metrics, and baselines~\citep{hu2020ogb}. KnowMe reads such graphs as sources, with OpenAlex supplying citation edges and early-citation outcomes and AMiner the starting catalogue, while its own library keeps only a reader's scope (Sec.~\ref{sec:expanded-discovery}). Against the clustering and classification methods and benchmarks of author name disambiguation~\citep{ferreira2012survey,subramanian2021s2and}, the identity rule of Sec.~\ref{sec:identity} is a narrow, precision-first variant: it decides only whether a byline may inherit a catalogue author's credit, keeps unresolved names visible and never merges through a linked paper. As in dataset versioning and provenance work~\citep{huang2017orpheusdb,buneman2001why}, the exported snapshots record the version and source of every relation. Learned representations such as SPECTER2 and generative rerankers such as RankGPT offer powerful alternatives~\citep{singh2023scirepeval,sun2023rankgpt}. Hosted inference and cached representations can reduce client-side hardware requirements while retaining server-side computation, storage, and service costs. \knowme{} instead keeps ordinary metadata ranking free of per-query embedding or LLM inference, while optional bridges, graph-export runtimes, and downstream agents retain their own requirements.
\setcounter{topnumber}{1}\setcounter{totalnumber}{1}

\Needspace{10\baselineskip}
\section{Discussion: from evidence to research value}
\label{sec:future}\label{sec:validation}
\subsection{An interface for literature-informed RSI}
\label{sec:plan-interface}
\begin{figure}[t!]\centering
\includegraphics[width=\linewidth]{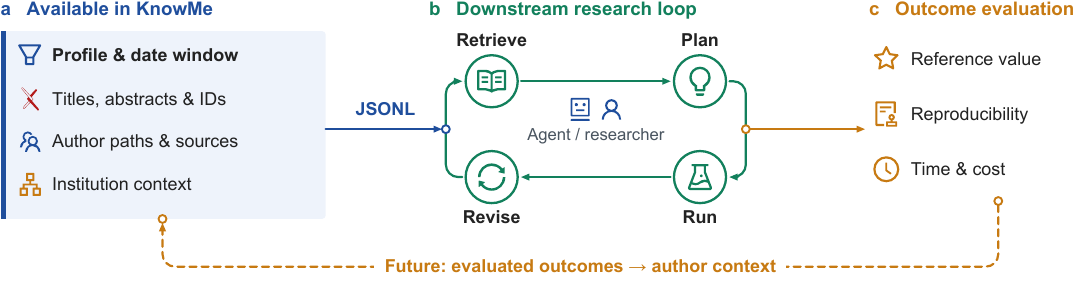}
\caption{\textbf{An integration boundary for research and RSI systems.} KnowMe supplies scoped evidence and source links. Downstream planning, interventions and independent outcome evaluation can consume these records; an outcome-driven reputation update is a future mechanism.}\label{fig:rsi-boundary}
\end{figure}
The ability to run many experiments makes the choice of which ideas to try more consequential; Chen and Schulman discuss this distinction between executing research and exercising research taste in the context of recursive self-improvement~\citep{wang2026workingtheory}. An agent deciding which baseline to implement has already inherited a choice about which papers are worth considering, and KnowMe can make that choice explicit: its literature input is traced back to particular work and can be reconsidered as the research question changes.

Concretely, an RSI controller can call the CLI with a topic profile, date window and reading budget and keep the returned JSONL manifest with its experiment plan, so that source identifiers, versions and fingerprints connect the literature input to later decisions. Fig.~\ref{fig:rsi-boundary} locates this interface in the research cycle, and App.~\ref{app:cli} gives the commands. RRSI and ModularRSI, which study constraints and evaluation separation for reusable harness changes~\citep{km260924972,km260914857}, are examples of the downstream loops that could consume such input.

\paragraph{A controlled literature-to-plan handoff.}
The handoff experiment selects fixed evidence packets across sixteen directions at budgets of twenty and fifty papers, pairing full-context packets with a title/abstract lexical baseline that uses the same fixed age penalty. The packets were frozen on 13 November 2026 from a full-context ranking that also weighted the fifteen-slot institution context, which Eq.~\eqref{eq:personal} leaves out.

A frozen Qwen3.8-27B planner receives full titles and abstracts, with author names, affiliations, ranking scores and arm labels withheld, and produces a source-linked hypothesis, baseline choices and an experiment plan. Fixed deterministic generation settings yield 64 plans, and a separate Jev~\citep{typesafe2026jev} assessment compares all 32 shuffled pairs for supported specificity, feasibility and source faithfulness. Because the planner and task are held fixed, differences reflect how the supplied evidence shapes the proposed experiment.

\input{experiments/planning_handoff_v42}

Among 56 schema-valid plans, every cited identifier belongs to the supplied evidence pool, and 94.8--98.5\% of quotes occur literally in their sources across the four conditions. Plan scores change little: full-context minus baseline specificity averages $+0.024$ at twenty papers and $+0.073$ at fifty, with direction-level SDs of 0.297 and 0.365, and feasibility differences are $+0.012$ and $-0.028$. What changes is which research histories reach the plan (Tab.~\ref{tab:planner-v42}b). At fifty papers, a mean 79.2\% of citations carry at least one history, institution or two-hop indicator under full context, against 56.4\% under the lexical baseline, and institution-supported shares are 58.3\% and 6.7\%; restricted to the selected experiment baselines, the shares with any such context are 78.6\% and 50.0\%. The slightly higher specificity mean (3.11 versus 3.04) thus accompanies substantially greater exposure to traceable research histories, as institutional and collaboration context flows from the selected pool into the agent's proposed baselines.

\subsection{Limitations and a prospective test}
\label{sec:limitations}
Four limits qualify the evidence. The credit strength of 1.25 was chosen after the May--August check months were seen, and on the already-opened September--October papers the list-level gain shrinks to +1.1 points ($-0.2$ to $2.4$) while only 73\% of newcomer winners are kept (Sec.~\ref{sec:heldout}). At fifty reads, credit-weighted lists are judged slightly less relevant ($-0.052$ on the 0--3 scale), and over long pools the head concentrates on prolific researchers (Sec.~\ref{sec:current-reading-response}). Collaboration credit has not beaten its placebo in either the sealed test or the pre-declared check of the full score, and names found only on the 2026 AI 2000 lists or in the CSRankings slice carried little hit signal (Sec.~\ref{sec:full-check}); institution profiles follow institution size and the coverage of the seed catalogue (Secs.~\ref{sec:institution-rule} and~\ref{sec:institution-profiles}). Finally, all outcomes so far are early citations, which the prospective study below replaces with readers' choices. Large reports without author-level affiliations needed a post hoc rule of their own, which binds only the catalogue members placed at the byline's institution, as the following example shows.

\begin{figure}[t!]\centering
\includegraphics[width=\linewidth]{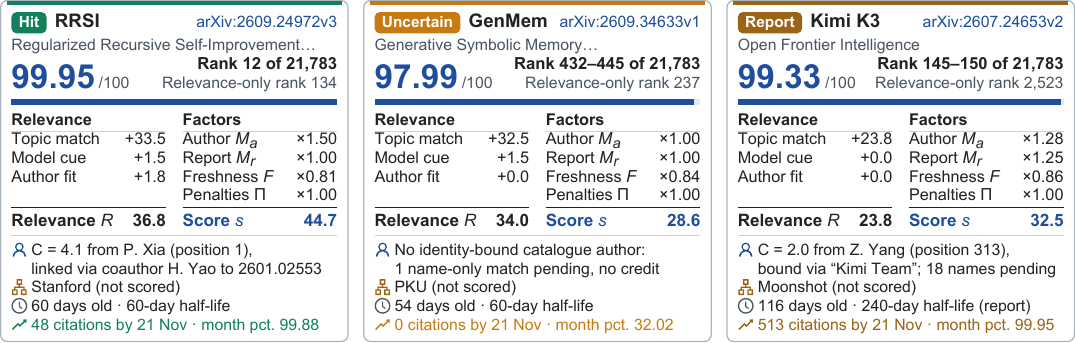}
\caption{\textbf{Every factor of a paper's score is visible: a coauthor-linked hit, a paper whose only catalogue name is pending, and a large technical report whose byline binds three members.} In a common reading pool, reasoning, model architecture, RSI and agent harnesses have equal priority, admitting 21,783 papers under the full score on the 27 November 2026 snapshot at the 21 November 2026 clock; each card lists the relevance terms, every factor of Eq.~\eqref{eq:personal}, the identity evidence behind author credit, the institution shown beside the score but not scored, the rank under relevance alone and early citations, with factors rounded for display and author names abbreviated. RRSI~\citep{km260924972} ranks 12th (percentile 99.95) and 134th under relevance alone: its relevance of 36.8 includes a personal author fit of 1.8, its first author and a coauthor are each linked to an earlier verified paper through the other, giving credit 4.1 and an author factor of 1.50 (the first author's collaboration credit, 1.3 points after the share, is smaller and does not count), and freshness is 0.81 at 60 days; its 48 citations by 21 November place it at the 99.9th percentile of its month. GenMem~\citep{km260934633} ranks 432--445 (percentile 97.99) and 237th under relevance alone: its one catalogue name is a pending match without identity evidence, so the author factor is 1.00, and with no citations by 21 November it sits at the 32nd percentile of its month. Kimi K3~\citep{km260724653} ranks 145--150 (percentile 99.33) and 2,523rd under relevance alone: its collective byline makes it an institutional report with a report factor of 1.25, the 240-day half-life keeps freshness at 0.86 after 116 days, the report rule binds the three of its 21 catalogue names that the catalogue places at Moonshot AI, of which the strongest, at byline position 313, gives credit 2.0 and an author factor of 1.28 while the other 18 stay pending, and the score uses version 2 while age remains anchored to the first submission; its 513 citations place it at the top of its month.}\label{fig:cases}
\end{figure}

\paragraph{When a technical report remains worth reading.}
A method can retain reference value across several publication cycles. Kimi K3, for example, documents an architectural change, compares its scaling efficiency with K2 and reports a weight release. Under a common four-topic profile (reasoning, model architecture, RSI and agent harnesses at equal priority) and the 21 November clock on the 27 November snapshot, it ranks 145th--150th among 21,783 papers (percentile 99.3) at 116 days of age, whereas its relevance of 23.8 alone would place it 2,523rd (Fig.~\ref{fig:cases}, right). The collective ``Kimi Team'' byline makes it an institutional report: the report factor is 1.25 and the 240-day half-life keeps $F=0.86$, where an ordinary paper of the same age keeps 0.59. Its 402 byline entries list 401 contributors after ``Kimi Team'' in surname order and match 21 catalogue names, but the paper gives no author-level affiliations, and coauthor links are not used above thirty authors. Name matching alone would credit the strongest of these names, a top-level researcher whom the catalogue places at another institution, and byline positions would treat every member as a minor author. The report rule instead binds the three names that the catalogue places at Moonshot AI, the institution the byline names, and counts them at least as senior authors: Z.~Yang, at byline position 313, gives credit 2.0 and $M_a=1.28$, and the other 18 names stay pending. Without the rule, Kimi K3 ranks 983rd--1,009th. RRSI ranks 12th, carried by two coauthor-linked authors, and GenMem, with near-maximal relevance but only a pending name, 432nd--445th (Fig.~\ref{fig:cases}, left and centre).

These three papers were chosen before the sealed test opened its outcomes and serve as illustrations, not evidence: RRSI is a hit by early citations, GenMem sits below its month's median, and Kimi K3 is a large technical report. Across the development months, 78\% of the 36 institutional reports reached the top citation decile, yet without the report factor their median position in a month's list is near the middle; 1.25 is the smallest factor within 0.005 of the best hit rate on the fitting months. Five papers drawn by hash from the October top tenth of the four-direction pool under the sealed-test score show ordinary cases (Tab.~\ref{tab:examples-v68}).

\paragraph{Toward a prospective reader study.}
Early citations measure attention, not taste: they favour fields and teams that cite quickly, and credit can further lift already visible authors. All evidence here comes from sixteen LLM directions; another field would need its own catalogue, would refit $a$ and $\tau$ under the same constraints and might need longer half-lives, since literatures age at different rates~\citep{burton1960halflife}. Papers first submitted from 23 November 2026, except 13 with relevance labels (App.~\ref{app:full-score}), are reserved for a prospective test of the full score, and a blind reader study will interleave credit-weighted and relevance-only lists without names or scores, so that readers' choices, rather than citations, decide whether credit and rising-star signals help. Signals publicized by a ranking can also invite strategic authorship. The burst rule of Eq.~\eqref{eq:author-credit} answers one form of it, and the prospective test tracks how bound credit and the gains it brings change over time. Signals that have not passed the held-out protocol enter below the top catalogue levels or capped; the post hoc report rule adds no source of credit, only bindings and byline weights for catalogue authors.

\section{Conclusion}
LLM-assisted research is making papers faster to produce than to evaluate, and in fast-moving fields the moment a preprint is most worth reading often comes long before citations can identify it. KnowMe Paper Radar fills this gap with evidence that exists on the first day: the track records of a paper's authors and the credit that flows through their collaborations. Researchers' care for their reputations and their taste in problems and collaborators tend to raise the floor on what a reader finds, and binding names to people keeps that credit with those who earned it. In a sealed test on 10,386 new papers, those with an identity-bound author were twice as likely to reach the top tenth of early citations. Collaboration credit, which may yet bring rising researchers into view before their names are familiar, did not pass the same test, and institution profiles, which adapt to each direction, still favour large institutions. The radar stays lightweight, since daily ranking needs no LLM inference and the browser extension and CLI share one reusable library. The same records form a scholarly knowledge graph for further study and could give research agents a traceable literature input, a promising direction for future autoresearch and recursive self-improvement. A prospective reader study will next decide whether credit and rising-star signals help readers, and other fields will need their own catalogues and a refit of $a$ and $\tau$. Like most papers in the field today, this one was written with substantial LLM assistance; as such assistance becomes the norm, knowing whose judgment stands behind a paper matters more, not less.

\section*{AI USE STATEMENT}
Generative AI tools assisted software implementation and data analysis, manuscript drafting, Chinese--English translation and language editing, secondary checks of mathematical expressions, and the preparation and refinement of figures. AI-assisted outputs were checked against the implementation, experiment records and source literature, and revised where necessary. The authors take full responsibility for the research decisions, claims and final content of this paper.

\paragraph{Acknowledgments.}
We thank arXiv, AMiner, CSRankings and OpenAlex for open data.

\Needspace{8\baselineskip}
\begingroup\let\FloatBarrier\relax\fontsize{9.2}{10.6}\selectfont\renewcommand{\bibfont}{\fontsize{9}{10.3}\selectfont}\setlength{\parskip}{0pt}\setstretch{1}\setlength{\bibsep}{.25pt}\raggedright
\interlinepenalty=0
\bibliographystyle{knowme-unsrtnat}\bibliography{main}\endgroup
\interlinepenalty=0

\appendix
\setcounter{topnumber}{2}\setcounter{totalnumber}{2}
\setcounter{topnumber}{3}\setcounter{totalnumber}{4}
\section{A research map for literature-informed workflows}
\label{app:landscape}

\input{literature/landscape_readout_v04}

\input{literature/categorized_citations}

\section{Expanded response and dense cost records}
\label{app:expanded}

\subsection{A matched-code retest on a larger-memory workstation}
The workstation comparison uses 106,043 distinct papers from retained history and September candidates. Forty-one nested workload sizes range from 1,000 to 100,000 papers. A recorded scoring implementation, input order, evidence clock and preparation procedure are held fixed across hosts, giving 246 Base/Personal observations on each workstation. Every matched output agrees (Tab.~\ref{tab:scale}). Workload sizes follow the same randomized schedule and method order alternates; the current graph-reuse service is evaluated separately in Tab.~\ref{tab:context-reuse}.

\begin{table}[tbp]\centering\small
\caption{\textbf{Identical ranking workload on two workstations.} Mean $\pm$ sample SD over three runs per host, size and method. This scoring benchmark uses Apple M4 Pro and Node.js v26.4.0 on both hosts; Tab.~\ref{tab:context-reuse} separately evaluates the persistent service on the Linux CPU host with the same Node.js version. During concurrent workstation use, core count, operating system and background workload differ. RSS includes input, prepared models and runtime state. The measurements compare the two complete workstation configurations.}\label{tab:scale}
\setlength{\tabcolsep}{4pt}\renewcommand{\arraystretch}{1.18}
\input{experiments/hardware_comparison_table}
\end{table}
\begin{table}[tbp]\centering\small
\caption{\textbf{Prepared HTTP responses under sixteen concurrent requests.} The recorded service configuration uses one, four or eight CPU workers for 500- or 1,000-record responses. Values are mean $\pm$ sample SD over three batches; latency is the mean of batch 95th percentiles. Bold marks the largest mean throughput per direction and size. All 48 conditions remain in the numerical supplement. Timing includes transport, context binding and serialization of supplied eligible records.}\label{tab:response-detail}
\setlength{\tabcolsep}{4pt}\renewcommand{\arraystretch}{1.18}
\input{analysis/visual_v42/service_v42/TABLE11_BODY}
\end{table}
\begin{figure}[t!]\centering
\includegraphics[width=\linewidth]{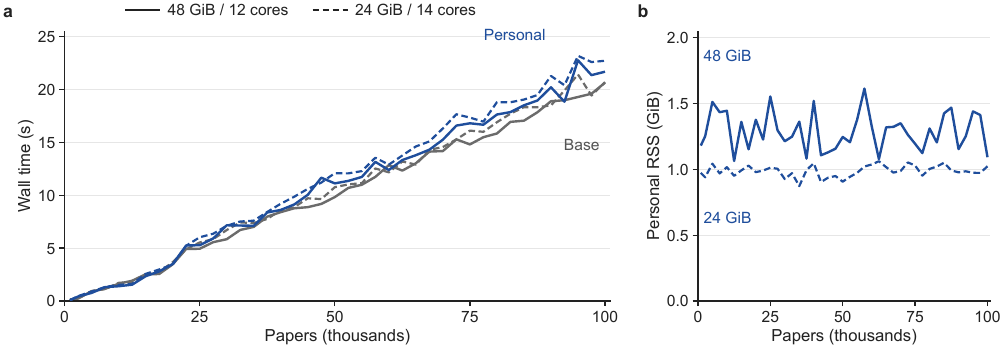}
\caption{\textbf{Scoring cost across forty-one workload sizes.} Each host contributes 246 measurements across two scoring modes and three repeats per workload. (a) Base (gray) and Personal (blue) scoring/sorting time; lines connect the means of three runs, solid for the 48 GiB, 12-core host and dashed for the 24 GiB, 14-core host. Sample SDs are at most 4.30 s; Tab.~\ref{tab:scale} lists them at 10k, 50k and 100k papers. (b) Personal RSS on both hosts; sample SDs are at most 0.42 GiB on the 48 GiB host and 0.22 GiB on the 24 GiB host. On the 48 GiB host, process CPU time is 0.86--1.36 times wall time across the 246 individual runs (median 0.99). The workstations differ in core count, operating system and ambient load. Timings cover scoring and sorting; loading, preparation, hashing, network access and browser IPC are measured separately.}\label{fig:latency}
\end{figure}
At 100k papers, the 48 GiB, 12-core machine takes \pmval{20.64}{0.68} s for Base and \pmval{21.68}{0.11} s for Personal, compared with \pmval{20.68}{0.50} s and \pmval{22.71}{1.31} s on the earlier 24 GiB, 14-core machine. The new Personal mean is 4.6\% lower; the Base means are almost identical. New-host Personal RSS averages 1.09 GiB, and process CPU time is close to elapsed time at this endpoint. Observed prepared-context cost is similar on the two hosts despite their different installed memory. One cold preparation takes 68.63 s on the new host versus 63.68 s previously.

Both workstations use normal caching and garbage collection during ordinary concurrent use. The earlier measurements were collected with other applications running and memory reported near capacity, providing a practical stress condition. Core counts, operating systems and ambient workloads also differ, so the comparison characterizes these two usage conditions without isolating the effect of memory capacity.

\paragraph{Complete-library CLI cost.} A separate fresh-process cost check uses 128,005 unique local records in a read-only SQLite snapshot. With the explicit RAG/Reasoning profile, all 12,441 eligible papers are ranked before five records are returned. Three fresh-process invocations take \pmval{152.44}{14.67} s end to end, including input reading, graph and personalization preparation, scoring and output; peak process RSS reaches 3.53 GiB. Returned identifiers and scores are identical across runs. The runs use normal workstation caching. The whole-library measurement exposes preparation costs: a small output limit alone does not make cold agent access fast.

\FloatBarrier
The supplementary CSV contains all 41 points; Fig.~\ref{fig:latency} plots every measured scale.

\section{Typed graph coverage in earlier source snapshots}
\label{app:graphs}

\subsection{Graph completeness exposes the limits of citation-only discovery}
\begin{table}[tbp]\centering\small
\caption{\textbf{Typed graph diagnostics on direction subcohorts.} Samples overlap and are not additive. Bound is the count with a successfully bound reference-list field; Nonempty counts source papers with a nonempty list. Empty lists retain their observed index state.}\label{tab:graphcounts}
\setlength{\tabcolsep}{5pt}\renewcommand{\arraystretch}{1.12}
\begin{tabularx}{\linewidth}{Yrrrrrrr}\toprule
\rowcolor{tablehead}\textcolor{white}{\textbf{Direction}} & \textcolor{white}{\textbf{Papers}} & \textcolor{white}{\textbf{Authorship}} & \textcolor{white}{\textbf{Coauthor}} & \textcolor{white}{\makecell{\textbf{Shared}\\\textbf{author}}} & \textcolor{white}{\textbf{Cites}} & \textcolor{white}{\textbf{Bound}} & \textcolor{white}{\textbf{Nonempty}}\\\midrule
\rowcolor{tablepale}
Agents&500&3,646&182,302&27&44&203&2\\
RAG&500&2,383&7,977&13&110&266&4\\
\rowcolor{tablepale}
Reasoning&500&3,883&187,959&27&13&207&1\\
Medical AI&314&1,802&8,673&17&116&160&2\\
\rowcolor{tablepale}
Multimodal&500&3,248&16,045&23&54&253&2\\
Safety&500&2,919&16,355&14&119&256&4\\\bottomrule
\end{tabularx}
\end{table}

An earlier audit exports up to 500 hash-ordered matching papers per direction (314 for Medical AI; Tab.~\ref{tab:graphcounts}). Authorship, co-authorship, author-mediated paper links, and citation assertions remain typed separately.

The September 18--19 OpenAlex cache binds a reference-list field for 40.6--53.2\% of each subcohort, but only \textbf{one to four papers per direction have nonempty lists}. Binding coverage and nonempty-reference coverage therefore describe separate source fields. Large-team bylines also inflate co-author clique projections. Preserving original authorship and source-paper evidence makes both limitations visible.

The counts use source observations dated September 18--19, 2026. These graph diagnostics do not feed candidate papers back into the historical ranking model.
\enlargethispage{4pt}
\Needspace{7\baselineskip}
\section{CS archive counts and segmented growth models}
\label{app:growth}

\subsection{Observed growth and conditional continuations}
\begin{figure}[t!]\centering
\includegraphics[width=\linewidth]{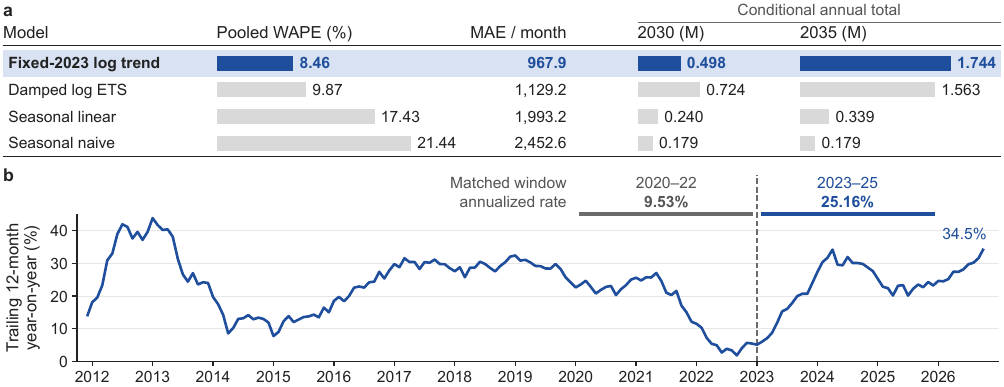}
\caption{\textbf{Forecast sensitivity and observed CS growth.} (a) Pooled weighted absolute percentage error (WAPE), mean absolute error (MAE), and conditional annual totals in millions of articles for all four declared models, sorted by WAPE; the 2030 and 2035 bars share one scale. Errors pool five overlapping twelve-month holdouts; the 60 forecast instances per model are not independent replications. The tinted fixed-2023 log trend has the lowest pooled WAPE. Damped log ETS denotes error--trend--seasonal exponential smoothing. The model range spans three fitted continuations. (b) Year-on-year change in trailing twelve-month primary-CS totals; the dashed line marks January 2023, the fitted period split. Bars along the top span the matched 2020--2022 and 2023--2025 windows, and the value above each bar is that window's annualized rate from the matched model; Tab.~\ref{tab:models} gives their Newey--West heteroskedasticity- and autocorrelation-consistent (HAC) standard errors.}\label{fig:growth}
\end{figure}

\label{sec:growth}
Fig.~\ref{fig:arxiv-growth-global} shows the complete 2010--October 2026 observed series. The projection fits use \keypoint{January 2020--October 2026}, retaining the original 82-month primary-CS window~\citep{arxivCSArchiveStats}. Incoming cross-listings are excluded, and CS months follow the archive's identifier-based counter. Growth is slower in 2020--2022, rather than absent: the earlier matched-window model estimates 9.5\% annualized growth, compared with 25.2\% in 2023--2025. The January 2023 marker is a fixed descriptive boundary.

We compare a seasonal-naive baseline, a seasonal linear trend, a log-linear trend with a fixed 2023 level/slope change, and a damped trend with seasonality fitted on log counts. Expanding-window forecasts start at five dates from October 2023 to October 2025 and each predict the next 12 months. The fixed-break log model has the lowest pooled weighted absolute percentage error (WAPE), \textbf{8.46\%}, followed by damped log ETS at 9.87\%, linear trend at 17.43\%, and seasonal naive at 21.44\%. The model is selected by this holdout criterion, not by the height of its long-range projection. App.~\ref{app:projection} gives the protocol, equations and software~\citep{seabold2010statsmodels}.

Continuing the selected fitted trend implies approximately \textbf{0.498 million new primary-CS articles in 2030} and \textbf{1.744 million in 2035}, or monthly averages of 41.5k and 145.4k. Across the three trend models, excluding the seasonal-naive reference, annual projections span 0.240--0.724 million in 2030 and 0.339--1.744 million in 2035. These differences expose sensitivity to the assumed growth law. The one-year holdouts do not validate the nine-year horizon; archive adoption, policy, research activity or publication practice may change the trajectory.

The observed near-term increase already matters for discovery: January--October counts rise from 114,533 in 2025 to \textcolor{knowgreen}{\textbf{156,364 in 2026 (+36.5\%)}}; October rises from 13,732 to \textbf{20,240 (+47.4\%)}. For a reader with budget $B$ and inspection cost $c$, direct coverage of $N$ candidates is $B/(cN)$. Growing supply makes timely selection more consequential, regardless of which long-range scenario eventually holds.

Seventeen official annual archive pages provide the 2010--2026 monthly counts. The official counter groups documents and separates primary listings from cross-listings. We retain both columns in the data package and use only primary-CS new articles in the analysis. The complete sequence contains 202 months, ending in October 2026, and appears in Fig.~\ref{fig:arxiv-growth-global}. The local retrieval date is 2 November 2026.

For monthly count $y_t$, with January 2023 at $t_0$, the primary specification is
\begin{equation}
\log y_t=\alpha+\beta t+\delta\,\mathbb{1}(t\geq t_0)
+\gamma(t-t_0)_+ + \sum_{m=2}^{12}\eta_m\mathbb{1}(\mathrm{month}(t)=m)+\epsilon_t.
\end{equation}
\begingroup\interlinepenalty=10000
Both level and slope can change. Annualized rates are $100[\exp(12\beta)-1]$ before the boundary and $100[\exp(12(\beta+\gamma))-1]$ afterward. Ordinary least squares (OLS) uses Newey--West heteroskedasticity- and autocorrelation-consistent (HAC) covariance with 12 lags and finite-sample correction. Displayed $\pm$ values are one delta-method HAC standard error of the annualized rate. The original asymmetric confidence intervals are retained in the analysis sidecar. The equal-window comparison (Tab.~\ref{tab:models}) uses 72 months with shared seasonal effects. The plan was fixed before fitting, after inspection of the source layout and selected raw counts; no breakpoint was searched to maximize apparent acceleration.\par
\endgroup

\begin{table}[tbp]\centering\small
\caption{\textbf{The post-period slope is higher under both declared windows.} Rates are annualized changes in fitted monthly counts, with one delta-method HAC standard error. These estimates have different windows and do not identify an AI treatment effect.}\label{tab:models}
\renewcommand{\arraystretch}{1.22}\setlength{\tabcolsep}{8pt}
\begin{tabularx}{\linewidth}{Yrr}\toprule
\rowcolor{tablehead}\textcolor{white}{\textbf{Comparison window}} & \textcolor{white}{\textbf{Pre-2023 rate (\%)}} & \textcolor{white}{\textbf{Post-2022 rate (\%)}}\\\midrule
\rowcolor{knowlight}\textbf{Equal complete windows: 2020--2022 / 2023--2025} & \pmval{9.53}{1.88} & \textbf{\pmval{25.16}{1.46}}\\
Full history: 2010--2022 / 2023--October 2026 & \pmval{22.74}{0.72} & \textbf{\pmval{29.50}{2.05}}\\\bottomrule
\end{tabularx}
\end{table}

Within the full-series log-response model family, a fixed-break segmented model has nominal Bayesian information criterion (BIC) $-221.7$, compared with $-211.4$ for a single trend with the same month effects. This is an in-sample fit diagnostic, not predictive validation. BIC values are not compared across observation counts or between level-count and log-response models. The matched level-count sensitivity estimates a rise in the monthly-count trend from about 485 additional articles per year to 2,042; those slopes are not annual totals.

\Needspace{12\baselineskip}
\subsection{Conditional projections to 2030 and 2035}
\label{app:projection}
The projection study uses monthly source counts from January 2020 through October 2026. Its model list, validation dates and selection rule were fixed before fitting this study. Each model is refitted using only observations available at its origin: October 2023, April 2024, October 2024, April 2025 and October 2025. Every origin has a 12-month holdout. These windows overlap; their 60 predictions are not independent replications. Pooled WAPE is $100\sum|y-\PaperCopyMath{D835DC660302}{\hat{y}}|/\sum y$ across those forecast instances.

The linear candidate has an intercept, monthly trend and calendar-month indicators. The segmented log candidate adds a level change and slope change at January 2023, with the same calendar terms. Log forecasts are mapped to the count scale using the mean exponentiated in-sample residual as a smearing factor. The damped candidate applies additive trend, damped continuation and 12-month additive seasonality to log counts, with estimated initial states. The seasonal-naive baseline repeats the most recent 12 monthly observations. NumPy 2.4.4 and statsmodels 0.14.6 implement the fits~\citep{seabold2010statsmodels}.

After holdout selection, each model is fitted to all 82 observations and extended through December 2035. Annual totals sum the corresponding 12 monthly projections. The numerical range across the linear, segmented-log and damped-log candidates measures structural model disagreement; Fig.~\ref{fig:growth} summarizes their annual totals. The forecast files provide every held-out prediction, model score and monthly extrapolation. No future counts are treated as measured data.

\FloatBarrier
\section{Browser controls and agent-facing evidence}
\label{app:agent-cli}\label{app:features}
\begin{figure}[t!]\centering
\includegraphics[width=.94\linewidth]{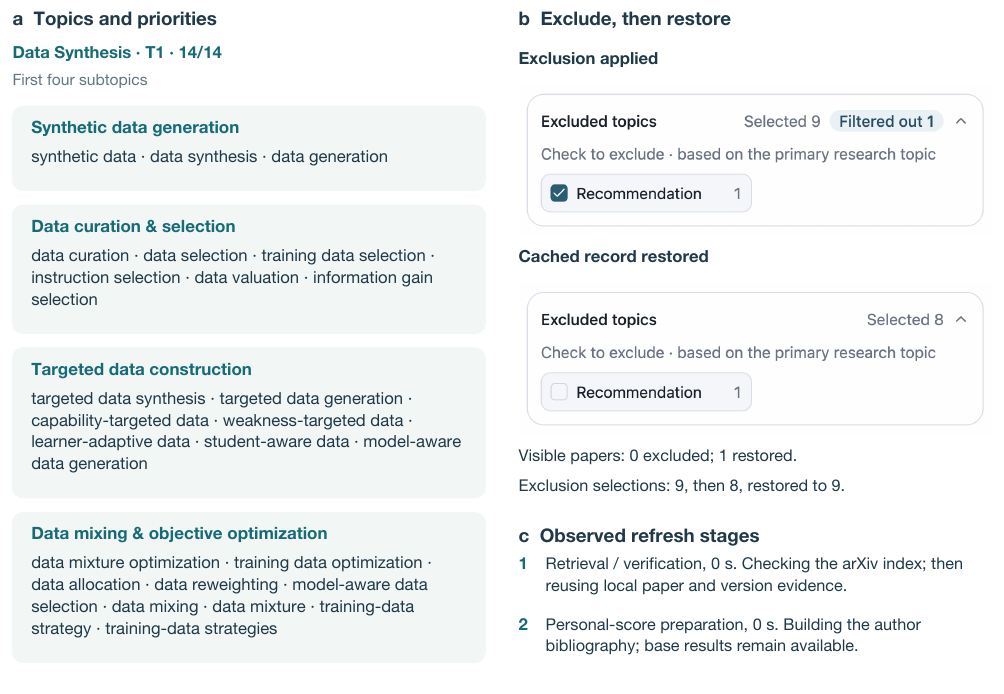}
\caption{\textbf{Scope controls make selection decisions visible.} Four topic cards preserve retrieval terms and priority. Original exclusion screenshots show one cached paper removed and restored. The adjacent stage readouts transcribe two observed preparation states. Topic and stage text is re-typeset.}\label{fig:controls}
\end{figure}

Tab.~\ref{tab:features} summarizes the interface controls, and Fig.~\ref{fig:controls} illustrates a recorded reading profile. The browser and command-line interfaces expose the same kinds of source-linked records.

\begin{table}[!tbp]\centering\small
\caption{\textbf{Implemented user-facing capabilities and their purpose.} The main text develops the scoring and data mechanisms; this map connects them to the visible workflow.}\label{tab:features}
\begin{tabularx}{\linewidth}{L{4.2cm}Y}\toprule
\rowcolor{tablehead}\textcolor{white}{\textbf{Capability}} & \textcolor{white}{\textbf{Researcher control or evidence}}\\\midrule
\grouphead{2}{\includegraphics[height=9pt]{assets/ant-icons/filter-blue.pdf}\enspace Discovery and overload control}
Topic subscriptions and priority tiers & Declare what matters and how strongly; customize directions\\
Topic exclusions & Suppress unwanted adjacent themes while preserving the positive goal\\
Dates, versions and cache state & Separate first submission, updated versions and available cached evidence\\
\grouphead{2}{\includegraphics[height=9pt]{assets/ant-icons/user-green.pdf}\enspace Interpretable personal selection}
Decomposed score views & Inspect content, author, institution, age and personal adjustments\\
Domain-specific author context & Reuse identity-bound bibliography; neutral fallback for sparse evidence\\
Adaptive institution context & Recognize topic-specific strengths through recent supported publications\\
Emerging-author candidates & Surface repeated paper evidence outside the fixed author catalog\\
\grouphead{2}{\includegraphics[height=9pt]{assets/ant-icons/branches-blue.pdf}\enspace Continuity and reuse}
Research atlas & Navigate paper-supported author and institution relationships\\
Opt-in local evolution & Add public observations with bounded, decaying attention\\
Favorites and retained bibliography & Preserve selected items and compact history across cache operations\\
Batch results and stopping & Read early results and retain an explicitly incomplete partial query\\
Typed graph export & Deliver nodes, evidence-bearing relations, triples and dated manifests\\\bottomrule
\end{tabularx}
\end{table}

\subsection{Ranked evidence packets}
From the project root, the following command selects an explicit profile, performs literal exclusions, ranks all eligible cached papers, and emits a manifest followed by five paper records:
\begin{commandpanel}[unbreakable]{Ranked evidence: explicit topic priorities and exclusions}
node scripts/rank-research-papers.mjs \
  --topics rag:1,reasoning:2 --from 2026-10-01 --to 2026-10-31 \
  --exclude survey --limit 5 --format jsonl
\end{commandpanel}

A topic priority is in 1--4, with 1 highest. An agent can instead pass frozen paper records through \mbox{\texttt{--input -}}. Output records contain identifiers/versions, titles and abstracts, a raw score decomposition, matched evidence, missingness, and shared snapshot fingerprints. The default command reads the library without changing it and makes no network or model requests. Preparation uses the complete supplied bibliography, including current candidates, with the date window selecting papers within that shared context. The output limit bounds returned items, not preparation cost. Private browser settings and manual author bonuses are not loaded.
\subsection{Typed graph snapshots}
\label{app:cli}
The following commands run from the project root. The first uses cached evidence only. The second refreshes a bounded set of citation records before export; \texttt{paper-ids.json} contains an array of public arXiv identifiers. Snapshot directories must be new, and Tab.~\ref{tab:cli-files} lists the files each export writes. The default one-page acquisition limit processes at most 100 selected identifiers per call, with progress retained for continuation.
\begin{commandpanel}[unbreakable]{Graph snapshots: cached export and bounded citation refresh}
node scripts/export-knowledge-graph.mjs \
  --category cs.CL --query retrieval --limit 1000 \
  --output snapshots/retrieval-20261101

node scripts/export-knowledge-graph.mjs --ids paper-ids.json \
  --refresh-citations --max-pages 1 \
  --citation-cache evidence/citations-20261101 \
  --output snapshots/citations-20261101
\end{commandpanel}

\begin{table}[!hbp]
\centering\small
\caption{\textbf{Machine-readable graph delivery.} The relation-rich export and the simple triple view serve different consumers. Both views distinguish unobserved reference metadata from observed citation counts.}\label{tab:cli-files}
\begin{tabularx}{\linewidth}{L{5cm}Y}
\toprule
\rowcolor{tablehead}\textcolor{white}{\textbf{Output}} & \textcolor{white}{\textbf{Information retained}} \\
\midrule
\rowcolor{tablepale}
\texttt{nodes.jsonl} & Typed IDs; paper version/date; author identity basis and resolution status \\
\texttt{edges.jsonl} & Subject, predicate, object, direction, and source/derivation evidence \\
\rowcolor{tablepale}
\texttt{triples.tsv} & Subject--predicate--object only; detailed provenance remains in the edge records \\
\texttt{citation-coverage.jsonl} & Coverage state; source time; bound work; available reference count \\
\rowcolor{tablepale}
\texttt{manifest.json} & Scope, time, input/code digests, relation counts and output hashes \\
\bottomrule
\end{tabularx}
\end{table}
Filters match explicit categories, literal title/abstract text, or selected identifiers. The selection is not a guarantee of semantic field coverage or the newest available papers. Export reads the existing library without changing it; optional source acquisition writes an export-side evidence cache. It does not install an automatic monitoring service or publish a dataset.

\input{experiments/cli_examples}

\section{Author-reference choices and paired comparisons}
\label{app:author-priors}
\input{literature/aminer_method_context}
\label{app:author-calibration}

\input{experiments/author_calibration_refresh_v32}

\Needspace{7\baselineskip}
\section{Reading priorities in a common candidate pool}
\label{app:runtime}
The common pool uses the score of Eq.~\eqref{eq:personal} as in the sealed test at the 21 November 2026 clock. Reasoning, model architecture, RSI and agent harnesses all have priority one, admitting 21,838 papers. Reading percentiles compare scores within this pool and give ties half credit. Fig.~\ref{fig:cases} shows three illustrative papers of these directions under the full score on the 27 November snapshot, where the pool admits 21,783 papers, with every term of their scores; Tab.~\ref{tab:examples-v68} adds five papers drawn by rule. Author names are abbreviated.

\input{analysis/revision_v68/TABLE_EXAMPLES_v68}

\section{Source attribution and large-team identity attachment}
\label{app:institution-network}
The current institutional readout combines two source relations. Author-specific affiliation evidence links a person to an institution in a paper; a version-verified collective byline links the paper itself to the named group. An exact, uniquely resolved catalogue alias is required for the latter, and a title or abstract mention is insufficient. Collective attribution supplies institutional context without asserting individual employment. For author credit, the report rule binds a catalogue member on such a byline only when the catalogue itself places the person at the named group's parent institution (Sec.~\ref{sec:identity}).

Large-team bylines do not form weak identity clusters. An occurrence may attach to an existing high-confidence identity when at least two repeatedly observed small-team collaborators appear uniquely on that byline, and exactly one candidate identity satisfies the condition. Duplicate names, initial-only given names and conflicting candidates remain unresolved. Each attachment records its supporting coauthors and source-paper identifiers. The full byline remains in the propagation denominator.

The fixed-corpus comparison preserves the identity population and original affiliation assertions while adding 1,007 source papers and 12,917 supported author occurrences. Controls cover exact collective aliases, title-only negatives, version mismatches, ambiguous identities, double-source deduplication, unresolved collective-only records and seedless components. Caching normalized source names reduces repeated preparation while preserving the complete resulting snapshot byte for byte.

\section{Source-window reuse and record-level deduplication}
\label{app:source-reuse}
\begin{table}[tbp]\centering\small
\caption{\textbf{Reuse at the source and paper levels.} The large analysis parses the same cached responses used by the canonical ingestion pipeline. Repeated source occurrences are not additional papers; panel B starts from 128,002 in-scope identifiers.}\label{tab:source-reuse}
\begin{tabularx}{\linewidth}{L{8.5cm}Y}\toprule
\rowcolor{tablehead}\textcolor{white}{\textbf{Quantity}} & \textcolor{white}{\textbf{Observed value}}\\\midrule
\grouphead{2}{A. Retained real source responses}
Complete source windows / response pages & 40 / 142\\
All observed source entries & 158,868\\
Occurrences retained by the expanded paper scope & 153,797\\
Unique eligible paper IDs (and unique ID/version pairs) & 109,285\\
Repeated eligible ID/version occurrences & 44,512\\
IDs absent from the 128,005-record starting library & 1,533\\
External requests for replayed pages / new payload copies & 0 / 0\\
Compressed source reused in place (MiB) & 104.3\\\midrule
\grouphead{2}{B. Completed expanded-range traversal}
Canonical in-scope IDs / net new IDs & 174,640 / 46,638\\
First submissions / revision-only records & 148,971 / 25,669\\
Raw / structured pages processed & 240 / 268\\
Raw / structured pages reused locally & 142 / 140\\
Daily repair windows / consolidated plan & 945 / 108\\
Verified metadata / unresolved field conflicts & 174,631 / 9\\\bottomrule
\end{tabularx}
\end{table}
The requested paper range is 1 October 2025--3 November 2026 (UTC). The existing canonical acquisition workflow includes cs.AI, cs.CV, cs.CL, cs.LG and cs.IR, independently of topic subscriptions. A paper is in scope when an original submission or revision lies in that range. First-submission counts and revision-only records must therefore be reported separately. Panel~A of Tab.~\ref{tab:source-reuse} isolates the retained-source replay, and panel~B reports the completed range traversal. The current canonical set contains 148,971 first submissions and 25,669 revision-only records. Nine visible field conflicts keep traversal completion distinct from metadata verification.

The standard CLI enables source-window reuse and repair-window consolidation automatically; no separate deduplication command is required. Consolidation retains started windows and their continuation state. A regression control preserves a newer verified record when an older cached response contains the same identifier. It retains a checkpoint and stops on a source rate limit. The scope, page digests, per-window traversal state, changed records and metadata-coverage results are saved together.
\begin{commandpanel}[unbreakable]{Canonical acquisition with automatic reuse}
node scripts/backfill-arxiv.mjs \
  --from 2025-10-01 --to 2026-11-03 --plan

node scripts/backfill-arxiv.mjs \
  --from 2025-10-01 --to 2026-11-03
\end{commandpanel}
The plan performs no external requests or canonical writes. A resumed run uses its existing evidence directory and frozen dates. Source-datestamp coverage, metadata verification, affiliation evidence and author identity resolution are reported as separate stages.

\input{experiments/legacy_discovery_appendix}

\input{experiments/discovery_evidence_appendix}

\setcounter{topnumber}{2}\setcounter{bottomnumber}{2}\setcounter{totalnumber}{3}
\Needspace{10\baselineskip}
\section{Institutional readouts, scope and complete-direction checks}
\label{app:graph-primary}
\input{experiments/institution_evidence_v68}
\input{experiments/full_score_check}

\input{experiments/semantic_appendix_v28}
\input{experiments/semantic_protocol_v28}
\input{experiments/trainable_graph_protocol_v32}

\input{experiments/data_value_appendix_v30}

\input{experiments/reading_choice_joint_protocol_v32}
\setcounter{topnumber}{2}\setcounter{bottomnumber}{2}\setcounter{totalnumber}{3}
\input{sections/semantic_portrait_appendix_v30}

\input{sections/lexical_appendix_v31}

\end{document}

%% file: analysis/final_review_v54/caption_copy/CAPTION_COPY_FINAL_PATCH.tex
\pdfmapline{=zpl-Regular-tlf-sc-t1 TeXGyrePagellaX-Regular " encntx-ec-tlf-pc ReEncodeFont " <[ntx-ec-tlf-pc.enc <analysis/final_review_v54/caption_copy/zplr-caption-sc.pfb}
\AtBeginDocument{%
  \DeclareCaptionFormat{copyable}{%
    \begingroup
    \everymath\expandafter{\the\everymath\pdfinterwordspaceoff\aftergroup\pdfinterwordspaceon}%
    \pdfinterwordspaceon
    #1#2#3\par
    \pdfinterwordspaceoff
    \endgroup
  }%
  \captionsetup{format=copyable}%
}

%% file: analysis/glyph_fix_v38/ADDITIONAL_GLYPH_MAP.tex
\pdfglyphtounicode{parenleftbig}{0028}
\pdfglyphtounicode{parenrightbig}{0029}
\pdfglyphtounicode{bracketleftbig}{005B}
\pdfglyphtounicode{bracketrightbig}{005D}
\pdfglyphtounicode{parenleftbigg}{0028}
\pdfglyphtounicode{parenrightbigg}{0029}
\pdfglyphtounicode{parenleftBigg}{0028}
\pdfglyphtounicode{parenrightBigg}{0029}
\pdfglyphtounicode{braceleftBigg}{007B}

%% file: experiments/context_reuse_v41.tex
\begin{table}[tbp]\centering\small
\caption{\textbf{Paired preparation for 1,000 returned records.} The recorded service measurements comprise three paired runs per direction (18 pairs), reported as mean $\pm$ sample standard deviation (SD); scores and bindings match in every pair. The target direction is uncached in both modes; retained-source requests follow a different-direction prime. Timing spans the native response call and complete serialization from loaded inputs. Fresh requests read 350 source pages; retained requests read none. Ratios use unrounded means. RAG denotes retrieval-augmented generation.}\label{tab:context-reuse}
\setlength{\tabcolsep}{4pt}\renewcommand{\arraystretch}{1.12}
\input{analysis/visual_v42/service_v42/TABLE3_BODY}
\end{table}

%% file: analysis/visual_v42/service_v42/TABLE3_BODY.tex
\begin{tabularx}{\linewidth}{Yrrrr}\toprule
\rowcolor{tablehead}\textcolor{white}{\textbf{Direction}}&\textcolor{white}{\makecell{Actual\\records}}&\textcolor{white}{\makecell{Fresh source\\(s)}}&\textcolor{white}{\makecell{Retained source\\(s)}}&\textcolor{white}{\makecell{Ratio of\\means}}\\\midrule
\rowcolor{tablepale}
Agent systems & 1,000 & \(168.27\,{\color{gray}\scriptstyle\pm}\,{\color{gray}8.24}\) & \(55.47\,{\color{gray}\scriptstyle\pm}\,{\color{gray}2.96}\) & 3.03$\times$\\
RAG & 1,000 & \(157.46\,{\color{gray}\scriptstyle\pm}\,{\color{gray}3.73}\) & \(45.72\,{\color{gray}\scriptstyle\pm}\,{\color{gray}2.12}\) & 3.44$\times$\\
\rowcolor{tablepale}
Reasoning & 1,000 & \(151.37\,{\color{gray}\scriptstyle\pm}\,{\color{gray}3.31}\) & \(42.07\,{\color{gray}\scriptstyle\pm}\,{\color{gray}0.73}\) & 3.60$\times$\\
Multimodal learning & 1,000 & \(151.04\,{\color{gray}\scriptstyle\pm}\,{\color{gray}1.48}\) & \(42.34\,{\color{gray}\scriptstyle\pm}\,{\color{gray}3.61}\) & 3.57$\times$\\
\rowcolor{tablepale}
Safety and alignment & 1,000 & \(152.02\,{\color{gray}\scriptstyle\pm}\,{\color{gray}1.15}\) & \(41.77\,{\color{gray}\scriptstyle\pm}\,{\color{gray}1.23}\) & 3.64$\times$\\
Efficient LLMs & 1,000 & \(155.02\,{\color{gray}\scriptstyle\pm}\,{\color{gray}1.68}\) & \(47.24\,{\color{gray}\scriptstyle\pm}\,{\color{gray}4.55}\) & 3.28$\times$\\
\bottomrule\end{tabularx}

%% file: experiments/graph_export_table_v40.tex
\begin{table}[tbp]\centering\small
\caption{\textbf{Typed evidence and delivery cost on 14,461 cached papers.} Both panels use the same complete recent-window export. Time is mean $\pm$ sample SD over three fresh processes; peak memory is resident set size (RSS). Acquisition is outside the timed operation.}\label{tab:graph-export-scale}
\setlength{\tabcolsep}{6pt}\renewcommand{\arraystretch}{1.14}
{\raggedright\color{knowgreen}\bfseries a\quad Export footprint\par}\vspace{3pt}
\begin{tabularx}{\linewidth}{@{}CCCCC@{}}\toprule
\rowcolor{tablehead}\textcolor{white}{\textbf{Paper records}}&\textcolor{white}{\textbf{Typed nodes}}&\textcolor{white}{\textbf{Relations}}&\textcolor{white}{\textbf{Time (s)}}&\textcolor{white}{\textbf{Peak resident set size (RSS, GiB)}}\\\midrule
\rowcolor{knowlight}\textbf{14,461}&\textbf{88,600}&\textbf{709,698}&\pmval{6.52}{0.32}&\textbf{2.95}\\\bottomrule
\end{tabularx}
\par\vspace{7pt}
{\raggedright\color{knowteal}\bfseries b\quad Relations and their source semantics\par}\vspace{3pt}
\begin{tabularx}{\linewidth}{@{}L{3.05cm}rL{3.0cm}Y@{}}\toprule
\rowcolor{tablehead}\textcolor{white}{\textbf{Connection}}&\textcolor{white}{\textbf{Links}}&\textcolor{white}{\textbf{Relation source}}&\textcolor{white}{\textbf{Retained evidence}}\\\midrule
\multicolumn{4}{@{}l}{\raisebox{-.5pt}{\includegraphics[height=9pt]{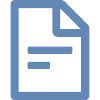}}\enspace\textcolor{knowteal}{\textbf{Source assertions}}}\\[2pt]
\rowcolor{tablepale}Paper $\rightarrow$ author&76,005&Recorded byline&Identity status and original author position\\
Paper $\rightarrow$ category&26,884&arXiv classification&Original category label\\
\rowcolor{tablepale}Paper $\rightarrow$ institution&58&Direct assertions&Verified paper-level affiliation evidence\\\addlinespace[3pt]
\multicolumn{4}{@{}l}{\raisebox{-.5pt}{\includegraphics[height=9pt]{assets/ant-icons/team-green.pdf}}\enspace\textcolor{knowgreen}{\textbf{Shared-author associations}}}\\[2pt]
Author $\leftrightarrow$ author&605,327&Coauthorship join&Supporting shared-paper identifier\\
\rowcolor{tablepale}Paper $\leftrightarrow$ paper&1,424&Resolved-author join&Common bound identity\\\addlinespace[3pt]
\multicolumn{4}{@{}l}{\raisebox{-.5pt}{\includegraphics[height=9pt]{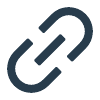}}\enspace\textcolor{knowgold}{\textbf{Indexed references}}}\\[2pt]
Paper $\rightarrow$ cited work&\textemdash&On-demand refresh&\texttt{--refresh-citations}; Sec.~\ref{sec:citation-snapshots}\\\bottomrule
\end{tabularx}
\par\smallskip\raggedright
The 58 direct affiliation assertions are paper--institution edges; institution context is computed separately from the institution's direction-active researchers (Sec.~\ref{sec:institution-rule}). The 73,962 author nodes comprise 7,364 resolved identities and 66,598 paper-scoped bylines. This cached export contains no observed citation lists; finite citation refresh is available on demand.
\end{table}

%% file: experiments/graph_export_scale.tex
\paragraph{A typed graph at the scale of a reading pool.}
The 14,461-paper recent reading pool of Sec.~\ref{sec:expanded-discovery} yields 88,600 typed nodes and 709,698 relations through the cached-only CLI (Tab.~\ref{tab:graph-export-scale}). Authorship assertions, shared-author associations and indexed citations carry different meanings, and each keeps its source. Three fresh processes take \pmval{6.52}{0.32} s for source selection, cached binding, graph construction, serialization and output, excluding source collection and operating-system (OS) cache flushing, and the fixed-clock node, edge, triple and coverage files have identical hashes across the runs.

%% file: sections/evaluation_roadmap_v34.tex
\begin{figure}[t!]
\centering
\includegraphics[width=\linewidth]{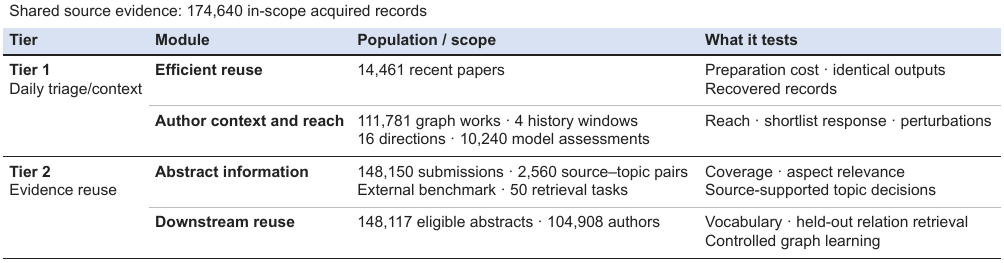}
\caption{\textbf{Two tiers of evaluation over shared source evidence.} The top line gives the shared acquisition of 174,640 in-scope records. Each row lists a module's tier, its own population or scope and what it tests: Tier 1 examines daily triage and context, and Tier 2 examines evidence reuse. The fifty expert-judged retrieval tasks come from an external benchmark.}
\label{fig:evaluation-map}
\end{figure}

%% file: experiments/discovery_v15_main.tex
\subsection{Corpus scope and the recent reading task}

\label{sec:expanded-discovery}
We first ask whether one reusable library can support timely selection across different research questions. The acquired library contains \textbf{174,640 unique in-scope paper identifiers}: 148,971 first submissions from 1 October 2025 through 3 November 2026 and 25,669 earlier papers updated during that period. Canonical identifiers collapse cross-listings and versions into single papers, and 174,631 records have verified metadata while nine retain field conflicts. Because the acquisition is AI-oriented, its scope differs from the official primary-CS archive of Sec.~\ref{sec:growth}. This scope, rather than the whole literature, sets the scale. From October 2025 through October 2026 all of arXiv received 389,047 new submissions~\citep{arxivMonthlySubmissions}, and global graphs such as OpenAlex, the Microsoft Academic Graph, OAG and Semantic Scholar integrate hundreds of millions of works~\citep{priem2022openalex,wang2020mag,zhang2019oag,ammar2018literature}; KnowMe reads such graphs as sources rather than replicating them. Its measurements therefore test robustness and usability at and beyond this design point: preparation on the 174,640 in-scope records (Sec.~\ref{sec:context-reuse}) and scoring on pools of 1,000 to 100,000 papers (App.~\ref{app:expanded}).

\begin{table}[tbp]\centering\small
\caption{\textbf{The declared LLM scope within the broader library.} Each family contains four existing directions. Counts are deduplicated within a family; papers can occur in several families.}\label{tab:llm-scope}
\setlength{\tabcolsep}{5pt}\renewcommand{\arraystretch}{1.16}
\begin{tabularx}{\linewidth}{lYr}\toprule
\rowcolor{tablehead}\textcolor{white}{\textbf{Family}}&\textcolor{white}{\textbf{Reading directions}}&\textcolor{white}{\textbf{Papers}}\\\midrule
Reasoning / learning&Reasoning; reinforcement learning (RL) for reasoning; distillation; alignment&19,507\\
\rowcolor{tablepale}Agents / retrieval&Agent harnesses; self-improvement; multi-agent; RAG&12,525\\
Model construction&Pretraining; efficiency; code; multimodal&22,440\\
\rowcolor{tablepale}Data / assessment&Data synthesis; evaluation; safety; hallucination&22,473\\\midrule
\textbf{Distinct union}&\textbf{All sixteen directions}&\textbf{58,094}\\\bottomrule
\end{tabularx}\end{table}

The longitudinal analysis follows sixteen LLM-related directions in four families (Tab.~\ref{tab:llm-scope}). The thirteen complete months from October 2025 through October 2026 contain \textbf{148,150 first submissions}, of which \textbf{58,094 distinct papers} match at least one direction. An explicit language-model expression appears in 36,958 of these records. Acquisition follows broad AI categories and title--abstract matching then defines the reading cohorts, so agents, RAG and reasoning can share papers. The broader forty-topic catalogue remains available for additional coverage checks, but neither dictionary is an exhaustive census of LLM research.

For a concrete reading task, we fix the recent 30-day window of 5 October--3 November 2026, which contains 14,461 papers before topic selection. Each of the 40 existing topics defines a priority-1 profile, and all profiles share the frozen library, the graph-based rule and the 4 November noon UTC clock. Existing topic definitions and weights determine eligibility through title/abstract matching, with no selection by citation outcomes.

Each experiment uses a frozen, matched configuration: the credit-versus-relevance replay uses the score of Sec.~\ref{sec:system} as frozen for the sealed test (the sealed-test score: $\phi=0$, unweighted mentions, archived catalogue) at the 21 November 2026 clock, while the service measurements, the author-reference ablation and the frozen reading-list and planner studies use their recorded mean-based configurations, with non-target terms fixed within each comparison; the report factor and type-aware half-lives, which those studies omit, are examined in Sec.~\ref{sec:limitations} and App.~\ref{app:runtime}. Together, corpus replays measure coverage, list response and computational reuse, expert retrieval judgments test text granularity, model assessments describe sampled topic evidence, and held-out authorship tasks examine semantic continuity. Independent reading and research-workflow utility are the downstream questions of Sec.~\ref{sec:future}.

Tables highlight favourable values within matched comparisons, including ties at the displayed precision. Confidence summaries use the estimate $\pm$ the larger 95\% half-width, with exact asymmetric limits in the numerical supplement. Sample standard deviations and missing-label bounds are identified separately.

\subsection{Held-out evidence for identity-bound credit}
\label{sec:heldout}\label{sec:adaptive-institution-test}
Identity-bound author credit predicts which new papers early citations later single out. Citations are a delayed and imperfect proxy for which papers deserve reading and also reward prominence~\citep{merton1968matthew,petersen2014reputation} (examined below), but they are computed apart from the score and arrive for every paper, so we test each part of the score against early citations in months it never saw. For the 43,870 of the 58,094 direction-matched first submissions from October 2025 to October 2026 that have an OpenAlex record~\citep{priem2022openalex}, the outcome is a paper's mid-rank citation percentile within its submission month, counted on 22 November 2026. A paper is a \emph{hit} in the top decile and clears the \emph{floor} above the month median; in September and October, when 64\% and 80\% of papers are still uncited, the floor means at least one citation. Track records of authors and institutions use only citations received by the end of 2025, and comparisons match papers on submission month and team size.

\begin{figure}[t!]\centering
\includegraphics[width=\linewidth]{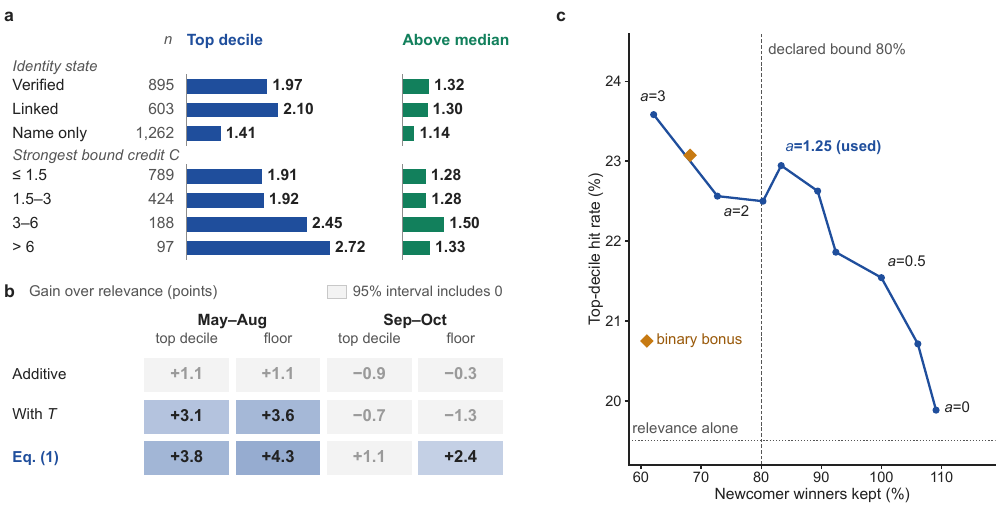}
\caption{\textbf{Identity-bound author credit predicts early-citation hits; name-only matches carry a weaker signal.} Outcomes are within-month percentiles of early OpenAlex citations; risk ratios match papers on submission month and team size against papers without any catalogue name. (a) Matched risk ratios (RR) for the top early-citation decile and for clearing the month median (floor), by identity state on January--August 2026 papers (verified on the paper, $n=895$; linked by an earlier coauthor only, $n=603$; name-only matches, which stay pending, $n=1{,}262$) and by credit $C$ of the strongest bound author (a top author in a lead role has $C=12.5$); every 95\% interval excludes 1. (b) Gain of each month's top tenth over relevance alone in the top-decile hit rate and in the floor rate, in points (grey cells: the 95\% bootstrap interval includes zero), for an additive score that adds author and institution evidence to relevance, a score that also multiplies in cached reading evidence $T$ and an institution factor, and Eq.~\eqref{eq:personal} as in the sealed test; the latter two are compared without the age factor, which a within-month outcome cannot evaluate. (c) Top-decile hit rate against newcomer winners kept on May--August papers as the strength $a$ of $M_a=1+a(1-e^{-C/\tau})$ varies ($\tau=8$; 1.25 is used); the diamond adds a binary bound-author bonus.}\label{fig:heldout}
\end{figure}

\emph{A sealed test of the modules.} We froze a protocol, the analysis code and its hash, and then opened the outcomes of 10,386 of the 11,749 September--October 2026 papers once. The 301 papers (2.9\%) with a bound author were hits twice as often as matched papers without one, 22.6\% against 11.5\% (risk ratio 1.97, 95\% interval 1.64--2.37), and cleared the floor more often (1.46, 1.30--1.63). That test's exposure also accepted shared OpenAlex institutions, whereas Fig.~\ref{fig:heldout}a and the list results use the rule of Sec.~\ref{sec:identity}, whose day-one link needs a shared coauthor, without the later report rule. Controlling for research direction, using only October papers, or counting only citations from other authors left the hit ratio between 1.72 and 1.99 and the floor ratio between 1.40 and 1.60. A second bound author gave no clear hit gain (0.96, 0.60--1.55, $n=52$). Collaboration with bound researchers did not beat activity-matched random researchers ($p=0.38$), and an institution's own track record (1.46) shrank under finer size control and reversed on declared top-venue acceptance (0.64), on which author credit was inconclusive (0.91, 0.44--1.87). Tab.~\ref{tab:sealed} lists every comparison. These results support author credit in Eq.~\eqref{eq:personal}. Institution records passed the frozen test but stay out by design because they track institution size (Sec.~\ref{sec:institution-rule}), and the report factor rests on development months, with only six reports in the sealed months.

\begin{table}[tbp]\centering\small
\caption{\textbf{The sealed test keeps author credit in the score; a pre-declared check tests the full score.} Matched risk ratios (RR) of a top-decile hit, with 95\% intervals, and of clearing the floor, on the 10,386 September--October 2026 papers with outcomes; $n$ counts exposed papers. The post hoc checks were run after opening and could only weaken the result: a research direction added to the matching strata (with the three identity states compared under the same strata), October papers only, citations only from other authors or only from other authors at other institutions, and papers submitted after the author catalogue was frozen on 19 September 2026. The institution row marks institutions in the top quartile of early citations to their 2025 papers, beyond their people; finer size control uses twenty volume bins. Only six institutional reports fall in these months. The bottom rows come from a check of the full score on the same, already-opened months and the 27 November 2026 snapshot, with the binding of Sec.~\ref{sec:identity} except the post hoc report rule, frozen with its code before the terms added to the sealed-test score met any September--October outcome and run once (Sec.~\ref{sec:full-check}); each catalogue row is matched on month and team size against papers without a catalogue name, and placebo $p$ compares a collaboration-credit ratio with 100 draws of activity-matched random seeds.}\label{tab:sealed}
\setlength{\tabcolsep}{5pt}\renewcommand{\arraystretch}{1.12}
\begin{tabularx}{\linewidth}{Yrlr}\toprule
\rowcolor{tablehead}\textcolor{white}{\textbf{Matched comparison}}&\textcolor{white}{\textbf{$n$}}&\textcolor{white}{\textbf{Hit RR, 95\% interval}}&\textcolor{white}{\textbf{Floor RR}}\\\midrule
\rowcolor{knowlight}\multicolumn{4}{l}{\textbf{Identity-bound author credit, in Eq.~\eqref{eq:personal}}}\\
\rowcolor{tablepale}Any bound author & 301 & \textbf{1.97}\,\ci{1.64}{2.37} & 1.46\\
\rowcolor{knowlight}\multicolumn{4}{l}{\textbf{Bound credit under post hoc checks}}\\
\rowcolor{tablepale}Direction added to the strata & 298 & 1.72\,\ci{1.42}{2.09} & 1.40\\
\quad verified on the paper & 39 & 2.28\,\ci{1.61}{3.23} & 1.52\\
\rowcolor{tablepale}\quad linked to an earlier verified paper & 258 & 1.71\,\ci{1.37}{2.12} & 1.39\\
\quad name only, pending & 236 & 1.37\,\ci{1.00}{1.86} & 1.13\\
\rowcolor{tablepale}October papers only & 165 & 1.94\,\ci{1.48}{2.56} & 1.60\\
Citations by other authors & 301 & 1.99\,\ci{1.61}{2.46} & 1.56\\
\rowcolor{tablepale}Independent citations & 301 & 1.83\,\ci{1.50}{2.23} & 1.59\\
Papers after the catalogue freeze & 230 & 2.06\,\ci{1.63}{2.59} & 1.53\\
\rowcolor{knowlight}\multicolumn{4}{l}{\textbf{Other signals in the sealed test}}\\
\rowcolor{tablepale}Second bound author & 52 & 0.96\,\ci{0.60}{1.55} & 0.92\\
Coauthor of a bound researcher & 1,442 & 1.18\,\ci{0.92}{1.53} & 1.05\\
\rowcolor{tablepale}Institution citation record & 2,248 & 1.46\,\ci{1.22}{1.75} & 1.19\\
\quad finer size control & 2,113 & 1.31\,\ci{1.07}{1.60} & 1.13\\
\rowcolor{tablepale}Institutional report & 6 & untested & --\\
\rowcolor{knowlight}\multicolumn{4}{l}{\textbf{Pre-declared check on the opened months}}\\
\rowcolor{tablepale}Collaboration credit, no catalogue name (placebo $p=0.46$) & 203 & 1.41\,\ci{0.98}{2.01} & 1.04\\
\quad credited, $\rho\geq1.5$ (placebo $p=0.48$) & 152 & 1.30\,\ci{0.88}{1.92} & 0.95\\
\rowcolor{tablepale}Bound author, full catalogue & 1,522 & 1.43\,\ci{1.20}{1.71} & 1.25\\
\quad only on AI 2000 (2026) & 512 & 1.05\,\ci{0.79}{1.41} & 1.08\\
\rowcolor{tablepale}\quad only in the CSRankings slice & 416 & 1.22\,\ci{0.91}{1.64} & 1.24\\
Bound author, archived lists only & 585 & 1.95\,\ci{1.63}{2.33} & 1.38\\
\bottomrule
\end{tabularx}\end{table}

\emph{Fame or the work?} Early citations also follow prominence, so credit could predict them by tracking fame alone. Yet the hit ratio stays at 1.99 without self-citations, at 1.83 when only citations by other authors at other institutions count, and at 2.06 for papers submitted after the catalogue freeze, while name-only matches, which show readers the same names, carry the weakest signal (Tab.~\ref{tab:sealed}). Merit and attention stay entangled, so we read the result as finding papers a field will attend to early. No reader follows every established group, and such papers are easy to miss: the 301 bound-author papers were 2.9\% of the sealed months' papers but contributed 68 of their 790 hits (8.6\%). Conversely, the other 91.4\% of hits had no bound author, and in a field that renews as fast as LLM research a researcher's best work can come at any point of a career, including the first papers~\citep{sinatra2016quantifying}. Credit therefore only reorders relevant papers and, like exposure and popularity-bias controls in ranking~\citep{singh2018exposure,abdollahpouri2017popularity}, the score must keep newcomers in view: the factor forms were chosen under a declared newcomer constraint (below), and collaboration credit points to rising researchers (Sec.~\ref{sec:rising}).

\emph{Identity and credit strength.} On development months (January--August 2026), papers with a verified author and papers with only a coauthor-linked author reach the top decile about twice as often as papers without any catalogue name, whereas pending names carry a weaker signal (Fig.~\ref{fig:heldout}a). Among bound authors, stronger credit predicts more hits (Fig.~\ref{fig:heldout}a), which motivates the convex strength curve; the saturating factor keeps any single byline from dominating. In January--October 2026, papers in first-author bursts reached the top decile about half as often as other papers of the same month (risk ratio 0.49, 0.30--0.70); the post hoc burst rule of Sec.~\ref{sec:score} shares their credit (App.~\ref{app:burst}).

\emph{Reading lists.} The factor forms and the report factor were chosen on January--April 2026 under two declared constraints: at most 5\% of each month's top tenth may fall below the median relevance, and at least 80\% of the newcomer winners, papers whose authors have no recorded history yet that reach the top decile, found by relevance alone must be kept. Of the 224 forms in the January--April grid, all 151 that met both constraints beat relevance alone on those months (15.2--18.7\% hits against 14.8\%), so the gain does not hinge on one setting. The grid preferred strength 1.5; we use 1.25, chosen after the May--August results were seen (App.~\ref{app:tuning}), so those months are not a clean test of this one parameter. On May--August papers, the credit-weighted top tenth of each month reaches the top decile in 23.3\% of cases against 19.5\% for relevance alone (+3.8 points, bootstrap interval 2.3--4.9) and the floor in 68.1\% against 63.9\%, with 2.8\% of picks below the median relevance and 86\% of newcomer winners kept (Fig.~\ref{fig:heldout}b). An additive score that adds author and institution evidence to relevance does not differ measurably from relevance alone. On the already-opened September--October papers the gains are smaller (+1.1 points, $-0.2$ to $2.4$; floor +2.4, 0.7--3.8), and only 73\% of newcomer winners are kept, below the declared 80\%. Fig.~\ref{fig:heldout}c shows the trade-off we accept: stronger credit raises hit rates but drops newcomer winners, and a binary bonus for any bound author would keep only 68\% of them. Papers first submitted from 23 November 2026 are reserved for a prospective test (Sec.~\ref{sec:limitations}).

\subsection{The full score on the opened months}
\label{sec:full-check}
The full score of Sec.~\ref{sec:system} differs from the sealed-test score in three terms: the 2026 AI 2000 and CSRankings entries of the catalogue (Sec.~\ref{sec:track-records}), mention weights in $R$ (Eq.~\eqref{eq:personal}) and collaboration credit at $\phi=0.15$ (Sec.~\ref{sec:rising}), all set on January--August outcomes and LLM relevance labels under the constraints above (App.~\ref{app:full-score}). The September--October outcomes had already been opened for the sealed test, so a new sealed test of these terms was impossible. Instead we froze a protocol, its code and its inputs before the full score met any of those outcomes, ran it once, and report its results.

\emph{Collaboration credit.} Among the 4,677 papers with prior author activity but no catalogue name, the 203 with collaboration credit were hits 1.41 times as often as matched papers (0.98--2.01). To ask whether this reflects the standing of the catalogue researchers the credit comes from, we replaced each of the 1,641 seed researchers by a random non-catalogue researcher of the same activity, seeded on as many papers at the same level, and reran the unchanged propagation 100 times. These placebo seeds gave a mean ratio of 1.43, and 45 of the 100 draws reached the real one ($p=0.46$). By the declared decision rule, collaboration credit is therefore not shown to reflect more than collaborators' activity. Among the 1,466 papers whose only catalogue names are unbound, those with collaboration credit cleared the floor more often (1.39, 1.13--1.71), consistent with the graph recognizing catalogue researchers that affiliations have not yet bound.

\emph{Catalogue.} On the 27 November snapshot, the full catalogue binds 1,522 papers against 585 for the archived lists alone, at a lower ratio (1.43 against 1.95; Tab.~\ref{tab:sealed}). Papers whose strongest bound author is on the archived lists keep a ratio of 1.83 (1.49--2.25), whereas names only on the 2026 AI 2000 lists (1.05) or only in the CSRankings slice (1.22) carry little hit signal, although the slice clears the floor more often (1.24, 1.08--1.42).

\emph{Reading lists.} The top-tenth hit rate did not change measurably. Age-neutral, the full score reaches the top decile with 13.6\% of its picks against 13.7\% for the sealed-test score (paired difference $-0.07$ points, $-1.05$ to $0.80$) and is 1.35 points above its own relevance alone (0.00 to 2.80). It places fewer picks below the median relevance (2.2\% against 3.6\%) and keeps 15 of 19 newcomer winners against 13 of 18; with age, 2.3\% against 3.5\% and 58\% against 56\%, so both scores miss the declared newcomer bound on these months. Age-neutral, collaboration credit alone changes 3.0\% of the top tenth and moves its hit rate by $+0.48$ points ($-0.19$ to $1.06$). Credited papers in the top two relevance fifths were hits 1.70 times as often as comparable papers without a known author, against 1.61 without sharing by role. The full score thus did not measurably change list quality.

\emph{Report rule.} The report rule of Sec.~\ref{sec:score} came after this check, so neither the sealed test nor this check includes it. Replayed on January--October 2026 in the sixteen-direction pool, it rescores ten papers, all collective reports and all early-citation hits, and changes no month's top-tenth hit rate: with age every month keeps the same top tenth, and age-neutral only June's changes, at the same hit rate (App.~\ref{app:full-score}). The case cards and the reading-list shares on the 27 November snapshot include it.

\subsection{Reading-list response at fixed budgets}

\label{sec:current-reading-response}
\begin{figure}[t!]\centering
\includegraphics[width=\linewidth]{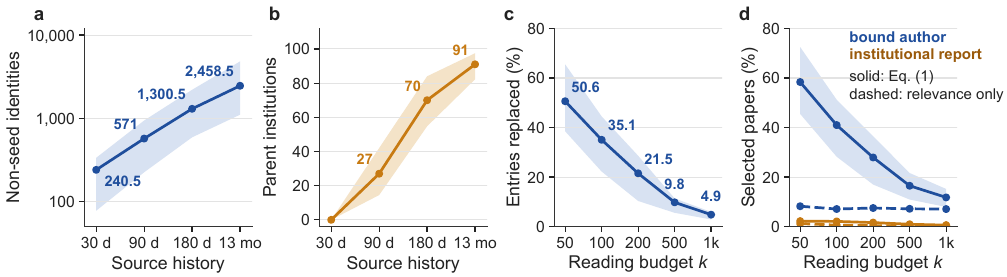}
\caption{\textbf{Historical graph coverage and the effect of credit on reading lists.} (a,b) Sixteen title-and-abstract directions use four source-history windows with fixed identities and clock; the panels show reached non-seed identities and graph-eligible parent institutions. (c,d) The sealed-test score is compared with a relevance-only ranking that keeps relevance, freshness and penalties but omits the author and report factors, at 50--1,000 reads. (c) Share of relevance-only entries replaced. (d) Share of selected papers with an identity-bound catalogue author and of institutional reports, credit-weighted (solid) and relevance-only (dashed). Lines are medians over the sixteen directions in (a,b) and means in (c,d), with values printed in (a--c); bands span the interquartile range across directions, in (d) of the credit-weighted bound-author share.}\label{fig:year-history}
\end{figure}
To measure how credit changes a shortlist, we score every eligible paper under the sealed-test score and under a relevance-only ranking that keeps the same relevance, freshness and penalty terms but drops the author and report factors. The sixteen candidate pools contain 88,643 direction--paper pairs from 58,094 distinct papers, scored natively at the 21 November 2026 clock; all directions support every budget.

Credit reshapes the head of each list. It replaces a mean 50.6\% of the relevance-only top fifty (16--80\% across directions), 35.1\% at 100 reads and 4.9\% at 1,000 (Fig.~\ref{fig:year-history}c). Papers with an identity-bound catalogue author make up 58.4\% of the credit-weighted top fifty against 8.2\% under relevance alone, and institutional reports 2.2\% against 1.2\% (Fig.~\ref{fig:year-history}d). On a later snapshot (27 November 2026), the full score replaces 47.9\% of the relevance-only top fifty (sealed-test score: 50.2\%), 64.4\% of its top fifty have a bound author (21.6\% under relevance alone), and collaboration credit sets the author factor for 5.0\% of the top fifty. The shift concentrates in the first lists a reader inspects and fades with the budget, as the bounded factors intend. Over long pools the head also concentrates on prolific researchers: in the thirteen-month common pool of App.~\ref{app:runtime}, 29 of the top thirty papers have a bound author, and two researchers account for eleven of them. Daily pools are far smaller, and the held-out gains of Sec.~\ref{sec:heldout} are measured on each month's top tenth.

\input{experiments/reading_choice_joint_main_v32}

\subsection{Institution profiles across sixteen directions}
\label{sec:institution-profiles}
\begin{figure}[t!]\centering
\includegraphics[width=\linewidth]{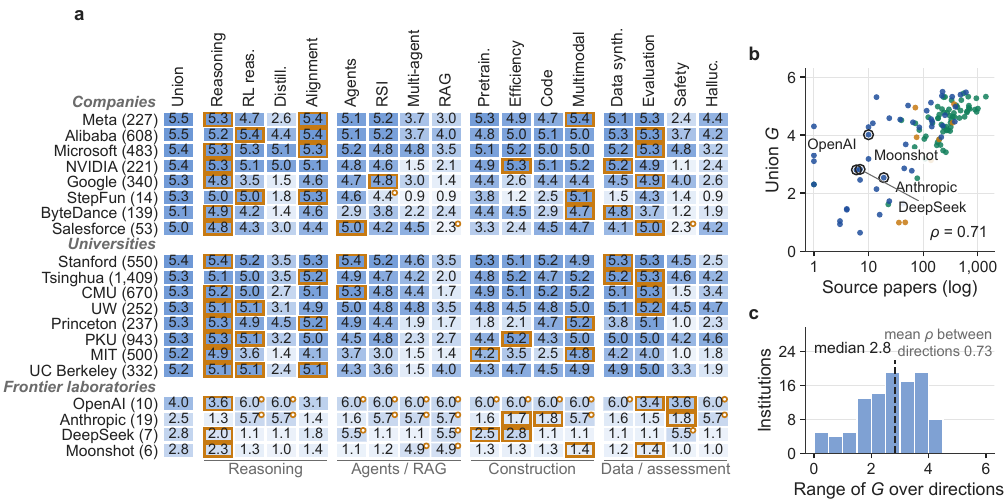}
\caption{\textbf{Institution profiles adapt to the reader's direction, but their level tracks size.} (a) Context $G$ of the institution card on its 0--6 scale for the union of sixteen directions and for each direction, at the 21 November 2026 clock. Rows show the eight companies and eight universities with the highest union context and four frontier laboratories chosen for comparison; parentheses give source papers. Gold outlines mark each row's three strongest directions relative to its own mean, among directions with graph evidence; faded cells with an open circle carry only the archived standing. (b) Union context against source papers for all 114 parents; blue marks companies, green universities and amber research institutes, and circled points are the four frontier laboratories. (c) Spread of $G$ across directions for the 104 institutions with graph evidence in at least four directions.}\label{fig:institutions}
\end{figure}

The institution card of Sec.~\ref{sec:institution-rule} covers 114 parent institutions in seventeen contexts, the sixteen directions and their equal-priority union, computed from 111,519 source works at the 21 November 2026 clock. Fig.~\ref{fig:institutions} shows the eight companies and eight universities with the highest union context, together with four frontier laboratories. The profiles adapt to the reader's question. Among the 104 institutions with graph evidence in at least four directions, the median gap between an institution's strongest and weakest direction is 2.8 points on the 0--6 scale, and the institution orderings of two directions have a mean rank correlation of 0.73, so choosing a direction reorders the list. Google, for example, scores 4.9 in model evaluation and 4.8 in RSI but 1.4 in RAG, whereas Alibaba ranges only from 3.7 to 5.4.

The level of a profile, however, tracks size. Across all 114 parents, union context correlates with source-paper count at Spearman $\rho=0.71$, whereas the archived standing alone does not ($-0.07$). Frontier laboratories that publish few papers therefore sit low: OpenAI, with ten source papers, has a union context of 4.0 and graph evidence in only four directions, and Anthropic, with nineteen, scores 1.3--1.8 wherever its graph evidence suffices. A fifteen-slot panel that a large institution fills easily remains a size advantage, which is why the score has no institution factor (Sec.~\ref{sec:heldout}).

Nor does the level reproduce publication-based rankings. Of the 114 parents, 28 are US universities ranked by CSRankings. Their union context follows their 2024--2026 CSRankings standing in artificial intelligence, machine learning and natural language processing, computed with its scoring rule from its public data~\citep{csrankings2026}, only weakly (Spearman $\rho=0.22$), whereas their paper counts in the library track it more closely ($\rho=0.63$). Within these universities the level tracks how many catalogue researchers fill the panel ($\rho=0.81$): the eight whose panels hold none, including Georgia Tech and Michigan, have a median union context of 4.3, against 4.7 for the other twenty. The profile thus inherits the coverage of the seed catalogue as well as institution size.

\subsection{Author reach under longer evidence histories}

A recent reading window may omit the older collaborations that connect a new paper to an established research history. We quantify this effect by varying only the depth of retained source context, holding fixed the verified identities, 228 catalogue priors, sixteen topic definitions and the 4 November 2026 decay clock. The 30-day, 90-day, 180-day and thirteen-month windows all end on 31 October 2026, and their 9,482, 25,806, 50,143 and 96,271 graph works yield 64 profile--window readouts, all matched on titles and cached abstracts.

From thirty days to thirteen months, the median number of reached non-seed identities rises from \textbf{240.5 to 2,458.5} and the median number of eligible institutions from zero to 91; Fig.~\ref{fig:year-history}(a,b) summarizes every direction, including the unsupported shortest-window institution contexts. Because identity bindings and catalogue priors stay fixed, this growth isolates the coverage that longer source histories supply.

\subsection{Source paths to people and institutions}
\begin{figure}[t!]\centering
\includegraphics[width=\linewidth]{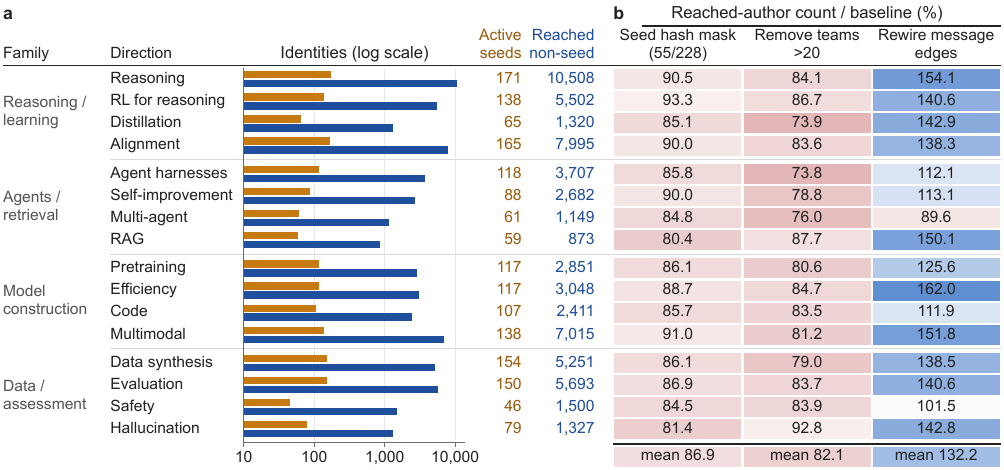}
\caption{\textbf{Direction-level author reach and count sensitivity.} (a) Active catalogue seeds (amber) and reached non-seed identities (blue) on one logarithmic axis starting at ten identities, with both counts printed; rows follow the families of Tab.~\ref{tab:llm-scope}. The source graph contains 111,781 works and 228 catalogue priors. (b) Reached non-seed count under each source control as a percentage of its unperturbed baseline, for all 48 direction--control pairs; shading is red below and blue above 100\%, and the last row gives arithmetic means. Directions can share researchers.}\label{fig:discovery-people-evidence}
\end{figure}

\label{sec:expanded-people}
The discovery graph uses the verified identity anchors and unique catalogue ownership defined in Secs.~\ref{sec:identity} and~\ref{sec:rising}. Each catalogue prior initializes one resolved identity, and supporting papers document every reached connection, so readers can trace how an unfamiliar author or institution enters their profile.

\input{experiments/discovery_graph_readout}

App.~\ref{app:legacy-discovery} retains the publication-separated replay and the fixed-reference versus leave-author-out comparison under their original source populations. These studies describe source reach and reading-list changes.
\begin{figure}[t!]\centering
\includegraphics[width=\linewidth]{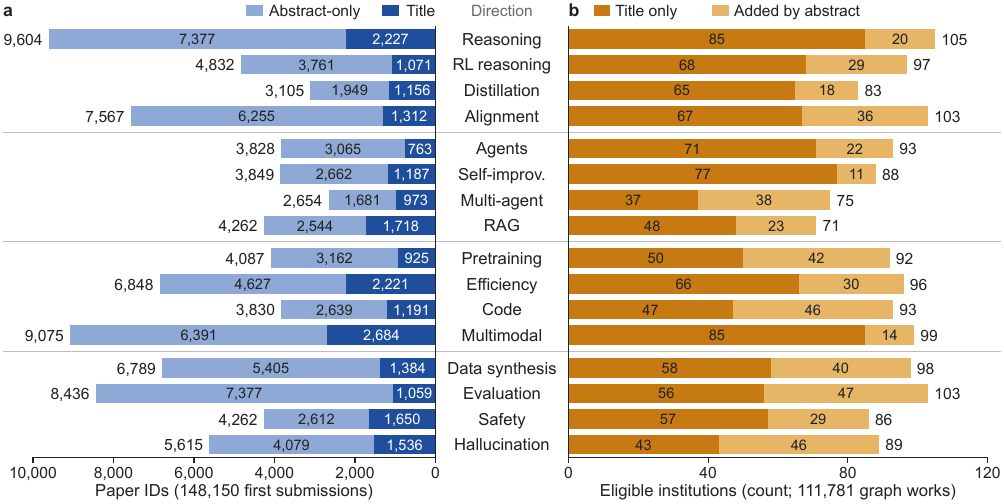}
\caption{\textbf{Paper and institution coverage from title and abstract evidence.} Mirrored bars use independent axes and units. The left side shows title-matched paper identifiers plus additional abstract-only matches among 148,150 first submissions. The right side shows eligible institution--direction contexts: title-only matching yields 980 cells, and cached abstracts add 491, giving 1,471. Numbers in the bars give the two parts and numbers at the bar ends the totals. Central labels retain the same sixteen directions and family order; memberships can overlap.}\label{fig:abstract-evidence}
\end{figure}
\input{experiments/facet_table_v39}
\input{experiments/abstract_main_v30}

\input{experiments/expert_retrieval_figure_v40}
\input{experiments/topic_support_figure_v40}
\input{experiments/data_value_main_v30}

\subsection{Publication volume and ranked-author participation}

Publication volume and researcher participation need not move together. To separate them, each ranked author contributes once per topic and month, weighted by the inverse square root of the archived rank, so repeated papers do not multiply an author's contribution. Paper shares divide by all acquired first submissions that month, and author shares by the total rank weight of distinct seeds in that month's focused cohort, which contains 218 uniquely attributed identities from the fixed AMiner archive; a zero author value means no eligible catalogue binding, even if papers match. Fig.~\ref{fig:topic-dynamics} covers all sixteen directions, and App.~\ref{app:topic-census} extends the analysis to forty.

The learning experiments below use the verified identity anchors. App.~\ref{sec:semantic-author} reports a semantic-history comparison on a snapshot without these anchors.

\input{experiments/learned_main_v32}

%% file: experiments/reading_choice_joint_main_v32.tex
\begin{figure}[t!]\centering
\includegraphics[width=\linewidth]{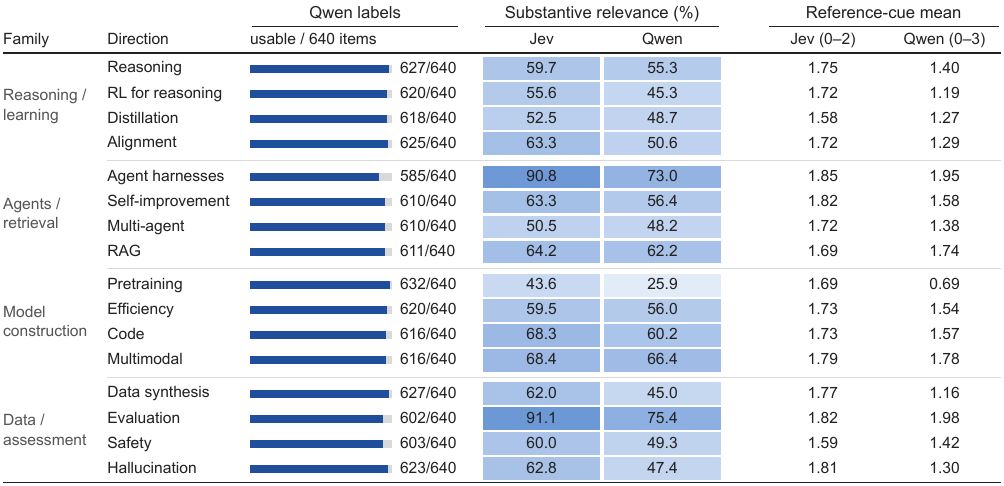}
\caption{\textbf{Assessment coverage and reading evidence in sixteen directions.} Each direction contributes 640 sampled paper--topic items; rows follow the four families of Tab.~\ref{tab:llm-scope}. Bars and counts give usable Qwen judgments, with grey marking unavailable labels. Substantive relevance is the percentage graded two or higher, and shading follows it on a common 0--100\% scale. Reference-cue means use each model's own scale. Jev labels every item; Qwen values are means over usable labels. The ranges compatible with the unavailable labels at the full denominator are at most 8.6 points wide for relevance and 0.26 for reference cues.}\label{fig:reading-pool-coverage-v32}
\end{figure}

\paragraph{Does credit trade relevance for prestige?}
We reuse the 10,240 existing Jev~\citep{typesafe2026jev} assessments of paper--topic items described below; the judge reads titles and abstracts without separate byline, affiliation or score fields, and no new judgments are made. They cover 75.0\% of the credit-weighted top-fifty memberships and 90.9\% of the relevance-only ones. On labelled items, the credit-weighted lists are judged slightly less relevant at fifty reads: the paired mean difference on the 0--3 scale is $-0.052$ (bootstrap 95\% interval $-0.100$ to $-0.003$; positive in 8 of 16 directions), and substantive relevance is 92.0\% against 93.4\%. At 100 reads the difference is $+0.001$ ($-0.025$ to $0.025$). Unlabelled memberships leave wide bounds ($-0.72$ to $+0.30$ at fifty reads). A score variant with a technical-evidence term appeared to raise judged relevance, but that term was derived from this same label pool; without it, credit buys early-citation precision at a small cost in topical focus at the very head of long lists.

\paragraph{The assessment pool.}
The labels come from a study of six additive component arms, which tested whether author and institutional context changes the topical relevance and reusable research cues of papers selected at a fixed reading budget. All arms rank the same eligible papers within each of sixteen directions from the 148,117-paper library. The fixed-penalty baseline combines topic and frontier scores with native recency, solo-author and venue penalties, and these precomputed penalties, including their catalogue-dependent rules, stay identical across arms.

The frozen pool contains 10,240 paper--topic items from 8,849 papers, including all 1,199 top-fifty union items, and Jev~\citep{typesafe2026jev} and Qwen3.8-27B~\citep{qwen38_27b2026} assess the same texts. The 1,967-item primary sample covers every $16\times6\times50$ list membership and the fixed crosscheck, and the complete readout adds 8,273 items while retaining the primary labels. Inputs omit scores, arm labels, bylines and affiliations as separate fields, and Qwen also supplies literal evidence.

Qwen supplies 9,845 usable judgments (96.1\%), with all 395 invalid outputs retained in the denominators, and the two judges agree on the four-level relevance grade for 6,761 of them (68.7\%; quadratic-weighted $\kappa=0.757$). Jev labels 6,500/10,240 items (63.5\%) substantively relevant, against fixed-denominator bounds of 51.8\% to 55.8\% for Qwen; Fig.~\ref{fig:reading-pool-coverage-v32} shows all sixteen directions and the separate cue rubrics.

In that study, the full-context arm changed Jev's top-fifty relevance relative to the fixed-penalty baseline by only \cival{-0.019}{0.038}{0.040} and its reusable-reference score by \cival{0.006}{0.011}{0.011}; substantive relevance was 93.2\% against 93.8\%, and reusable-reference means were 1.887 and 1.881. The original 1,967-item Qwen sample retains 64 schema/rubric violations and 32 literal-evidence failures, including 66 unavailable mandatory items, and both primary Qwen contextual contrasts cross zero under shared-item missingness bounds. These close readouts show that the added author paths accompanied broadly similar topical content, with uncertainty reported for each paired contrast; App.~\ref{app:reading-choice-joint-v32} reports all arms, directions, the author-reference control and the uncertainty definitions.

%% file: experiments/discovery_graph_readout.tex
The full source graph contains 111,781 works and 114 parent-grouped institutional contexts, including 14,988 works from before October 2025 and 522 from after 31 October 2026, beyond the complete-month acquisition cohort. Across the sixteen primary directions, 46--171 catalogue seeds are active, 873--10,508 non-seed identities receive support (Fig.~\ref{fig:discovery-people-evidence}a) and 71--105 parent institutions are eligible, and every reached identity keeps the papers that supply its connections.

Source controls that retain source affiliations move reach in both directions (Fig.~\ref{fig:discovery-people-evidence}b): the fixed seed mask keeps a mean 86.9\% of the original reached-author count and excluding large teams keeps 82.1\%, whereas degree-preserving rewiring reaches 132.2\% on average, so greater reach alone need not identify more informative paths.

%% file: experiments/facet_table_v39.tex
\begin{table}[tbp]\centering\small
\caption{\textbf{Cached abstracts support an external relevance test.} The CSFCube expert relevance benchmark supplies 50 query--facet tasks, 34 distinct query papers and 6,242 non-self expert-judged pairs. BM25 is the lexical baseline; SPECTER2, a scholarly-document encoder, and the Qwen3-Embedding encoders use the same full title--abstract input. Entries are NDCG@10 $\times100$; the last column of panel (a) uses the mean $\pm$ the larger 95\% bootstrap half-width. Panel (a) reports frozen representations by relevance facet; panel (b) reports the fixed model readouts under each input condition, with the 8B cosine baseline. The unrun 8B title-only condition is marked by a dash. No benchmark grade selects a configuration.}\label{tab:facet-retrieval}
\noindent\textit{a\quad Frozen representations and facet relevance}\par\smallskip
\setlength{\tabcolsep}{5pt}\renewcommand{\arraystretch}{1.16}
\begin{tabularx}{\linewidth}{Yrrrr}\toprule
\rowcolor{tablehead}\textcolor{white}{\textbf{Representation / input}}&\textcolor{white}{\textbf{Background}}&\textcolor{white}{\textbf{Method}}&\textcolor{white}{\textbf{Result}}&\textcolor{white}{\textbf{All facets}}\\\midrule
\rowcolor{tablepale}
BM25: title&30.88&24.40&24.84&\cival{26.62}{6.44}{6.48}\\
BM25: title + abstract&58.21&32.14&42.69&\cival{44.07}{6.65}{6.55}\\
\rowcolor{tablepale}
SPECTER2: full abstract&61.19&40.26&54.43&\cival{51.78}{6.76}{6.65}\\
Qwen3-Embedding-4B&\textbf{64.89}&\textbf{46.24}&57.20&\cival{\mathbf{55.94}}{6.55}{6.66}\\
\rowcolor{tablepale}
Qwen3-Embedding-8B&63.84&43.56&\textbf{59.21}&\cival{55.37}{6.75}{6.83}\\
\bottomrule\end{tabularx}
\par\medskip\noindent\textit{b\quad Input granularity and fixed relevance readouts}
\par\medskip
{\small\setlength{\tabcolsep}{5pt}\renewcommand{\arraystretch}{1.08}
\begin{tabularx}{\linewidth}{Yrrr}\toprule
\rowcolor{tablehead}\textcolor{white}{\textbf{Input view}}&\textcolor{white}{\textbf{Qwen3.8-27B}}&\textcolor{white}{\textbf{Qwen3-32B}}&\textcolor{white}{\textbf{8B cosine}}\\\midrule
Title only&41.22&41.13&---\\
\rowcolor{tablepale}Title + abstract&59.80&\textbf{52.97}&55.37\\
Query facet + full candidate&\textbf{61.51}&52.25&\textbf{57.07}\\\bottomrule
\end{tabularx}}
\end{table}

%% file: experiments/abstract_main_v30.tex
\subsection{Matching evidence from cached abstracts}
\label{sec:abstract-evidence}

To isolate the information carried by abstracts, we apply the same sixteen direction definitions to two views of the thirteen-month corpus. Titles admit 20,599 distinct papers; titles plus abstracts admit 58,094. Thus \textbf{37,495 papers, or 64.5\% of the focused cohort, have no matching title signal} (App.~\ref{app:semantic}, Fig.~\ref{fig:abstract-profiles}). Fig.~\ref{fig:abstract-evidence} includes every direction and uses text already retained in the library.

The input change also broadens institutional coverage under the same core-contributor and parent-group rule. Eligible institution--direction combinations rise from \textbf{980 to 1,471}, with 491 additions and no losses (Tab.~\ref{tab:abstract-detail}); union eligibility rises from 107 to 109. The paired comparison keeps identities, catalogue priors, source attributions and scoring definitions fixed.

For author profiles, each paper divides one unit across its matching directions. With abstracts, 24,490 bound authors have at least three matching papers; 16,994 fall below that threshold with titles alone. Among the 7,496 authors eligible in both views, the median direction count rises from three to seven, and 30.6\% share no highest-weight direction, allowing joint maxima. The added matches change both profile eligibility and the distribution of topic evidence within an author's stored publications. Background mentions can contribute, which motivates the following test of the \emph{aspect} a reader seeks.

\input{experiments/facet_retrieval_v28}

%% file: experiments/facet_retrieval_v28.tex
\subsection{Does richer text improve facet-specific retrieval?}
\label{sec:facet-retrieval}

To evaluate the text substrate that a downstream reader or agent can reuse, we turn to CSFCube's independently produced judgments of background, method and result relevance~\citep{mysore2021csfcube}. With the candidate pools and lexical settings fixed, the comparison varies title-only BM25~\citep{robertson2009bm25}, title--abstract BM25 and three frozen semantic encoders, SPECTER2-base~\citep{singh2023scirepeval} and Qwen3-Embedding-4B/8B~\citep{zhang2025qwenembedding}, whose representations and ranking rules were specified before these results were examined.

Adding abstracts raises lexical NDCG@10 from \textbf{26.62 to 44.07}, and with the same title--abstract input SPECTER2, Qwen4B and Qwen8B reach \textbf{51.78, 55.94 and 55.37}, so the larger Qwen backbone has a similar full-abstract mean here. When both sides use their benchmark facet labels, a condition that evaluates retrieval with the supplied reading facet, Qwen8B reaches 58.78. App.~\ref{app:semantic} retains every fixed lexical and semantic configuration, their paired differences and the original self-match correction.

%% file: experiments/expert_retrieval_figure_v40.tex
\begin{figure}[t!]\centering
\includegraphics[width=\linewidth]{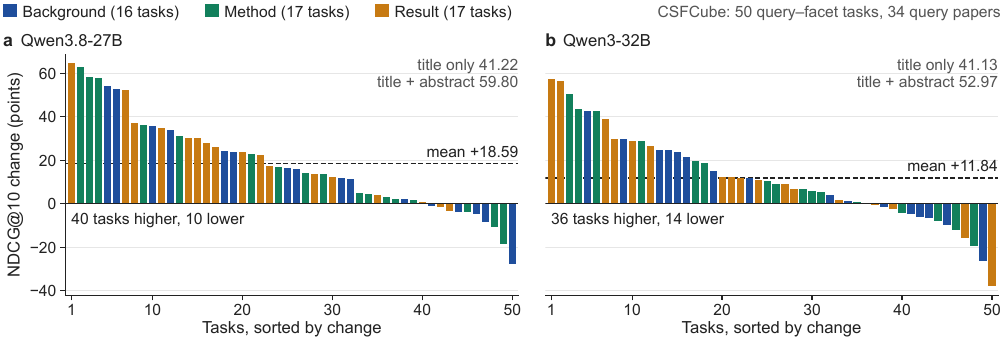}

\caption{\textbf{Abstracts support finer relevance distinctions across expert reading tasks.} Change in NDCG@10 ($\times100$) from adding abstracts (title plus abstract minus title only, paired within each task) for the two fixed scorers, (a) Qwen3.8-27B and (b) Qwen3-32B. Bars are the fifty tasks, sorted by change and coloured by facet. The dashed line gives the mean change, whose conditional 95\% interval over 34 query-document clusters is in the text; grey text gives the mean NDCG@10 ($\times100$) of each input. All recorded input conditions, including the full-input means of both scorers and of the frozen Qwen3-Embedding-8B cosine reference, appear separately in Tab.~\ref{tab:facet-retrieval}(b); the 8B title-only condition was not run.}\label{fig:expert-readouts-v30}
\end{figure}

%% file: experiments/topic_support_figure_v40.tex
\begin{figure}[t!]
\centering
\includegraphics[width=\linewidth]{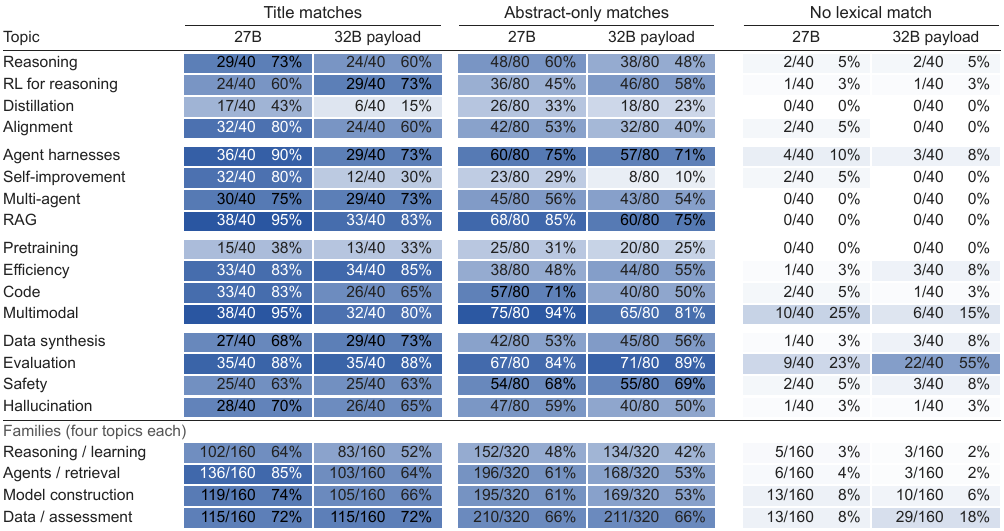}
\caption{\textbf{Source-quoted topic judgments at two levels of granularity.} Adjacent columns report Qwen3.8-27B and Qwen3-32B counts with valid literal quotes. Each topic retains 40 title matches, 80 abstract-only matches and 40 no-lexical-match pairs. Topic rows are grouped by the four families of Tab.~\ref{tab:llm-scope}, and the lower band pools the same rows by family, with denominators of 160, 320 and 160. Each cell prints the count and its percentage of all sampled rows, and shading follows that percentage on a 0--100\% scale. The 32B columns use the extracted JSON payload; complete response and validation records accompany the analysis.}
\label{fig:topic-support-v30}
\end{figure}

%% file: experiments/data_value_main_v30.tex
\begin{figure}[t!]\centering
\includegraphics[width=\linewidth]{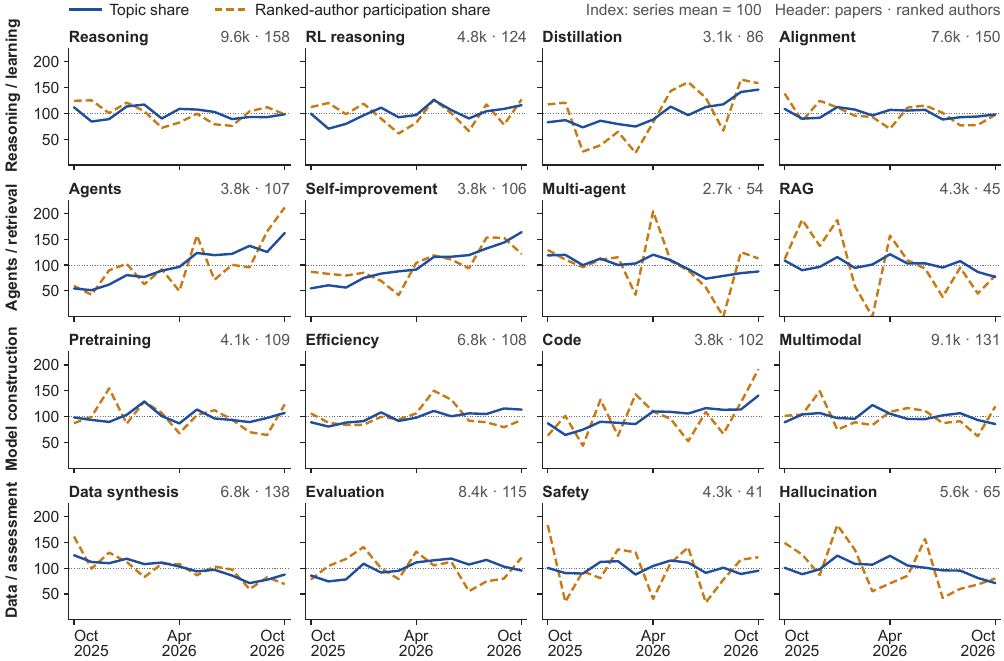}
\caption{\textbf{Volume and ranked-author participation need not move together.} Monthly topic share (solid blue) and ranked-author participation share (dashed amber) are each indexed to their own thirteen-month mean of 100; rows group the four direction families. This supports within-series trend comparison without equating the denominators. Each uniquely attributed catalogue author contributes once per topic/month, using verified identity anchors and archived ranks; the two numbers beside each direction give its distinct papers and ranked authors over the complete period.}\label{fig:topic-dynamics}
\end{figure}

\subsection{Expert relevance and source-supported topic readouts}
\label{sec:data-value-v30}
Broader topic coverage helps only if the added text also distinguishes relevant methods and results, so we test this link with a fixed relevance scorer on the same CSFCube candidate pools. Qwen3.8-27B~\citep{qwen38_27b2026} scores title-only, full title--abstract and supplied query-facet inputs for the same 50 tasks and 6,242 non-self expert-judged pairs, and matching frozen Qwen3-Embedding-8B baselines cover the full and query-facet conditions. The model and prompts were fixed before scoring, and expert grades were withheld.

Fig.~\ref{fig:expert-readouts-v30} and Tab.~\ref{tab:facet-retrieval}(b) separate input content from readout choice. Adding abstracts raises Qwen3.8-27B NDCG@10 from \textbf{0.412 to 0.598}, a paired change of \cival{18.59}{6.60}{7.76} points (conditional 95\% interval), and query-facet input reaches \textbf{0.615} against 0.571 for the matching 8B embedding baseline, a difference of \cival{4.45}{4.12}{4.62} points. App.~\ref{app:data-value-v30} retains every condition, deeper-list metric and interval.

% BEGIN QWEN32 MATCHED COMPARATOR
A fixed Qwen3-32B comparator~\citep{qwen3_32b2025}, given identical 21,286 input rows, prompts, caps and decoding settings, reaches an expert NDCG@10 of 0.411 for titles, 0.530 for titles plus abstracts and 0.522 for query-facet input. Abstracts again add \cival{11.84}{5.86}{6.54} points, and its query-facet result is \cival{-9.26}{3.25}{3.36} points relative to Qwen3.8-27B.
% END QWEN32 MATCHED COMPARATOR

A sample from the local corpus then examines the roles topics play in abstracts. Each of sixteen topics contributes 40 title matches, 80 abstract-only matches and 40 papers without a lexical match; Qwen3.8-27B sees only the abstract and the topic definition, with the stratum withheld, and a primary or secondary relation requires an exact supporting substring.

All 2,560 Qwen3.8-27B outputs pass strict JSON and schema checks, and 1,262 of 1,267 primary/secondary judgments carry a literal source quote. These quote-supported relations cover \textbf{472/640} title matches, \textbf{753/1,280} abstract-only matches and \textbf{37/640} papers without a lexical match (Fig.~\ref{fig:topic-support-v30}). Among abstract-only matches they range from 23/80 for self-improvement to 75/80 for multimodal fusion, so the same input expansion exposes very different amounts of topic evidence across directions.

% BEGIN QWEN32 TOPIC AGREEMENT
In the Qwen3-32B payload analysis, literal source support covers 1,133 of 1,230 primary/secondary choices; the extracted relation labels agree with Qwen3.8-27B on 1,918 of 2,560 rows (74.9\%; Cohen's $\kappa=0.616$), and 998 rows are supported by both models. Fig.~\ref{fig:topic-support-v30} reports the same stratified counts for each model, and App.~\ref{app:data-value-comparator-v30} describes the payload extraction.

\input{sections/radar_main_v31}

%% file: sections/radar_main_v31.tex
The family totals beneath the topic rows in Fig.~\ref{fig:topic-support-v30} summarize the same judgments at a coarser research granularity, and keeping the three strata separate shows whether abstract evidence concentrates in particular directions without confounding the comparison with their different sampling allocations.

%% file: experiments/learned_main_v32.tex
\subsection{Learning from the acquired author graph}
\label{sec:learned-graph}
\input{experiments/learned_table_v32}

To test whether the acquired histories support learning about research continuity, we reconstruct held-out authorship relations: each author's remaining source-bound papers provide the context for ranking the withheld ones. The graph contains 104,908 author nodes, 146,393 training author--paper relations and 935,262 directed message edges, and every method ranks the same 5,797 test candidates for 15,694 query authors (App.~\ref{app:learned-graph} specifies the paper-level splits and validation-only selection).

Frozen histories, matched neural projections, one- and two-layer messages, learned gates and degree-preserving rewiring separate learned adaptation, semantic preservation and graph structure; residual readouts retain 90\% of the frozen semantic cosine and learn the remaining 10\%. With the 8B backbone (Tab.~\ref{tab:learned-static}), the validation-selected residual graph reaches a mean reciprocal rank (MRR, $\times100$) of \pmval{17.80}{0.07} and \pmval{41.84}{0.12}\% recall at fifty, against 15.89 and 36.56\% for frozen abstracts, while the matched residual multilayer perceptron (MLP) reaches \pmval{17.82}{0.15} MRR and \pmval{41.46}{0.24}\% recall. The graph thus adds 0.38 recall points over the MLP while lowering MRR by 0.014 points, so most of the gain over frozen histories comes from semantic preservation and adaptation rather than graph structure.

\textbf{Semantic retention and optional graph refinement.} With the same frozen 8B encoder, title--abstract histories reach 15.89 MRR against 10.06 for titles, a 58\% relative increase before any task-specific training, whereas comparing the matched residual MLP with the graph model reveals a much smaller contribution from graph structure. This pattern motivates retaining complete abstract histories for downstream models while keeping ordinary discovery independent of neural inference.

\begin{figure}[t!]\centering
\includegraphics[width=\linewidth]{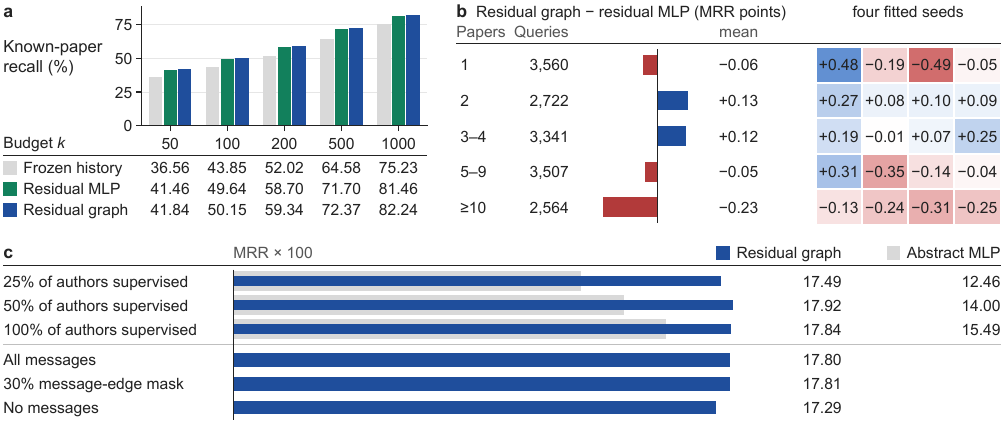}
\caption{\textbf{Learning and structural controls on a reusable author graph.} All panels use the same 15,694 held-out author queries and 5,797 candidates with frozen Qwen3-Embedding-8B features. (a) Recall from the same frozen checkpoints and features at $k=50,100,200,500,1000$; the values below the bars are means over four fitted seeds (sample SD at most 0.24 points). (b) Graph-minus-residual-MLP MRR by history size in papers; query counts are shown for every stratum. Bars and values give the mean over the fitted seeds, and the shaded cells give the individual seeds. (c) MRR on one zero-based scale, each model's values in its own column. Upper rows: the abstract MLP and residual graph at three supervision fractions, using one fixed initialization and the full context graph. Lower rows: the same fitted graph models under all messages, a 30\% message-edge mask and no messages, as means over the four fitted seeds (sample SD at most 0.08). The mask removes message-edge endpoint units, masks symmetric pairs together and renormalizes rows. Queries can share papers, and targets are observed authorship relations.}\label{fig:learned-mechanisms}
\end{figure}

At a budget of 1,000 candidates (Fig.~\ref{fig:learned-mechanisms}a), additional readouts of the saved checkpoints reach \pmval{82.24}{0.22}\% recall for the residual graph, \pmval{81.46}{0.09}\% for the matched residual MLP and 75.23\% for frozen abstracts; the shared Recall@50/100 points reproduce their original per-query values, and the extended budgets reuse the same features, splits and selected parameters. With text and fitted parameters fixed, removing all messages reduces 8B MRR to 17.29, whereas masking 30\% of message edges retains 17.81 (Fig.~\ref{fig:learned-mechanisms}c). At 25\%, 50\% and 100\% supervised-author coverage, the single-seed graph obtains 17.49, 17.92 and 17.84 MRR (Fig.~\ref{fig:learned-mechanisms}c), a non-monotone curve that distinguishes source dependence from a uniform benefit of more supervision. Chronological, target-team and dependent-observation analyses appear in App.~\ref{app:learned-graph}.

%% file: experiments/learned_table_v32.tex
\begin{table}[tbp]\centering\small
\caption{\textbf{Three semantic backbones on the same held-out relations.} Graph neural network (GNN) and multilayer perceptron (MLP) readouts use the same 15,694 queries and 5,797 candidates. Values are mean reciprocal rank (MRR) and recall@50 multiplied by 100; grey $\pm$ values are sample SD over four initializations. Bold marks the largest test mean in each column; $\dagger$ identifies the family selected by validation MRR before test access. Frozen rows have no fitted-seed dispersion. Residual models retain 90\% of the original semantic cosine and learn the remaining 10\%.}\label{tab:learned-static}
\setlength{\tabcolsep}{3.3pt}\renewcommand{\arraystretch}{1.13}
\begin{tabularx}{\linewidth}{Yrrrrrr}\toprule
\rowcolor{tablehead}\textcolor{white}{\textbf{Readout}}&\multicolumn{2}{c}{\textcolor{white}{\textbf{SPECTER2}}}&\multicolumn{2}{c}{\textcolor{white}{\textbf{Qwen3-4B}}}&\multicolumn{2}{c}{\textcolor{white}{\textbf{Qwen3-8B}}}\\
\rowcolor{tablehead}&\textcolor{white}{MRR}&\textcolor{white}{R@50}&\textcolor{white}{MRR}&\textcolor{white}{R@50}&\textcolor{white}{MRR}&\textcolor{white}{R@50}\\\midrule
\rowcolor{tablepale}
Frozen title&\nopm{7.14}&\nopm{20.49}&\nopm{10.03}&\nopm{27.12}&\nopm{10.06}&\nopm{27.73}\\
Frozen abstract&\nopm{12.17}&\nopm{31.28}&\nopm{15.46}&\nopm{35.77}&\nopm{15.89}&\nopm{36.56}\\
\midrule
\rowcolor{tablepale}
Title MLP&\pmval{7.92}{0.03}&\pmval{25.88}{0.20}&\pmval{9.41}{0.10}&\pmval{29.08}{0.45}&\pmval{9.50}{0.11}&\pmval{29.83}{0.03}\\
Abstract MLP&\pmval{13.72}{0.08}&\pmval{37.35}{0.17}&\pmval{15.35}{0.13}&\pmval{39.21}{0.33}&\pmval{15.35}{0.21}&\pmval{39.17}{0.43}\\
\rowcolor{tablepale}
Mean GNN (1)&\pmval{13.81}{0.11}\rlap{$^{\dagger}$}&\pmval{38.12}{0.14}\rlap{$^{\dagger}$}&\pmval{15.59}{0.12}&\pmval{40.48}{0.36}&\pmval{15.70}{0.30}&\pmval{40.58}{0.12}\\
Mean GNN (2)&\pmval{13.84}{0.19}&\pmval{38.37}{0.23}&\pmval{15.29}{0.22}&\pmval{40.80}{0.24}&\pmval{15.52}{0.19}&\pmval{40.84}{0.51}\\
\rowcolor{tablepale}
Gated GNN (1)&\pmval{13.81}{0.11}&\pmval{38.10}{0.18}&\pmval{15.51}{0.16}&\pmval{40.20}{0.20}&\pmval{15.69}{0.30}&\pmval{40.61}{0.23}\\
Gated GNN (2)&\pmval{13.60}{0.19}&\pmval{\mathbf{38.51}}{0.14}&\pmval{15.43}{0.11}&\pmval{40.91}{0.24}&\pmval{15.85}{0.22}&\pmval{40.89}{0.17}\\
\rowcolor{tablepale}
Rewired GNN (2)&\pmval{\mathbf{13.87}}{0.17}&\pmval{37.66}{0.14}&\pmval{15.18}{0.22}&\pmval{39.23}{0.29}&\pmval{15.45}{0.16}&\pmval{39.49}{0.69}\\
Residual MLP&\pmval{12.03}{0.04}&\pmval{30.91}{0.12}&\pmval{16.77}{0.51}&\pmval{39.59}{1.67}&\pmval{\mathbf{17.82}}{0.15}&\pmval{41.46}{0.24}\\
\rowcolor{tablepale}
Residual GNN (2)&\pmval{12.08}{0.05}&\pmval{30.97}{0.07}&\pmval{\mathbf{17.29}}{0.05}\rlap{$^{\dagger}$}&\pmval{\mathbf{41.24}}{0.08}\rlap{$^{\dagger}$}&\pmval{17.80}{0.07}\rlap{$^{\dagger}$}&\pmval{\mathbf{41.84}}{0.12}\rlap{$^{\dagger}$}\\
\bottomrule\end{tabularx}\end{table}

%% file: experiments/planning_handoff_v42.tex
\begin{table}[!t]\centering\small
\caption{\textbf{Source-linked plans at two reading budgets.} Sixteen directions per condition; 64 plans and 32 paired model assessments. (a) Source checks and model scores (0--4) for the original fixed packets. (b) Post-hoc proportions of distinct cited papers per schema-valid plan with each recorded source indicator; percentages are mean $\pm$ sample SD across usable plans. History requires a resolved author with ten earlier dated works; institution context requires explicit attribution to an eligible parent with a mean-based reference contribution $G_{\mathrm{mean}}\geq3$ in that direction; two-hop paths use saved contributions for source-attributed non-seed authors. ``Any'' is their union. Bold marks the larger observed mean within a budget.}\label{tab:planner-v42}
\setlength{\tabcolsep}{4pt}\renewcommand{\arraystretch}{1.18}
\noindent\textcolor{primary}{a\quad Source checks and plan assessments}\par\smallskip
\begin{tabularx}{\linewidth}{>{\raggedleft\arraybackslash}p{.04\linewidth}>{\raggedright\arraybackslash}p{.22\linewidth}>{\raggedleft\arraybackslash}p{.10\linewidth}*{4}{>{\raggedleft\arraybackslash}X}}\toprule
\rowcolor{tablehead}\textcolor{white}{\textbf{$k$}}&\textcolor{white}{\textbf{Evidence selection}}&\textcolor{white}{\textbf{Schema}}&\textcolor{white}{\makecell[r]{Valid\\IDs (\%)}}&\textcolor{white}{\makecell[r]{Quotes\\(\%)}}&\textcolor{white}{\textbf{Specificity}}&\textcolor{white}{\textbf{Feasibility}}\\\midrule
\rowcolor{tablepale}20&Full context&14/16&100.0&94.8&\pmval{\mathbf{3.15}}{0.19}&\pmval{\mathbf{2.44}}{0.24}\\
20&Lexical + recency&13/16&100.0&98.5&\pmval{3.13}{0.23}&\pmval{2.42}{0.32}\\\midrule
\rowcolor{tablepale}50&Full context&14/16&100.0&95.6&\pmval{\mathbf{3.11}}{0.21}&\pmval{2.43}{0.23}\\
50&Lexical + recency&15/16&100.0&97.4&\pmval{3.04}{0.27}&\pmval{\mathbf{2.46}}{0.28}\\\bottomrule
\end{tabularx}
\par\smallskip\noindent\textcolor{primary}{b\quad Research context of cited papers (\%)}\par\smallskip
\input{analysis/institution_profile_v48/postprocess/TABLE9B}
\end{table}

%% file: analysis/institution_profile_v48/postprocess/TABLE9B.tex
\begin{tabularx}{\linewidth}{>{\raggedleft\arraybackslash}p{.04\linewidth}>{\raggedright\arraybackslash}p{.22\linewidth}>{\raggedleft\arraybackslash}p{.10\linewidth}*{4}{>{\raggedleft\arraybackslash}X}}\toprule
\rowcolor{tablehead}\textcolor{white}{\textbf{$k$}}&\textcolor{white}{\textbf{Evidence selection}}&\textcolor{white}{\makecell[r]{Usable\\plans}}&\textcolor{white}{\makecell[r]{Prior author\\history}}&\textcolor{white}{\makecell[r]{Institution\\context}}&\textcolor{white}{\makecell[r]{Two-hop\\path}}&\textcolor{white}{\makecell[r]{Any source\\context}}\\\midrule
\rowcolor{tablepale}20&Full context&14/16&\pmval{\mathbf{83.3}}{32.2}&\pmval{\mathbf{56.5}}{44.1}&\pmval{\mathbf{66.7}}{39.4}&\pmval{\mathbf{86.9}}{29.4}\\
20&Lexical + recency&13/16&\pmval{48.5}{39.7}&\pmval{13.1}{29.8}&\pmval{34.6}{38.2}&\pmval{53.8}{43.6}\\\midrule
\rowcolor{tablepale}50&Full context&14/16&\pmval{\mathbf{75.0}}{34.6}&\pmval{\mathbf{58.3}}{32.4}&\pmval{\mathbf{56.0}}{40.7}&\pmval{\mathbf{79.2}}{31.6}\\
50&Lexical + recency&15/16&\pmval{56.4}{40.6}&\pmval{6.7}{25.8}&\pmval{28.0}{36.5}&\pmval{56.4}{40.6}\\
\bottomrule\end{tabularx}

%% file: literature/landscape_readout_v04.tex
Within the selected set, research systems and benchmarks each contribute 18 papers, while retrieval and ideation contribute 10 each.
All eight harness/improvement papers fall in August--October 2026, whereas the execution and benchmark samples extend across earlier months of the observation window.
\knowme{} is positioned at the retrieval entry point, supplying inspectable candidate evidence for subsequent research activities; the outlined row marks this conceptual role rather than adding another paper to the sample.

%% file: literature/categorized_citations.tex
% Optional insertion into Related Work. All 75 new references below are task-screened.
% Scope: recent LLM research first submitted from 2025-11-01 to 2026-11-01.
% This taxonomy describes reported mechanisms and evaluation objects; it does not
% compare incompatible headline scores or claim that KnowMe already improves them.

\paragraph{From research assistance to autonomous workflows.}
Recent systems organize research around persistent workspaces, baseline papers,
and recoverable artifacts, rather than a single text-generation call
\citep{km260900365,km251104583,km260503042,km260531468}.
Execution-grounded search and resource-aware state management connect proposals
to experiments in MARS, execution-grounded AI research, ScienceFlow, and EurekAgent
\citep{km260202660,km260114525,km260814354,km260613662}.
Other systems coordinate exploration, experimentation, reporting, and human
intervention through multiple agents
\citep{km260702807,km260308127,km261018933,km260520025,km260813558}.
Industrial research, learned outcome prediction, and harness efficiency introduce
additional constraints on how much exploration a fixed budget can support
\citep{km260910922,km260812564,km261012235,km260920519}.
A recent survey frames this transition as workflow-level research automation
\citep{km260523204}. These systems are prospective consumers of the literature selected by \knowme{}.

\Needspace{10\baselineskip}
\paragraph{Evaluating research requires more than successful execution.}
ResearchGym, AIRS-Bench, ResearchClawBench, and AARRI-Bench instantiate different
starting conditions, hidden information, and research behaviors
\citep{km260215112,km260206855,km260607591,km260607462}.
InnoGym and evaluations of creativity distinguish novelty from usefulness, while
DataSense-Bench checks whether data-selection judgments predict downstream
training outcomes \citep{km251201822,km260830047,km261012190}.
RSIBench-Data, MMPostTrainBench, and AI4AI-Bench constrain the experimental
substrate to inspect data strategies, model regressions, or training-algorithm
changes \citep{km260725886,km261005398,km260820318}.
Process traces, lifecycle failure taxonomies, pivotal research decisions, and
experimental-effect prediction expose failures that a final score can conceal
\citep{km260909203,km260814905,km261021082,km260927490}.
AutoSciRub makes task-specific scientific requirements executable before revision
\citep{wang2026autoscirub}. Evidence-grounded answer evaluation,
TruthInsightBench, and SDABench further distinguish following supplied evidence,
establishing an agent-originated claim, and exercising specific scientific
capabilities \citep{km261004915,km260905079,km260711079}.
These distinctions motivate separate endpoints for retrieval, faithful use of
evidence, and independently verified experimental progress.

\paragraph{Judging and generating research ideas.}
Novelty assessment can be decomposed into evidential dimensions and historical
research context \citep{km260911234,km260320884}.
SoundnessBench and IdeaAMBIG examine whether an idea is viable and sufficiently
specified for implementation \citep{km260530329,km260910539}.
Idea-level critics, pre-execution prediction, and alternative hypothesis-scoring
methods study the cost of deciding what to verify
\citep{km261008993,km260105930,km260817270}.
Quality-diversity search and graph-derived falsifiable ideas offer complementary
ways to structure the candidate space \citep{km260722375,km260820361}.
At the same time, style sensitivity in LLM judging can confound an apparent
improvement in idea quality \citep{km260801666}.
A \knowme{} evaluation should therefore preserve the distinction between a
paper's topical fit, a proposed use, a judge's preference, and the outcome of
actually pursuing that use.

\paragraph{Personalized and relation-aware scholarly retrieval.}
Personalized Deep Research, Sci-Surf, and MyScholarQA place user requirements and
feedback at different points in the discovery workflow
\citep{km260510530,km260811973,km260316120}.
Structured graph exploration, RATIO, and SciNet examine bounded navigation,
operation-specific retrieval, and relations among scientific works
\citep{km260824809,km260827394,km260103260}.
FootprintRAG and ReCite make the connection between retrieved material, consumed
evidence, and claim support explicit \citep{km260919601,km260909156}.
Recursive literature retrieval studies how the retrieval process itself may
adapt \citep{km260514306}; work on literature-search evaluation cautions that
human citation lists are an incomplete target for judging retrieval
\citep{km260529234}.
Together, these works provide closer comparisons for \knowme{} than generic
question answering: user-specific usefulness, relationship-sensitive discovery,
source attribution, and the cost of reaching an adequate evidence set.

\paragraph{Evidence reuse and recursive harness improvement.}
RRSI and ModularRSI investigate constraints and evaluation separation intended to
favor reusable harness changes \citep{km260924972,km260914857}.
AutoSaddler treats the harness as an editable object selected through validation,
while ActiveSaddler adapts the scenarios used to produce optimization feedback
\citep{km260823041,km261000906}.
Mixture-of-branches optimization diversifies proposal policies and development
subsets \citep{km260937834}. GenMem explores stable symbolic memory addressing
while stored payloads evolve \citep{km260934633}; DataFoundry applies recursive
revision to data-preparation procedures \citep{km260829966}.
AIDE$^2$ makes an AI research agent's own code the object of the next improvement
round \citep{km260926457}.
These mechanisms clarify what an outcome-trained extension of \knowme{} would
need to specify: the editable component, source of feedback, selection budget,
and evaluation tasks withheld from adaptation.

\paragraph{Knowledge graphs, reusable skills, and data curation.}
Mechanist combines scientific knowledge resources, method skills, and agent
orchestration to connect hypotheses with execution
\citep{wang2026mechanist}.
DataCOPE discovers reusable data-analysis skills using indirect verifier signals,
then evaluates them on held-out tasks \citep{qiu2026datacope}.
Autonomous agentic data engineering instead treats training-data strategy as a
research decision whose outcome is measured through model specialization
\citep{km260530407}.
These distinct paths from literature or experience to action motivate
attributable evidence packets and explicit downstream tests of \knowme{}'s
author, institution, and preference signals.

\paragraph{Validity, exploration breadth, and total cost.}
Research-agent reward hacking, experimental-fidelity audits, and scientific-slop
benchmarks identify ways in which a plausible report or a high metric can fail
to support its scientific claim
\citep{km260928614,km260826753,km261000531}.
Distributional analyses of generated ideas raise a separate concern about
concentration near seed literature \citep{km260527905}.
REFLECT evaluates judges through controlled errors in research-agent traces,
rather than assuming their scores are reliable feedback
\citep{km260519196}.
Studies of wasted computation, pre-execution idea selection, and localized
harness gains show why search, verification, and budget allocation must be
accounted for separately \citep{km260810424,km260727687,km260902889}.
Accordingly, the relevant objective for \knowme{} is useful, independently
supported progress per total budget, reported alongside omissions and exposure
concentration; metadata scores and automated judgments remain intermediate
signals.

\paragraph{Academic-network mining and historical author signals.}
AMiner integrates researcher-profile extraction, name disambiguation, joint
modeling of papers, authors and venues, and topic-level expertise search, alongside
relationship mining and community evolution \citep{wan2019aminer}.
This is direct prior art for author-centered scholarly navigation, independent of
the recent wave of research agents. AI2000 is separately attributed as an AMiner
ranking resource \citep{aminerai2000}. \knowme{} uses documented historical
candidate-list snapshots as bounded priors; it does not claim to reproduce the
current website ranking or to turn author visibility into a certificate of
paper quality.

%% file: experiments/hardware_comparison_table.tex
\begin{tabularx}{\linewidth}{>{\raggedright\arraybackslash}p{.20\linewidth}*{4}{>{\raggedleft\arraybackslash}X}}
\toprule
\rowcolor{tablehead}\textcolor{white}{\textbf{Host}} & \textcolor{white}{\textbf{Papers}} & \textcolor{white}{\textbf{\makecell[r]{Base\\(s)}}} & \textcolor{white}{\textbf{\makecell[r]{Personal\\(s)}}} & \textcolor{white}{\textbf{\makecell[r]{Personal RSS\\(GiB)}}}\\\midrule
24 GiB / 14 cores & 10,000 & \pmval{1.72}{0.55} & \pmval{1.52}{0.41} & \pmval{\mathbf{1.02}}{0.22}\\
24 GiB / 14 cores & 50,000 & \pmval{10.76}{0.29} & \pmval{12.08}{0.61} & \pmval{\mathbf{0.91}}{0.08}\\
24 GiB / 14 cores & 100,000 & \pmval{20.68}{0.50} & \pmval{22.71}{1.31} & \pmval{\mathbf{1.02}}{0.09}\\
\rowcolor{knowlight}48 GiB / 12 cores & 10,000 & \pmval{\mathbf{1.63}}{0.58} & \pmval{\mathbf{1.39}}{0.38} & \pmval{1.45}{0.32}\\
\rowcolor{knowlight}48 GiB / 12 cores & 50,000 & \pmval{\mathbf{9.82}}{0.44} & \pmval{\mathbf{11.12}}{0.42} & \pmval{1.25}{0.32}\\
\rowcolor{knowlight}48 GiB / 12 cores & 100,000 & \pmval{\mathbf{20.64}}{0.68} & \pmval{\mathbf{21.68}}{0.11} & \pmval{1.09}{0.02}\\
\bottomrule
\end{tabularx}

%% file: analysis/visual_v42/service_v42/TABLE11_BODY.tex
\begin{tabularx}{\linewidth}{>{\raggedright\arraybackslash}p{.20\linewidth}*{4}{>{\raggedleft\arraybackslash}X}}\toprule
\rowcolor{tablehead}\textcolor{white}{\textbf{Direction}}&\textcolor{white}{\makecell[r]{Actual records\\per request}}&\textcolor{white}{\textbf{Workers}}&\textcolor{white}{\makecell[r]{QPS\\(requests/s)}}&\textcolor{white}{\makecell[r]{Batch p95 latency\\(ms)}}\\\midrule
\rowcolor{tablepale}
RAG & 500 & 1 & \(28.46\,{\color{gray}\scriptstyle\pm}\,{\color{gray}\phantom{0}0.21}\) & \(637.74\,{\color{gray}\scriptstyle\pm}\,{\color{gray}28.47}\)\\
RAG & 500 & 4 & \(103.19\,{\color{gray}\scriptstyle\pm}\,{\color{gray}\phantom{0}2.71}\) & \(228.07\,{\color{gray}\scriptstyle\pm}\,{\color{gray}\phantom{0}6.51}\)\\
\rowcolor{tablepale}
RAG & 500 & 8 & \(\mathbf{184.56}\,{\color{gray}\scriptstyle\pm}\,{\color{gray}\phantom{0}8.90}\) & \(113.02\,{\color{gray}\scriptstyle\pm}\,{\color{gray}\phantom{0}6.80}\)\\
\midrule
RAG & 1,000 & 1 & \(12.03\,{\color{gray}\scriptstyle\pm}\,{\color{gray}\phantom{0}0.25}\) & \(1624.14\,{\color{gray}\scriptstyle\pm}\,{\color{gray}19.78}\)\\
\rowcolor{tablepale}
RAG & 1,000 & 4 & \(56.93\,{\color{gray}\scriptstyle\pm}\,{\color{gray}\phantom{0}0.54}\) & \(414.79\,{\color{gray}\scriptstyle\pm}\,{\color{gray}\phantom{0}3.99}\)\\
RAG & 1,000 & 8 & \(\mathbf{103.33}\,{\color{gray}\scriptstyle\pm}\,{\color{gray}10.18}\) & \(213.65\,{\color{gray}\scriptstyle\pm}\,{\color{gray}14.72}\)\\
\midrule
\rowcolor{tablepale}
Reasoning & 500 & 1 & \(15.47\,{\color{gray}\scriptstyle\pm}\,{\color{gray}\phantom{0}0.25}\) & \(1342.41\,{\color{gray}\scriptstyle\pm}\,{\color{gray}28.37}\)\\
Reasoning & 500 & 4 & \(80.95\,{\color{gray}\scriptstyle\pm}\,{\color{gray}\phantom{0}0.50}\) & \(297.41\,{\color{gray}\scriptstyle\pm}\,{\color{gray}\phantom{0}5.04}\)\\
\rowcolor{tablepale}
Reasoning & 500 & 8 & \(\mathbf{141.79}\,{\color{gray}\scriptstyle\pm}\,{\color{gray}\phantom{0}9.60}\) & \(168.59\,{\color{gray}\scriptstyle\pm}\,{\color{gray}31.86}\)\\
\midrule
Reasoning & 1,000 & 1 & \(8.06\,{\color{gray}\scriptstyle\pm}\,{\color{gray}\phantom{0}0.09}\) & \(2533.01\,{\color{gray}\scriptstyle\pm}\,{\color{gray}40.19}\)\\
\rowcolor{tablepale}
Reasoning & 1,000 & 4 & \(28.47\,{\color{gray}\scriptstyle\pm}\,{\color{gray}\phantom{0}1.01}\) & \(605.15\,{\color{gray}\scriptstyle\pm}\,{\color{gray}46.92}\)\\
Reasoning & 1,000 & 8 & \(\mathbf{79.61}\,{\color{gray}\scriptstyle\pm}\,{\color{gray}\phantom{0}4.04}\) & \(285.49\,{\color{gray}\scriptstyle\pm}\,{\color{gray}\phantom{0}5.73}\)\\
\bottomrule\end{tabularx}

%% file: experiments/cli_examples.tex
\Needspace{8\baselineskip}
\subsection{Profiles and explicit frozen inputs}
\begingroup\interlinepenalty=10000
The commands below use existing flags. A topic's display name is not necessarily its command identifier: for example, the Agent/Harness identifier is \texttt{agent-harness}. Enumerate identifiers before constructing a profile. A frozen input has the shape \texttt{\{"records":[\{"paper":\{...\}\}]\}}; each paper carries its identifier, title, abstract in \texttt{summary}, byline, dates and categories.\par
\endgroup
\begin{commandpanel}[unbreakable]{Discover topic identifiers and save a reusable profile}
node scripts/rank-research-papers.mjs --list-topics > topics.json
cat > profile.json <<'JSON'
{"name": "Agent research",
 "topics": {"reasoning": 1, "agent-harness": 2}}
JSON
\end{commandpanel}

\begin{commandpanel}[unbreakable]{Rank a frozen input, or stream it from another process}
node scripts/rank-research-papers.mjs \
  --input frozen-papers.json --profile profile.json \
  --from 2026-09-01 --to 2026-09-30 --exclude survey \
  --limit 20 --format json > ranked.json

cat frozen-papers.json | node scripts/rank-research-papers.mjs \
  --input - --profile profile.json \
  --from 2026-09-01 --to 2026-09-30 --exclude survey \
  --limit 20 --format jsonl > ranked.jsonl
\end{commandpanel}
The file and stream forms share the same ranking path. JSON contains one object with \texttt{manifest} and \texttt{papers}; JSONL begins with a manifest record, followed by records whose \texttt{type} is \texttt{paper}. An agent should check the process exit status and parse stdout, while retaining stderr for diagnostics.

\subsection{From ranked evidence to retrieval documents}
This small consumer turns the returned papers into a source-linked retrieval input and an explicit identifier list. It preserves the snapshot identifier so that a downstream answer can be traced back to the selection that supplied its context.
\begin{commandpanel}{Extract RAG documents and identifiers from JSONL}
python3 - <<'PY'
import json
from pathlib import Path
rows = [json.loads(s) for s in
        Path('ranked.jsonl').read_text().splitlines() if s.strip()]
papers = [r for r in rows if r.get('type') == 'paper']
docs = [{'id': p['id'], 'title': p['title'],
         'text': p['abstract'], 'source': p['url'],
         'snapshotId': p['snapshotId']} for p in papers]
Path('rag-documents.jsonl').write_text(
    ''.join(json.dumps(d) + '\n' for d in docs))
Path('selected-ids.json').write_text(
    json.dumps([p['id'] for p in papers], indent=2))
PY
\end{commandpanel}
These documents can feed an existing retrieval system or an agent's context assembly. For training-data preparation, retain the same source identifiers and evidence as a reviewable candidate pool; selection scores do not supply factual labels or permission to reuse a source. The supplementary consumer additionally writes \texttt{source-candidates.jsonl} with review status and provenance, without creating training supervision.

\subsection{Scoped graph export and inspection}
The selected identifiers can bound a graph independently of the display limit in the browser. The first command below uses only the supplied input and cached metadata. The edge bound limits export size; the manifest records the resulting scope and coverage.
\begin{commandpanel}{Export exactly the selected input and inspect its manifest}
node scripts/export-knowledge-graph.mjs \
  --input frozen-papers.json --ids selected-ids.json \
  --max-edges 50000 --output snapshots/selected-01

python3 -m json.tool snapshots/selected-01/manifest.json
python3 - <<'PY'
import json
from collections import Counter
from pathlib import Path
p = Path('snapshots/selected-01/citation-coverage.jsonl')
states = Counter(json.loads(s)['status']
                 for s in p.read_text().splitlines() if s.strip())
print(dict(states))
PY
\end{commandpanel}
The coverage file separates unobserved reference metadata from an observed empty list. The recorded coverage state identifies which reference observations are available.

\subsection{A bounded citation refresh and continuation}
A citation refresh is the explicit network-enabled path. Keep the input snapshot and citation-cache directory fixed across continuations, while choosing a fresh graph-output directory for each export.
\begin{commandpanel}{Acquire one citation batch; continue a paused checkpoint}
node scripts/export-knowledge-graph.mjs \
  --input frozen-papers.json --ids selected-ids.json \
  --refresh-citations --max-pages 1 \
  --citation-cache evidence/selected \
  --output snapshots/selected-02

python3 -m json.tool evidence/selected/citation-checkpoint.json

# Only when the checkpoint state is "paused":
node scripts/export-knowledge-graph.mjs \
  --input frozen-papers.json --ids selected-ids.json \
  --refresh-citations --max-pages 1 \
  --citation-cache evidence/selected \
  --output snapshots/selected-03
\end{commandpanel}
A completed acquisition is reused. A partially completed refresh returns exit code 2 and retains its checkpoint. Rate-limited, access-blocked and explicitly stopped states are not automatically retried. Input changes require a new citation-cache directory.

\paragraph{Example verification.}
An offline check on six frozen public paper records enumerated the topic identifiers, exercised both file and stdin/profile ranking, and returned the same four paper IDs and scores. Exporting those four IDs produced 41 nodes and 207 relations. All four citation-coverage states were \texttt{not-observed}; no network refresh or model call ran in this check. These are interface checks on a small fixture, not measurements of retrieval quality or citation coverage. The commands require the product project and its dependencies.

%% file: literature/aminer_method_context.tex
% Proposed insertion into the fixed-author-prior subsection; local implementation facts.
% The AI2000 citation covers the archived lists and the 2026 edition.
\paragraph{Historical author priors and source attribution.}
The fixed catalogue includes locally retained AMiner-derived candidate lists
\citep{wan2019aminer,aminerai2000}. The checked implementation labels the LLM-era
windows as 2022--2025 and 2023--2025, and the recent-author window as 2024--2025.
Repeated catalogue entries are consolidated using the available name and identity keys, and supplied scores, ranks, and
rising-author tags pass through bounded local transformations before contributing
to a paper score. These transformations are our ranking rules, not probabilities
of paper correctness supplied by AMiner. More recent paper-level identity
checks do not refresh the historical ranking windows. The catalogue also includes the
2026 edition of the AI 2000 lists, read once on 26 November 2026, and a slice of
CSRankings faculty~\citep{csrankings2026}, which the extension downloads on the
reader's device; names shared with an archived list keep their archived level.

% Optional precise identity clarification if the surrounding manuscript makes
% a universal "all author matches are verified" statement.
Direct paper or author-affiliation bindings are marked as verified, and a coauthor
shared with earlier verified work links a name to its person; any other catalogue
match stays pending and earns no credit (Sec.~\ref{sec:identity}). Pending matches
remain distinguishable from a resolved identity and from the user's
topic-conditioned adjustment.

%% file: experiments/author_calibration_refresh_v32.tex
\subsection{Calibrating author concentration against background prevalence}
\label{sec:author-calibration}
A fixed 50\% neutral point can penalize a narrow-field author even when their concentration exceeds the bibliography's prevalence. We compare this fixed-threshold reference with a leave-author-out reference, changing only the author-profile weight. The complete thirteen-month population and sixteen declared LLM directions are fixed independently of citation or reading-utility outcomes.

Of 148,150 papers first submitted from October 2025 through October 2026, excluding one administrative record and 32 exact duplicates leaves 148,117 candidates. Title/abstract matching yields 58,079 papers and \textbf{88,619 direction--paper pairs}; 90,038 unmatched candidates remain outside these lists. Both arms use the complete frozen title bibliography, identical anchored identities and a 4 November 2026 scoring clock. This archived ablation holds its mean-based institution term fixed and compares the two author references; Sec.~\ref{sec:current-reading-response} instead compares the credit-weighted score with relevance alone.

\begin{figure}[t!]\centering
\includegraphics[width=\linewidth]{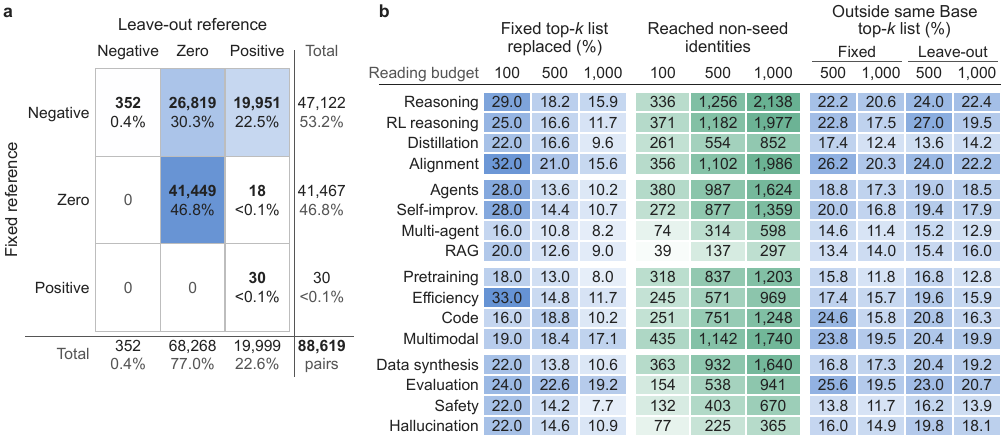}
\caption{\textbf{Author-reference calibration in a frozen institution context.} (a) Author-adjustment sign for all 88,619 direction--paper pairs under the fixed reference (rows) and leave-out calibration (columns), with each count's share of all pairs and the marginal totals. Zero includes rounded-zero adjustments and insufficient author evidence. (b) One row per direction, grouped by the four research families: the percentage of each fixed-reference top-$k$ list replaced by leave-out calibration at 100, 500 and 1,000 reads; distinct reached non-seed identities in the corresponding Personal lists; and the percentage outside the same Base top-$k$ list under each reference at 500 and 1,000 reads. Shading follows the printed values within each column group (log scale for identities). Directions may share papers and authors.}\label{fig:author-calibration}
\end{figure}

\Needspace{12\baselineskip}
Median leave-out reference fits range from 0.68\% to 2.08\%, far below the fixed 50\% threshold. Negative, zero and positive adjustments change from 47,122, 41,467 and 30 under the fixed reference to 352, 68,268 and 19,999 under the leave-out reference. Top-100 lists replace 16--33 entries per direction (mean 23.5; Tab.~\ref{tab:author-calibration} and Fig.~\ref{fig:author-calibration}).

\begin{table}[!hbp]\centering\fontsize{9}{11}\selectfont
\caption{\textbf{Author-reference list changes at three reading budgets.} Each count is the number of entries in the fixed-reference top-$k$ list replaced under leave-out calibration, with the frozen mean-based institution term shared by both rules. All sixteen directions have complete lists at 100, 500 and 1,000 reads. Matching counts sum to 88,619 direction--paper pairs and overlap across directions. The final row gives arithmetic means of replaced-entry counts; these list changes are descriptive selection outcomes.}\label{tab:author-calibration}
\renewcommand{\arraystretch}{1.30}\setlength{\tabcolsep}{4pt}
\input{analysis/institution_profile_v48/repaired_dependents/TABLE15_BODY}
\end{table}

\begingroup\interlinepenalty=10000
The \textbf{141,600 eligible author--profile records} comprise 8,850 identities in sixteen directions. Of 133,977 bound profiles per direction, 125,127 lack sufficient title evidence and retain weight 1. Each author's profile weight applies across their matching papers. Both arms fix identity bindings, position weights, time, a zero manual bonus, institutional context and the $\pm6$ author-adjustment bound; all non-author terms match. Zero paper adjustments include insufficient evidence and small eligible contributions rounded to zero.\par
\endgroup

Author--profile and paper counts are distinct units: one identity can contribute to many papers, and one paper can match several directions. At budgets of 500 and 1,000, the reference change replaces an average of 79.4 and 116.4 entries, respectively.

%% file: analysis/institution_profile_v48/repaired_dependents/TABLE15_BODY.tex
\begin{tabularx}{\linewidth}{Yrrrr}\toprule
\rowcolor{tablehead}\textcolor{white}{\textbf{Direction}} & \textcolor{white}{\makecell{Matching\\papers}} & \textcolor{white}{\makecell{Top-100\\replaced}} & \textcolor{white}{\makecell{Top-500\\replaced}} & \textcolor{white}{\makecell{Top-1,000\\replaced}}\\\midrule
\rowcolor{tablepale}
Reasoning & 9,603 & 29 & 91 & 159\\
RL reasoning & 4,830 & 25 & 83 & 117\\
\rowcolor{tablepale}
Distillation & 3,105 & 22 & 83 & 96\\
Alignment & 7,565 & 32 & 105 & 156\\
\midrule
\rowcolor{tablepale}
Agent harnesses & 3,826 & 28 & 68 & 102\\
Self-improvement & 3,849 & 28 & 72 & 107\\
\rowcolor{tablepale}
Multi-agent & 2,654 & 16 & 54 & 82\\
RAG & 4,260 & 20 & 63 & 90\\
\midrule
\rowcolor{tablepale}
Pretraining & 4,087 & 18 & 65 & 80\\
Efficiency & 6,846 & 33 & 74 & 117\\
\rowcolor{tablepale}
Code LLMs & 3,828 & 16 & 94 & 102\\
Multimodal & 9,073 & 19 & 92 & 171\\
\midrule
\rowcolor{tablepale}
Data synthesis & 6,789 & 22 & 69 & 106\\
Evaluation & 8,431 & 24 & 113 & 192\\
\rowcolor{tablepale}
Safety & 4,260 & 22 & 71 & 77\\
Hallucination & 5,613 & 22 & 73 & 109\\
\midrule
\rowcolor{tablepale}Mean across directions & --- & 23.50 & 79.38 & 116.44\\
\bottomrule\end{tabularx}

%% file: analysis/revision_v68/TABLE_EXAMPLES_v68.tex
\begin{table}[tbp]\centering\small
\caption{\textbf{Five papers drawn by rule rather than chosen.} The papers are the first five by SHA-256 of their identifiers among the 266 papers in the top tenth of October 2026 in the common pool of 2,664 October papers. Ranks are within all 21,838 papers of the pool; $R$ is relevance, $M_a$ the author factor, and the last name the strongest identity-bound author.}\label{tab:examples-v68}
\setlength{\tabcolsep}{4pt}\renewcommand{\arraystretch}{1.12}
\begin{tabular}{lrrrrlr}\toprule
\rowcolor{tablehead}\textcolor{white}{\textbf{Paper}}&\textcolor{white}{\textbf{Rank}}&\textcolor{white}{\textbf{Relevance-only}}&\textcolor{white}{$\boldsymbol{R}$}&\textcolor{white}{$\boldsymbol{M_a}$}&\textcolor{white}{\textbf{Bound author}}&\textcolor{white}{$\boldsymbol{s}$}\\\midrule
\radarpreprint{2610.22732}&75&941&26.1&1.43&J.~Gao&35.8\\
\radarpreprint{2610.15782}&165&303&30.5&1.15&Y.~Qiao&32.5\\
\radarpreprint{2610.18813}&218&86&33.0&1.00&--&31.1\\
\radarpreprint{2610.17266}&301&131&32.2&1.00&--&30.1\\
\radarpreprint{2610.16577}&105&814&27.5&1.35&H.~Zhao&34.4\\
\bottomrule\end{tabular}\end{table}

%% file: experiments/legacy_discovery_appendix.tex
\Needspace{12\baselineskip}
\section{Personalization in the September candidate cohort}
\label{app:legacy-discovery}
\subsection{Shared candidates and publication-separated context}
The six-direction mechanism and cost comparisons use a fixed-threshold author reference. A twelve-direction extension uses the leave-author-out reference on the same September history and candidate pool; Sec.~\ref{sec:author-calibration} separately evaluates author-reference calibration on the thirteen-month, sixteen-direction cohort. Interface examples retain their recorded scores and observation dates.
From the 112,663-record source snapshot, the replay selects \textbf{96,253 historical papers} whose first publication and current-version update precede September 2026, and \textbf{9,790 September candidates} whose current version predates October. Target IDs are excluded from history. Six declared directions cover Agents, RAG, Reasoning, Medical AI, Multimodal, and Safety/Alignment; these are configurations, not sampled human participants.

Five variants compare topic plus age, the public Base score, Base plus personal author context, Base plus personal institution context, and full personalization (Tab.~\ref{tab:ablation}). Topic matching and scoring use the same functions, without fitting weights or thresholds to these results. Historical contexts exclude candidates and later external author metrics. Publication and version dates determine which papers enter the historical graph. The replay uses the catalogue and implementation archived for this study; its institution term belongs to an additive score and is not part of Eq.~\eqref{eq:personal} (App.~\ref{app:institution-audit}).

\begin{table}[!hbp]\centering\small
\caption{\textbf{Reading-list changes under personal context.} Means use the fixed-threshold author reference across six configurations and 9,790 candidates. Overlap measures agreement with Full; catalogue/network exposure and the institution-label Herfindahl--Hirschman index (HHI) summarize list composition. Tint highlights the complete configuration.}\label{tab:ablation}
\setlength{\tabcolsep}{8pt}\renewcommand{\arraystretch}{1.12}
\begin{tabularx}{\linewidth}{L{5.3cm}CCCC}\toprule
\rowcolor{tablehead}\textcolor{white}{\textbf{Variant}} & \textcolor{white}{\makecell{Top-100\\overlap (\%)}} & \textcolor{white}{\makecell{Catalog\\exposure (\%)}} & \textcolor{white}{\makecell{Network\\exposure (\%)}} & \textcolor{white}{\makecell{Institution\\HHI}}\\\midrule
\grouphead{5}{A. Content and public evidence}
Topic + age &67.2&1.2&8.8&0.054\\
Base score &90.7&10.8&13.8&0.039\\\midrule
\grouphead{5}{B. Adding direction-conditioned evidence}
+ personal author &99.0&9.7&11.3&0.040\\
+ personal institution &90.3&10.8&13.8&0.041\\
\rowcolor{knowlight}\textbf{Full personal} &100.0&9.8&11.5&0.041\\\bottomrule
\end{tabularx}
\end{table}

\subsection{Reading-budget response and evidence coverage}
Full personalization replaces \textbf{7, 13, 9, 12, 8, and 7} Base top-100 entries across the six directions, averaging \textbf{9.33\%}. Author terms explain most changes; institution-only terms replace one or two entries per direction. Eligible personal-author bindings cover 1,749 of 9,790 candidates (17.87\%), and institution context covers 1,996 (20.39\%). Candidates without eligible evidence retain neutral adjustments.
\begin{figure}[t!]\centering
\includegraphics[width=\linewidth]{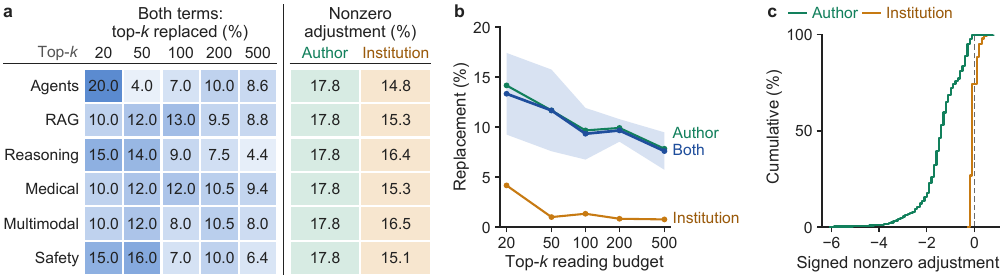}
\caption{\textbf{Reading-budget response, evidence coverage, and signed adjustments.} All panels use the fixed-threshold reference and 9,790 September candidates. (a) Base-top-$k$ replacement with both personal terms at $k=20,50,100,200,500$; the last two columns give the percentage of candidates with a nonzero author or institution adjustment. (b) Mean replacement across the six profiles for each term set; the band is $\pm$ one sample SD for the Both set. (c) Cumulative distributions of nonzero signed adjustments. Repeated direction--paper pairs are dependent.}\label{fig:response}
\end{figure}
Fig.~\ref{fig:response} shows that replacement varies with both direction and reading budget, which a single top-100 value cannot capture. Of 10,479 nonzero personal-author direction--paper adjustments, only one is positive. Under the fixed threshold, authors with limited directional match in their available title histories mostly receive lower adjustments. This motivates calibration against background prevalence.

The history contains 75,122 inferred author profiles, of which 3,117 meet the ten-title requirement. A separate author-value calculation identifies 361 eligible profiles, including 277 beyond the fixed catalogue after overlap removal. These locally inferred candidates retain the publications supporting their discovery.

\subsection{Cross-direction overlap and institutional exposure}
\begin{figure}[t!]\centering
\includegraphics[width=\linewidth]{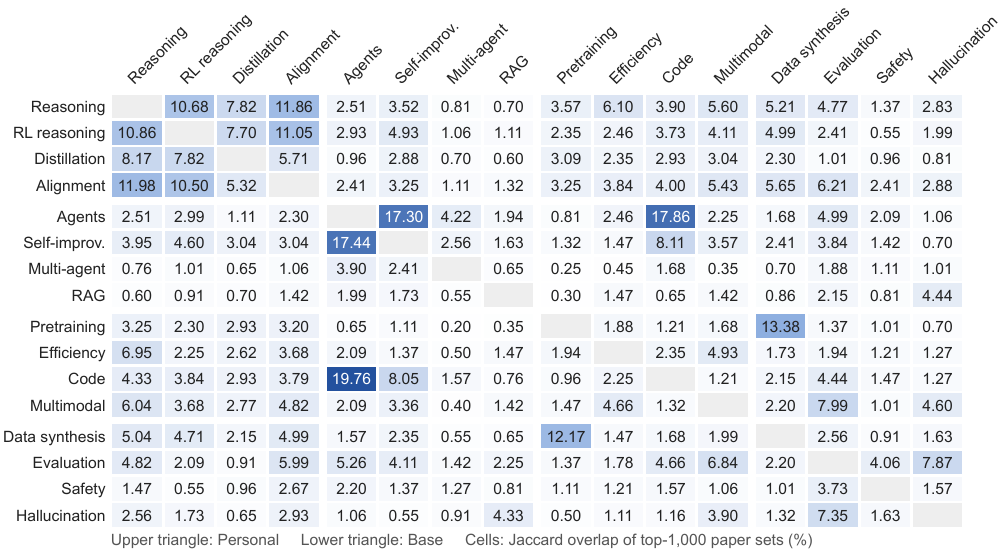}
\caption{\textbf{Overlap between related reading directions.} Each cell prints the Jaccard overlap (\%) between the top-1,000 paper sets of two directions: Base below the diagonal and the archived additive Personal score, without document-level terms, above it. The diagonal is blank; both triangles share one 0--20\% shading scale. All 500- and 1,000-read values remain in the supplement.}\label{fig:cross}
\end{figure}

With the fixed-threshold reference in six directions, mean pairwise top-100 overlap decreases from 23.40\% for Base to 21.73\% for Full. Verified catalogue exposure changes from 10.83\% to 9.83\%, while institution-label Herfindahl--Hirschman index (HHI) rises from 0.0390 to 0.0412. Thus the lists become more differentiated without uniformly reducing concentration.
Fig.~\ref{fig:cross} adds a sixteen-direction comparison under the archived additive Personal score at 1,000 reads: the split matrix prints all 120 pairs under both scores.

A separate twelve-direction extension of the historical replay applies the leave-author-out reference to the six declared profiles and six additional catalogue profiles, chosen before inspecting their outputs. The history and September candidates remain fixed. Mean overlap increases from 22.65\% to 23.03\% across 66 pairs, providing no evidence of a general increase in differentiation. The direction set and author reference differ from the six-direction comparison. Greater or smaller overlap can be appropriate when research topics share methods.

HHI divides each paper's unit mass among its verified listed institution IDs and excludes unknown affiliations from the denominator. It summarizes the distribution of evidenced institutions within each reading list.

%% file: experiments/discovery_evidence_appendix.tex
\section{Reading-list evidence, topic roles and reproducible scope}
\label{app:discovery-evidence}
\begin{figure}[t!]\centering
\includegraphics[width=\linewidth]{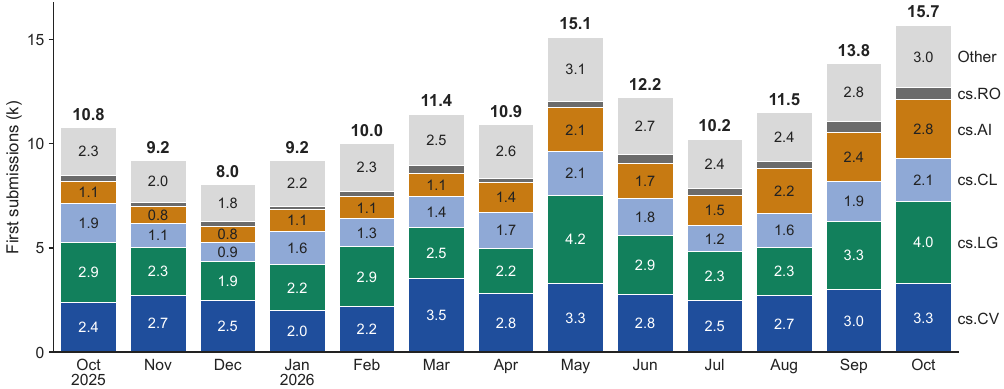}
\caption{\textbf{The expanded library exposes both research volume and topic coverage.} Monthly bars show 148,150 unique first submissions from October 2025 through October 2026 (13 complete months), grouped by primary category. Numbers are thousands of submissions; the thin cs.RO segments (0.2--0.6 thousand) are unlabelled. The incomplete November tail is omitted from this display. The half-year topic comparison appears in App.~\ref{app:topic-census}; complete monthly series accompany the source package. Counts describe this acquired library, not all arXiv or historical topic assignments.}\label{fig:corpus-topic-census}
\end{figure}

The frozen recent-window replay uses all 40 declared topics without outcome-based selection and retains its original scoring strategy. The acquisition contains 174,640 in-scope records (Fig.~\ref{fig:corpus-topic-census}), and the recent reading task contains 14,461 papers. Topic filtering produces 14,887 direction--paper pairs spanning 8,618 papers. These counts separate the acquired library, recent paper pool, direction--paper memberships and distinct topic-matched papers. The same native functions produce Content, Content + recency, Base and Personal scores. A preselected RAG profile reproduces all 200 ordinary-CLI identifiers and scores exactly, and the source digests match.

\subsection{Changes across successive half-year windows}
\label{app:topic-census}
\begin{figure}[t!]\centering
\includegraphics[width=\linewidth]{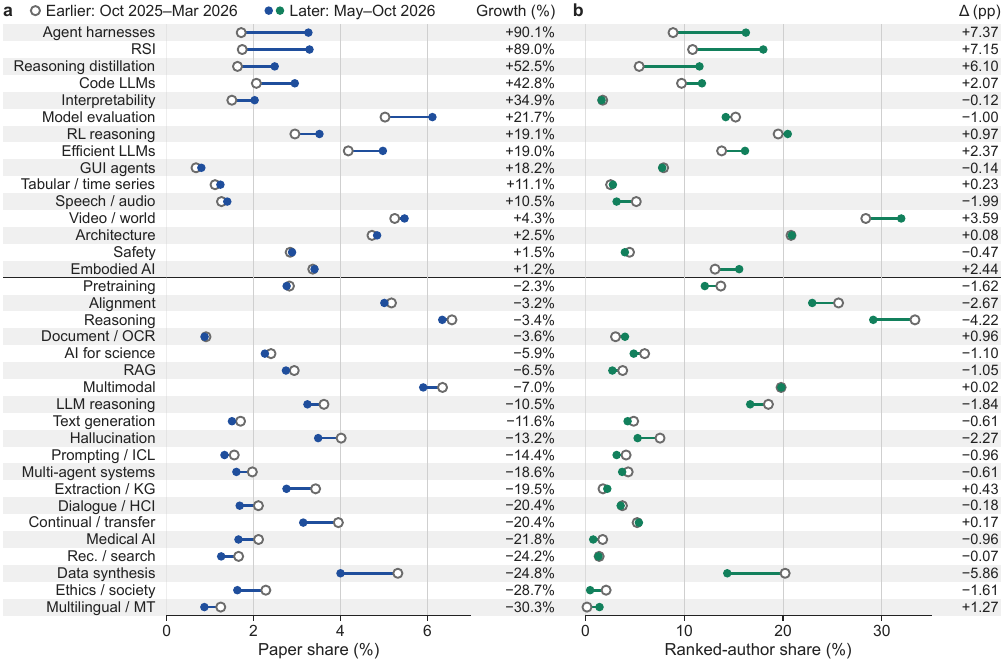}
\caption{\textbf{Changes in publication share and ranked-author participation.} (a) Paper share, with relative growth $100(p_L/p_E-1)$ printed; (b) ranked-author share, with the percentage-point change $p_L-p_E$ printed. Open and filled markers show the earlier (October 2025--March 2026) and later (May--October 2026) shares. Rows are ordered by the relative change in paper share, and the horizontal rule separates growing from shrinking paper shares. Paper shares use 58,645 and 78,585 acquired papers; author shares use summed monthly rank-weighted participation in the complete graph-admitted library. Authors count once per topic/month. The same 35 directions meet the retained support rule of ten ranked owners and six nonzero months. All forty directions and thirteen monthly observations remain in the numerical supplement.}\label{fig:corpus-topic-matrix}
\end{figure}

The comparison uses two equally long six-month windows to separate a direction's share change from growth of the whole collection. Fig.~\ref{fig:corpus-topic-matrix} orders directions by relative paper-share growth and shows ranked-author participation alongside it; the two measures capture different changes in research activity.

\subsection{A topic hit needs a reading purpose}
Title/abstract evidence can identify a method contribution, an application or a contextual mention. Tab.~\ref{tab:topic-role-cases} retains three RAG cases spanning these roles; ranks are read from that frozen recent-window replay. These source-role annotations are not independent relevance or quality judgments. A retrieval-method researcher may choose differently from an engineer seeking applications. The source text lets each reader judge relevance.

\begin{table}[tbp]\centering\small
\caption{\textbf{Three roles behind a literal RAG match.} Fixed recent-window ranks and exact source excerpts; cases remain in the reported lists. The method does not automatically infer the role annotations shown here.}\label{tab:topic-role-cases}
\setlength{\tabcolsep}{5pt}\renewcommand{\arraystretch}{1.18}
\begin{tabularx}{\linewidth}{L{1cm}L{5.3cm}Y}\toprule
\rowcolor{tablehead}\textcolor{white}{\textbf{Rank}}&\textcolor{white}{\textbf{Source and observed role}}&\textcolor{white}{\textbf{Abstract evidence}}\\\midrule
\rowcolor{tablepale}
1&UniGFM-RAG~\citep{km261034866}: method contribution&``a unified end-to-end GraphRAG framework''\\
77&Visual Tokens on Demand~\citep{km261018858}: related-method context&``retrieval-augmented visual reasoning'' appears among existing approaches; the proposed module budgets visual tokens.\\
\rowcolor{tablepale}
156&Indigenous Community Cosmovision~\citep{km261025298}: application component&``deployed in a conversational interface with retrieval-augmented generation (RAG)''\\\bottomrule
\end{tabularx}
\end{table}

\subsection{Numerical and source boundaries}
The numerical supplement provides all 40-profile summaries, 780 direction-pair overlaps, monthly topic counts and 120 sensitivity results, with frozen anonymous scores and export manifests for reproducing reading-budget metrics. Public identifiers and short source excerpts accompany the three cases; aggregate tables and plotting sources accompany the paper.

\Needspace{5\baselineskip}
Three fixed-clock exports have identical content hashes. Their counts separate source records, shared-author relations, paper-scoped identities and unobserved citations. These checks provide reproducible inputs for downstream RAG and research-agent evaluations.

\subsection{Related directions should share some papers}
\label{sec:current-cross-direction}
\begin{figure}[t!]\centering
\includegraphics[width=\linewidth]{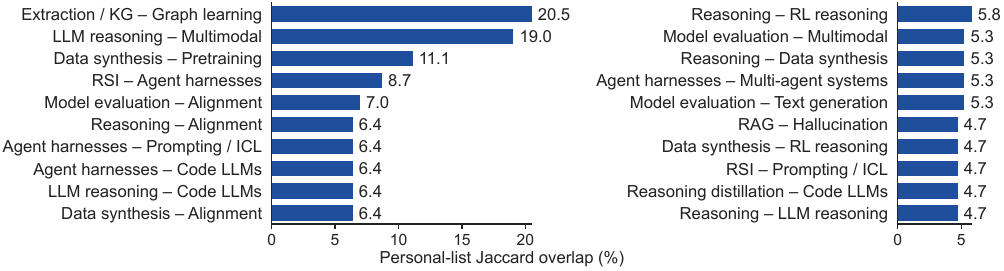}
\caption{\textbf{Shared papers connect related reading scopes.} The twenty pairs with the largest Personal-list overlap are shown in descending order, left block first, on one common scale; all 780 pairs remain in the supplement. Bars report actual Jaccard percentages.}\label{fig:discovery-cross-direction}
\end{figure}

Related research questions can legitimately share papers. We report list overlap alongside candidate counts and evidence coverage to show these relationships. Fig.~\ref{fig:discovery-cross-direction} compares up to 100 actual results per topic using Jaccard overlap, including smaller eligible sets.

\input{experiments/discovery_overlap_readout}

%% file: experiments/discovery_overlap_readout.tex
\Needspace{5\baselineskip}
In the frozen 40-topic recent-window replay, mean Jaccard overlap across all 780 distinct pairs is 0.955\% for Base and 0.943\% for Personal. It falls for 45 pairs, is unchanged for 703, and rises for 32. The three smallest directions retain their actual 11--19 candidates, rather than being padded to a nominal budget.

%% file: experiments/institution_evidence_v68.tex
\subsection{Institutions: evidence and why they stay out of the score}
\label{app:institution-audit}
An additive variant added an institutional term that mixed the direction activity of an institution's papers with a documented LLM-field standing $B$ (8 October 2026), and a multiplicative variant multiplied each paper by an institution factor that also added a QS/THE reputation axis for universities. Eq.~\eqref{eq:personal} carries no institution weight, for three reasons, each measured on the same library.

\paragraph{Documentation, not evidence, set the level.} In the multiplicative factor, $B$ explained 98\% of the variance of the union institution context across 114 parents, and 77 parents shared one level because unassessed axes counted as zero. Of the 121 catalogue entries, 87 rested on a single source domain, so a reader could not trace why one laboratory outranked another beyond the catalogue itself.

\paragraph{People aggregates predict, but measure size or coverage.} The fifteen-slot people core of the institution card, computed at each month's start from earlier papers only, adds 1.5 points (0.3--2.7) to the hit rate and 2.1 points to the floor of the check months' top tenth beyond author credit. It correlates with an institution's paper count at 0.73 on the 21 November snapshot and 0.69 at the end of August, and it ranks OpenAI, DeepSeek, Moonshot and Anthropic between 83rd and 102nd of 105 because their few papers leave the core underfilled. The share of catalogue researchers among an institution's direction-active authors is size-free but depends on how fully the seed catalogue covers its roster: Google sits 65th of 113 and Alibaba 76th, while small laboratories with one or two catalogue researchers lead. Size-free variants that only ever raise a paper changed fitted hit rates and floors by at most 0.8 points combined, and their check-month top-decile intervals included zero. Because the design requires that institution weight not grow with headcount, the people core stays a display.

\paragraph{An external signal exists but is confounded.} In the sealed test, an institution's track record of early citations to its 2025 papers predicted top-decile papers beyond their people (risk ratio 1.46, 1.22--1.75), and so did the density of bound researchers on its roster (1.39, 1.16--1.67). Both fell under finer size control (1.31 and 1.26); the track record correlates 0.58 with publication volume; and both reversed on declared top-venue acceptance (0.64 and 0.72). Over the 125 institutions with at least twenty papers, a three-month track record placed Notre Dame first, Tsinghua 26th, Peking University 32nd and Google 58th, which is not a sound basis for a reading prior.

The institution card therefore reports the people behind a paper's context, and affiliations serve identity binding (Sec.~\ref{sec:identity}). A future institution term must beat author credit on held-out lists without tracking size, which requires at least a year of outcomes and audited parent mappings.

\subsection{First-author bursts}
\label{app:burst}
The burst rule of Eq.~\eqref{eq:author-credit} uses first submissions on record, byline names normalized as in Sec.~\ref{sec:identity} with teams of at most thirty, and TF--IDF cosines of title and abstract with document frequencies over the papers in candidate pairs. Of 176,220 dated papers in the local store, which holds the library and the author-graph records, 2,831 fall in a burst; the median cosine between burst siblings is 0.37. Among the 37,904 papers with outcomes from January to October 2026, the 447 burst papers (1.2\%) reached the top early-citation decile half as often as other papers of the same month (Mantel--Haenszel risk ratio 0.49, 95\% bootstrap interval 0.30--0.70) and cleared the month median slightly less often (0.89, 0.78--0.99). The direction holds in each window, with wide intervals: 0.57 (0.25--0.96) in January--April, 0.33 (0.11--0.59) in May--August and 0.68 (0.22--1.30) in September--October. Because the window looks both ways, an earlier paper shares its credit once a later sibling appears, and 126 of the 447 papers are in a burst only through a later sibling; for the other 321, whose sibling existed by their own submission day, the ratio is 0.23 (0.07--0.41). The rule is post hoc: it was added after the sealed test, and only 13 of these 447 papers carry a bound author, so it changes no monthly top-tenth list of Sec.~\ref{sec:heldout} and replaces one top-100 paper in each of two direction pools (efficient LLMs and reasoning) and none elsewhere (Sec.~\ref{sec:current-reading-response}). It is meant to keep one author's closely related papers from collecting the same credit twice, not to raise measured hit rates.

\subsection{Choosing the score form}
\label{app:tuning}
Factor forms were chosen on papers first submitted from January to April 2026 and checked on May--August, using the top tenth of each month and two constraints declared before the search: at most 5\% of picks below the month's median relevance, and at least 80\% of the newcomer winners found by relevance alone. A grid over exponential and saturating factors, shares of non-strongest authors and a relevance gate selected the saturating strongest-author form as the simplest within 0.005 of the best feasible hit rate (Tab.~\ref{tab:tuning-v68}). Of the grid's 224 forms, 151 met both constraints, and all of them beat relevance alone on the fitting months (Sec.~\ref{sec:heldout}). The grid rule preferred strength 1.5; we use 1.25 after seeing the check months, where it kept more newcomer winners (83.3\% against 80.3\%) at a similar hit rate, so the check months are not a clean estimate for this one parameter. Papers submitted from 23 November 2026, except 13 with relevance labels (App.~\ref{app:full-score}), are reserved for a prospective test. A Python reimplementation of Eq.~\eqref{eq:personal} as in the sealed test reproduces the frozen scores on all 58,094 union-profile papers; the burst rule (App.~\ref{app:burst}) is post hoc and outside this check.

\begin{table}[tbp]\centering\small
\caption{\textbf{Choices behind the score form.} Hit rates are the shares of each month's top tenth in the top citation decile; relevance alone reaches 14.8\% on the fitting months and 19.5\% on the check months. Newcomers is the share of newcomer winners kept relative to relevance alone. Values come from the tuning replay, which recomputes each form from recorded score inputs; the frozen scores at $a=1.25$ give 23.3\% hits on the check months (Sec.~\ref{sec:heldout}).}\label{tab:tuning-v68}
\setlength{\tabcolsep}{4pt}\renewcommand{\arraystretch}{1.14}
\begin{tabularx}{\linewidth}{L{3.6cm}YYL{2.3cm}}\toprule
\rowcolor{tablehead}\textcolor{white}{\textbf{Choice}}&\textcolor{white}{\textbf{Fitting months (Jan--Apr)}}&\textcolor{white}{\textbf{Check months (May--Aug)}}&\textcolor{white}{\textbf{Decision}}\\\midrule
\rowcolor{tablepale}Strongest author, $a=1.25$&18.4\% hits; newcomers 82.5\%&22.9\% hits; newcomers 83.3\%&Used\\
Strongest author, $a=1.5$&18.7\% hits; newcomers 82.5\%&22.5\% hits; newcomers 80.3\%&Not used\\
\rowcolor{tablepale}Binary bound-author bonus&+2.5 points of hits and floor&newcomers 68.2\%&Rejected\\
All bound authors summed&within 0.005 of strongest only&second author: risk ratio 0.96 (sealed test)&Not used\\
\rowcolor{tablepale}Institution context&size-free variants: at most +0.8 points&people core: +1.5 points, but tracks size&Shown only\\
Report factor 1.25&median report in the top 12\% of its month&top 10\%&Used\\
\rowcolor{tablepale}Cached reading evidence $T$&--&--&Not used\\\bottomrule
\end{tabularx}\end{table}

%% file: experiments/full_score_check.tex
\Needspace{8\baselineskip}
\section{The full score: choices and a pre-declared check}
\label{app:full-score}
The full score differs from the sealed-test score in three terms: the 2026 AI 2000 and CSRankings entries of the catalogue with their levels, the mention weight and the collaboration rule. These were chosen on papers first submitted from January to August 2026, against early-citation outcomes and the two constraints of App.~\ref{app:tuning}, with May--August as check months; the catalogue levels and the mention weight also used 94 relevance labels (0--2) on one reader's papers of 27 October--26 November 2026 (13 first submitted from 23 November), given by an LLM judge that saw only titles, abstracts and the reader's directions, blind to the score. No constant was fitted on September--October outcomes.

\paragraph{Catalogue.} The 2026 AI 2000 lists name 2,744 researchers; 2,661 are kept, after removing 50 persons who share a name with another AMiner person, 23 entries whose organization conflicts with an archived entry of the same name, 6 without an organization and 4 held out as identity risks. Rating the 262 names that are already on an archived list by the archived citation thresholds would have raised their catalogue input by 1.4 points on average and moved off-topic papers by famous authors into the reader's top ten (mean label 1.90 to 1.50), so these names keep their archived level; new names enter at level B with at least 75,000 citations at top AI venues, C with 20,000 and D otherwise. Placing every new name at level C cost about six points of newcomer recall. CSRankings faculty enter only from the top slice, the 531 faculty with at least six author-weighted papers at fourteen CCF A (full weight) and B (half weight) AI venues in 2021--2025 at the sixty departments with the most such output, and the 350 not already in the catalogue only at level D: all 3,204 listed faculty at their first levels broke the newcomer constraint (recall 0.69--0.76), and on the check months papers led by the slice reached the top decile about as often as papers by new AI 2000 names, other faculty about as often as papers without a known author. The CSRankings table is licensed CC BY-NC-ND 4.0, so the extension downloads it on the reader's device and nothing derived from it ships.

\paragraph{Mentions.} A topic term counts fully in the title, the opening sentence, a stated contribution or more than one sentence, three quarters as a single passing mention and zero when negated. A weight of one half cost 1.4 points of the relevance-only hit rate on the check months, whereas three quarters cost 0.8 ($-1.7$ to $0.2$) and separated the reader's off-topic papers nearly as well.

\paragraph{Collaboration rule.} The narrowest variant credited only collaborators with strongly on-topic evidence papers (base score at least 24) and seeded only from authors verified on the paper. Seeding also from linked authors and removing this floor raised the share of each month's top tenth with a credited author from 3\% to 37\% (at a 30\% share), and credited papers still reached the top decile about twice as often as comparable papers without a known author. A second round added 6.7 points of coverage, and its papers were still 1.3 times as often hits as expected; a third added 0.6 points with the weakest papers (hit rate 0.213) and lowered newcomer recall with age from 0.862 to 0.845, so the rule stops at two rounds. Excluding names held by four or more identities from the second round removed name collisions, raised the hit rate by 0.3 points and age-neutral newcomer recall from 0.741 to 0.776. Tab.~\ref{tab:collab-variants} compares the byline-role split with its alternatives at the chosen share; sharing by role lowered coverage but raised the hit ratio of credited papers, most in the second round. On the reader-scale graph of 40,386 papers, the chosen rule builds 642 profiles (450 with collaboration credit) in 14.7~s. Equal-contribution marks are absent from arXiv metadata, so a co-first second author counts as second.

\begin{table}[tbp]\centering\small
\caption{\textbf{Sharing by byline role trades coverage for precision.} Monthly top tenth of the check months (May--August 2026; 27 November snapshot), sixteen directions at equal priority; hit rate with age and age-neutral, newcomer winners kept against relevance alone (age-neutral), and the share of the top tenth with a credited author. The last column is the hit ratio of credited papers in the top two relevance fifths on January--August against papers of the same month, relevance fifth and team size without a known author (round 1 / round 2 for the split). All rows meet both declared constraints (age-neutral); rows below the first are the full score with each collaboration rule, at the share $\phi=0.15$ with cap 1.5.}\label{tab:collab-variants}
\setlength{\tabcolsep}{4pt}\renewcommand{\arraystretch}{1.12}
\begin{tabularx}{\linewidth}{Yrrrr}\toprule
\rowcolor{tablehead}\textcolor{white}{\textbf{Variant}}&\textcolor{white}{\textbf{Hit rate (\%)}}&\textcolor{white}{\textbf{Newcomers}}&\textcolor{white}{\textbf{Coverage}}&\textcolor{white}{\textbf{Credited ratio}}\\\midrule
\rowcolor{tablepale}Sealed-test score, no collaboration share&23.8 / 23.5&85.3\%&0\%&--\\
Two rounds, no sharing among coauthors&24.2 / 23.7&81.0\%&29.7\%&1.92 (round 2: 1.32)\\
\rowcolor{tablepale}Role split, $\kappa_0=5$&23.6 / 23.5&82.8\%&9.3\%&--\\
\textbf{Role split, $\kappa_0=2.5$ (chosen)}&24.0 / 23.7&81.0\%&23.1\%&\textbf{2.18} (2.16 / 2.23)\\
\rowcolor{tablepale}\quad last author weighted 1&24.0 / 23.6&81.0\%&26.2\%&--\\
\quad role weights also on the scored paper&23.7 / 23.6&81.0\%&22.9\%&--\\\bottomrule
\end{tabularx}\end{table}

Against the sealed-test score, the full score with the chosen rule changes the check-month hit rate by $+0.19$ points ($-0.97$ to $0.97$) with age and $+0.25$ ($-1.03$ to $1.15$) age-neutral, so we claim no difference. Weighting the last author at 1 credited more senior authors but lowered the second-round hit rate from 0.354 to 0.300, and role weights on the scored paper itself lowered the hit rate; neither was adopted. The share of 15\% is the largest that kept both constraints on the check months (20\% missed the age-neutral newcomer bound by one winner before the role split), and shares of 15--30\% differed in hit rate by at most 0.4 points.

\paragraph{The pre-declared check.} The September--October 2026 outcomes were opened once on 22 November 2026 for the sealed test (Sec.~\ref{sec:heldout}) and were reused descriptively in this paper, so no sealed test of the full score is possible. We wrote a protocol on 27 November 2026 before the full score met any of those outcomes, hashed it together with its code and inputs, and ran the outcome join once. It scores all September--October candidates of the sixteen directions at the clock of 21 November 2026 on a library snapshot of 27 November (177,500 records) and uses the outcomes of the sealed test: the top decile of OpenAlex citations within the submission month, counted on 22 November, and the floor above the month median. A harness check confirmed that the protocol's graph equals the extension's graph and that its collaboration-credit exposure equals that of the full score on all 11,741 candidates.
\begin{itemize}
\item \emph{Collaboration credit against placebo seeds (decisive).} Papers with an outcome, no catalogue name under either catalogue and prior author activity, as for the collaboration signal of the sealed test: 4,677 papers, 203 with collaboration credit. Each draw replaces every one of the 1,641 seed researchers by a random non-catalogue researcher from the same activity bucket ($\lfloor2\log_2 n\rfloor$ for $n$ papers in the library), seeded on as many of their papers at the seed's catalogue level, and reruns the unchanged propagation; papers that name a placebo seed leave that draw. The statistic is the matched risk ratio of a hit with the strata of the sealed test, and $p=(1+\#\{\text{draws}\geq\text{real}\})/101$ over 100 draws. Declared rule: supported only if the hit interval lies above 1, $p\leq0.05$ and the floor interval does not lie below 1.
\item \emph{Catalogue sources, lists and role split (descriptive).} Matched ratios by the source of the strongest bound author; each month's top tenth under relevance alone, the sealed-test score and the full score, with paired bootstrap intervals (1,000 draws within month); and the role-split ratio above on these months.
\end{itemize}

\paragraph{Results.} Collaboration credit gave a hit ratio of 1.41 (0.98--2.01) and a floor ratio of 1.04 (0.84--1.28), while the placebo seeds gave a mean hit ratio of 1.43 (95th percentile 2.42); 45 of 100 draws reached the real ratio ($p=0.46$; floor $p=0.80$). For credited authors ($\rho\geq1.5$) the hit ratio was 1.30 (0.88--1.92, $p=0.48$). The signal is therefore not shown. For reference, the sealed test's own exposure, a coauthor of a bound researcher, gives 1.13 (0.84--1.52) on the same population. On the 1,466 papers whose only catalogue names are unbound, the 180 with collaboration credit gave 1.44 (0.97--2.13) for hits and 1.39 (1.13--1.71) for the floor. The catalogue and list results are in Tab.~\ref{tab:sealed} and Tab.~\ref{tab:full-lists}: the full catalogue binds 2.6 times as many papers (196 hits among them against 110) at a lower ratio, and the full score shows no measurable hit-rate change while placing fewer picks below the median relevance. Credited papers in the top two relevance fifths were hits 1.70 times as often as comparable papers without a known author under the role split (441 papers), against 1.61 without it (633).

\begin{table}[tbp]\centering\small
\caption{\textbf{No measurable hit-rate change on the opened months.} Top tenth of September and October 2026 on the 27 November snapshot (1,036--1,038 papers per score), sixteen directions at equal priority; hits and floor are top-decile and above-median early citations; newcomers are newcomer winners kept against relevance alone (19 for the full score, 18 for the sealed-test score). Relevance alone differs between the two because the full score weights topic mentions. Paired differences, full minus sealed-test score: $-0.17$ points ($-1.24$ to $0.80$) with age and $-0.07$ ($-1.05$ to $0.80$) age-neutral.}\label{tab:full-lists}
\setlength{\tabcolsep}{4pt}\renewcommand{\arraystretch}{1.12}
\begin{tabularx}{\linewidth}{Yrrrr}\toprule
\rowcolor{tablehead}\textcolor{white}{\textbf{Top tenth}}&\textcolor{white}{\textbf{Hits}}&\textcolor{white}{\textbf{Floor}}&\textcolor{white}{\textbf{Below median}}&\textcolor{white}{\textbf{Newcomers}}\\\midrule
\rowcolor{tablepale}Relevance alone, sealed-test / full&12.6\% / 12.3\%&35.2\% / 35.1\%&0\%&--\\
Sealed-test score, with age&12.5\%&35.4\%&3.5\%&56\%\\
\rowcolor{tablepale}Sealed-test score, age-neutral&13.7\%&37.9\%&3.6\%&72\%\\
Full score, with age&12.4\%&34.7\%&2.3\%&58\%\\
\rowcolor{tablepale}Full score, age-neutral&13.6\%&37.9\%&2.2\%&79\%\\\bottomrule
\end{tabularx}\end{table}

Both scores miss the declared newcomer bound of 80\% on these months, as Sec.~\ref{sec:heldout} reports for the sealed-test score, and both keep the below-median share under 5\%. Recomputing the full score without the collaboration share changes 2.1\% of the top tenth with age and 3.0\% age-neutral, and moves the hit rate by $+0.19$ points ($-0.29$ to $0.68$) and $+0.48$ ($-0.19$ to $1.06$). The prospective window from 23 November 2026 and the blind reader study of Sec.~\ref{sec:limitations} remain the tests that can still validate these terms.

\paragraph{Report rule (post hoc).} The Kimi K3 card (Fig.~\ref{fig:cases}) exposed two gaps on collective bylines: none of its 21 catalogue names could be bound, because the byline gives no author-level affiliations, and byline positions read its surname-ordered roster as roles. On a version-verified collective byline, the rule therefore binds a uniquely resolved catalogue name when the catalogue places the person at the same parent institution as the byline, and gives every bound author on it at least the last-author weight of a small team, 0.68; lead positions and top-level members keep their larger weight, and only the strongest author counts. Binding alone, binding with this floor and binding with a floor of 1 leave every month's top-tenth hit rate unchanged, so the data cannot choose the weight; 0.68 reuses an existing constant, and Kimi K3's resulting rank played no part in the choice. Among the 17 collective-byline papers first submitted in January--October 2026 in any direction, the rule binds members on ten reports from four institutions and raises the weight of members already bound by author-level affiliations on two others. In the sixteen-direction pool it rescores ten papers, all reports and all early-citation hits, whose ranks within their month rise, for Kimi K3 from 164th to 32nd with age. With age every month keeps the same top tenth; age-neutral only June's changes, at the same hit rate. On May--August, newcomer recall stays at 0.90 with age and rises from 0.81 to 0.83 age-neutral, and the share of picks below the median relevance stays at 4.15\% with age and moves from 2.04\% to 2.11\% age-neutral. The case cards and the reading-list shares on the 27 November snapshot include the rule; the choices above and the pre-declared check do not. Names that the catalogue places elsewhere stay pending, so members who also hold another affiliation are missed, and company-led teams without a collective byline keep the ordinary rules.

%% file: experiments/semantic_appendix_v28.tex
\Needspace{8\baselineskip}
\section{Abstract evidence, semantic retrieval and temporal author context}
\label{app:semantic}
\subsection{Source fields and analysis views}
The source audit reads the same 148,150 complete-period records. Every record has a nonempty summary field, but one contains an arXiv administrative removal notice rather than a scientific abstract. Twenty-seven summaries contain fewer than thirty whitespace-delimited words, and sixteen pairs of records share an identical normalized abstract. These are source-quality flags, not grounds for silently deleting canonical records or asserting that every repeated abstract denotes the same work. The administrative record has no selected LLM-topic match. The temporal experiment excludes it and all members of the exact-abstract duplicate groups; the complete corpus accounting retains the original source population.

Of the complete-period records, 105,569 currently retain version 1. For 105,565 of these, sources are verified and the canonical first-submission and retained source-created days agree. The temporal experiment uses this subset and guards availability by the later of that day and the arXiv identifier-month start. It does not claim to reconstruct the original announcement service or the historical availability of identity bindings.

\begin{figure}[t!]\centering
\includegraphics[width=\linewidth]{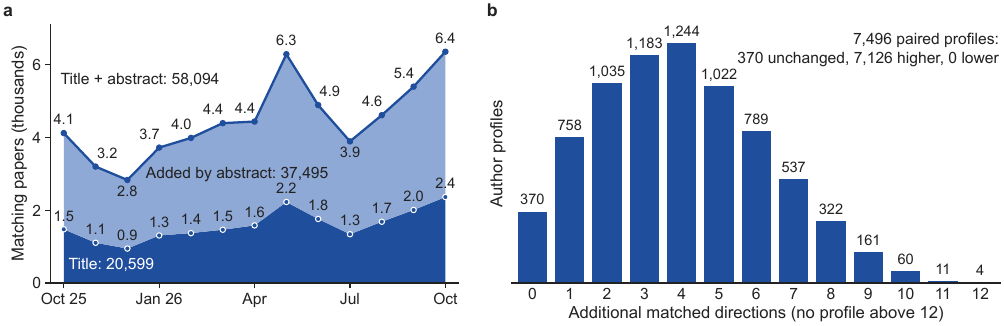}
\caption{\textbf{Title and abstract matches in a frozen library.}
(a) Monthly distinct-paper counts in the same sixteen-direction union over October 2025--October 2026; dark and light shading separate title matches from the additional abstract matches. Labels give thousands per month and the 13-month totals.
(b) The original direction-increment histogram for 7,496 profiles with at least three matched paper units under both text conditions, using an independent frozen author-profile snapshot. Bar labels give the nonzero profile counts.
Papers split one unit across their matched directions. The panels describe the topical evidence exposed by the two text views.}\label{fig:abstract-profiles}
\end{figure}
\input{experiments/abstract_detail_v30}

\input{experiments/semantic_author_v28}

\subsection{An index for reconstructing the study corpus}
The source index provides arXiv identifiers, observed versions, dates, categories, licenses, abstract checksums and both topic-membership views. Abstract text and personal identity maps are excluded. Source assertions, derived matches and expert judgments have separate fields; author and institution scores are labelled as computed quantities. Semantic arrays remain experiment artifacts, and raw-text redistribution requires a separate review of source licenses.

CSFCube supplies the independent facet judgments under its own license~\citep{mysore2021csfcube}; the annual index itself has no expert paper-quality labels. Its versioned source fields, quality flags and fixed splits support future task-specific annotation and adjudication.

%% file: experiments/abstract_detail_v30.tex
\begin{table}[tbp]\centering\small
\caption{\textbf{Paired paper and institution coverage from cached abstracts.} Paper columns use the complete first-submission cohort; institution columns use the same 114-parent, identity-anchored graph as Fig.~\ref{fig:abstract-evidence}. Adding abstracts expands eligible direction--institution pairs from 980 to 1,471, with 491 additions and no losses. Titles and abstracts are compared with identities, seed assignments, source attribution and graph edges held fixed.}\label{tab:abstract-detail}
\setlength{\tabcolsep}{4pt}\renewcommand{\arraystretch}{1.12}
\begin{tabularx}{\linewidth}{Yrrrrrr}\toprule
\rowcolor{tablehead}\textcolor{white}{\textbf{Direction}}&\textcolor{white}{\makecell{Title\\papers}}&\textcolor{white}{\makecell{+ abstract\\papers}}&\textcolor{white}{\makecell{Added\\(\%)}}&\textcolor{white}{\makecell{Title\\contexts}}&\textcolor{white}{\makecell{+ abstract\\contexts}}&\textcolor{white}{\makecell{New\\contexts}}\\\midrule
\rowcolor{tablepale}
Reasoning&2,227&\textbf{9,604}&76.8&85&\textbf{105}&20\\
RL reasoning&1,071&\textbf{4,832}&77.8&68&\textbf{97}&29\\
\rowcolor{tablepale}
Distillation&1,156&\textbf{3,105}&62.8&65&\textbf{83}&18\\
Alignment&1,312&\textbf{7,567}&82.7&67&\textbf{103}&36\\
\rowcolor{tablepale}
Agent harnesses&763&\textbf{3,828}&80.1&71&\textbf{93}&22\\
Self-improvement&1,187&\textbf{3,849}&69.2&77&\textbf{88}&11\\
\rowcolor{tablepale}
Multi-agent&973&\textbf{2,654}&63.3&37&\textbf{75}&38\\
RAG&1,718&\textbf{4,262}&59.7&48&\textbf{71}&23\\
\rowcolor{tablepale}
Pretraining&925&\textbf{4,087}&77.4&50&\textbf{92}&42\\
Efficiency&2,221&\textbf{6,848}&67.6&66&\textbf{96}&30\\
\rowcolor{tablepale}
Code LLMs&1,191&\textbf{3,830}&68.9&47&\textbf{93}&46\\
Multimodal&2,684&\textbf{9,075}&70.4&85&\textbf{99}&14\\
\rowcolor{tablepale}
Data synthesis&1,384&\textbf{6,789}&79.6&58&\textbf{98}&40\\
Evaluation&1,059&\textbf{8,436}&87.4&56&\textbf{103}&47\\
\rowcolor{tablepale}
Safety&1,650&\textbf{4,262}&61.3&57&\textbf{86}&29\\
Hallucination&1,536&\textbf{5,615}&72.6&43&\textbf{89}&46\\
\bottomrule\end{tabularx}\end{table}

%% file: experiments/semantic_author_v28.tex
\subsection{Semantic histories without verified anchors}
\label{sec:semantic-author}
\begin{table}[tbp]\centering\small
\caption{\textbf{Semantic histories on a snapshot without verified anchors.} The snapshot supplies 4,777 author queries and 4,026 later candidate papers. Candidate bylines define outcomes but are hidden from retrieval. Mean reciprocal rank (MRR) and recall are multiplied by 100; recall uses each author's known later papers. Validation fixes the mixing weight $\alpha$.}\label{tab:semantic-author}
\setlength{\tabcolsep}{5pt}\renewcommand{\arraystretch}{1.16}
\begin{tabularx}{\linewidth}{Yrrrrr}\toprule
\rowcolor{tablehead}\textcolor{white}{\textbf{Historical context}}&\textcolor{white}{$\alpha$}&\textcolor{white}{\textbf{MRR}}&\textcolor{white}{\textbf{Recall@10}}&\textcolor{white}{\textbf{Recall@50}}&\textcolor{white}{\makecell{Non-seed\\MRR}}\\\midrule
\rowcolor{tablepale}
Title history&---&4.81&7.08&14.39&4.82\\
Title + abstract history&---&\textbf{8.29}&\textbf{11.69}&\textbf{23.34}&\textbf{8.32}\\
\rowcolor{tablepale}
Coauthor mixture&0.5&8.18&11.50&22.86&8.21\\
Semantic-gated mixture&0.5&8.18&11.51&22.80&8.21\\
\rowcolor{tablepale}
Shuffled-neighbor control&0&\textbf{8.29}&\textbf{11.69}&\textbf{23.34}&\textbf{8.32}\\
\bottomrule\end{tabularx}\end{table}
The time-split comparison tests whether earlier publications represent an author's subsequent research. Recency-weighted semantic centroids rank later candidate texts by cosine similarity, with candidate bylines withheld. Observed subsequent authorship provides the continuity target; unselected papers have no independent relevance label. All conditions share the candidate pools and dates, keeping the text and coauthor comparisons paired within this source snapshot.

Validation uses history through March and 2,747 April--June candidates for 2,311 authors. Test uses 38,135 history papers through June and 4,026 July--October candidates for 4,777 authors, including 4,665 outside the seed catalogue. Records retain version 1 and verified dates; target bylines are complete and source-bound, with exact-abstract duplicate families excluded across periods. These identity bindings omit the verified anchors used in the learned-graph study in Sec.~\ref{sec:learned-graph}.

Title--abstract histories raise test MRR from \textbf{4.81 to 8.29} (Tab.~\ref{tab:semantic-author}) and authored-paper recall@50 from \textbf{14.39\% to 23.34\%}; non-seed MRR rises from 4.82 to 8.32. Ordinary and semantic-gated coauthor mixtures, with weights chosen on validation, each reach 8.18 MRR. The shuffled-neighbor control selects zero weight. Within this snapshot, richer author text accounts for the observed gain; the coauthor mixtures add no further mean-MRR improvement.

%% file: experiments/semantic_protocol_v28.tex
\subsection{Fixed semantic retrieval protocol}
BM25~\citep{robertson2009bm25} uses $k_1=1.2$, $b=0.75$, a fixed English function-word stoplist and document-frequency estimates from the 4,207 benchmark documents. Query terms are counted once. The three original SPECTER2 conditions use the same base checkpoint, without task adapters or fine-tuning, concatenating title, separator and selected abstract text. Representations use the first token, L2 normalization and cosine similarity. The maximum length is 512 tokens. Qwen3-Embedding-4B and 8B reuse the same three full/facet input definitions with the same token limit, last non-padding-token pooling and L2 normalization. Each added backbone retains all three conditions; no relevance grade fits the encoders or changes their pooling and ranking settings. We do not assume that the benchmark documents were absent from pretraining.

Query facets and predicted candidate sentence facets come from the benchmark. Background combines its background and objective labels. Across all documents, missing background, method and result spans cause 36, 305 and 337 full-abstract fallbacks respectively. Scores use the released candidate pools and adjudicated grades, with two exact self-candidates removed uniformly: query 8781666 appears in its own background and result pools. This leaves 6,242 evaluated pairs from 6,244 released judgments. Original-pool outputs for the initial lexical/SPECTER2 runs remain in the numerical supplement. No query facet requires fallback. The primary NDCG uses gain $2^{y}-1$; precision counts grades of at least two. We additionally report linear-gain NDCG at twenty percent of each candidate pool to expose the metric convention. Confidence intervals resample the 34 distinct query documents 2,000 times, keeping repeated facets together, with seed 1108.

Tabs.~\ref{tab:external-all-v29} and~\ref{tab:facet-extra} report every condition and endpoint.

\input{experiments/external_all_v29}
\input{experiments/external_extra_v29}

The paired full-abstract gains over BM25 are 7.71 points for SPECTER2, 11.87 for Qwen4B and 11.30 for Qwen8B; their query-document-cluster intervals exclude zero. The 8B-minus-4B full-abstract difference is -0.56 points and its interval includes zero. Facet-dependent improvements should likewise be read with the complete paired intervals retained in the numerical supplement.

\subsection{Temporal context construction and selection}
For each historical paper, credit decays with a 180-day half-life and is divided across the full source byline. Author centroids normalize the resulting weighted mean. Messages use historical coauthorship from papers with at most twenty authors; self-loops are omitted. Ordinary messages average neighboring centroids, while semantic gates multiply each edge by the squared nonnegative cosine between the two historical centroids. The query is the normalized mixture $(1-\alpha)h_a+\alpha m_a$. A shuffled control permutes neighbor features within historical-paper-count bins while retaining the observed graph.

\begingroup\interlinepenalty=10000
We select $\alpha\in\{0,0.1,0.25,0.5\}$ independently for each message condition by validation macro MRR, breaking ties toward zero (Tab.~\ref{tab:semantic-selection}). Ordinary and gated contexts select 0.5; the shuffled control selects zero. These choices are saved before scoring the test period. History graphs contain 336,596 and 553,996 directed coauthor associations for validation and test, respectively. The test contains 8,073 source-bound author--candidate labels. The candidate universe is restricted to the predeclared LLM directions and complete bound bylines; incomplete identities remain outside this particular proxy benchmark, rather than being interpreted as negative authorship. Unseen-name discovery and independent reading judgments are separate tasks.\par
\endgroup

\begin{table}[tbp]\centering\small
\caption{\textbf{Validation selects the context mixture before test scoring.} Entries are validation MRR $\times100$ on the same 2,311 queries. The selected coefficient for each condition is bold.}\label{tab:semantic-selection}
\begin{tabularx}{\linewidth}{Yrrrr}\toprule
\rowcolor{tablehead}\textcolor{white}{\textbf{Context}}&\textcolor{white}{$\alpha=0$}&\textcolor{white}{$0.1$}&\textcolor{white}{$0.25$}&\textcolor{white}{$0.5$}\\\midrule
\rowcolor{tablepale}
Coauthor mixture&8.765&8.825&8.779&\textbf{8.839}\\
Semantic-gated mixture&8.765&8.812&8.811&\textbf{8.849}\\
\rowcolor{tablepale}
Shuffled-neighbor control&\textbf{8.765}&8.326&7.308&4.043\\
\bottomrule\end{tabularx}\end{table}

\paragraph{Execution and reuse.}
In the initial SPECTER2 run, eight existing NVIDIA A800-SXM4-80GB GPUs encode 312,399 inputs: 148,150 title-only texts, 148,150 title--abstract texts and 16,099 distinct benchmark texts. Encoder weights are frozen and verified against the source checksum. The slowest worker finishes in 132.8 seconds, including that worker's model/input loading and output; model download, transfer, merging and readout evaluation are outside this number. Peak framework-allocated memory is 1.47 GiB per worker, and 85 inputs exceed the 512-token limit. These timings describe the SPECTER2 run. The added Qwen backbones have separately recorded encoding and merge costs; their representations are each computed once and reused across the learned-model comparisons in App.~\ref{app:learned-graph}. Ordinary KnowMe ranking remains encoder-free.

%% file: experiments/external_all_v29.tex
\begin{table}[tbp]\centering\small
\caption{\textbf{All predeclared external text conditions.} The same 50 query--facet tasks and 6,242 judged non-self pairs underlie every row. Values are NDCG@10 $\times100$ with 95\% query-document-cluster bootstrap errors. Query and candidate facets are supplied by CSFCube; the encoders receive no benchmark fitting or adaptation.}\label{tab:external-all-v29}
\setlength{\tabcolsep}{5pt}\renewcommand{\arraystretch}{1.16}
\begin{tabularx}{\linewidth}{Yrrrr}\toprule
\rowcolor{tablehead}\textcolor{white}{\textbf{Representation / input}}&\textcolor{white}{\textbf{Background}}&\textcolor{white}{\textbf{Method}}&\textcolor{white}{\textbf{Result}}&\textcolor{white}{\textbf{All facets}}\\\midrule
\rowcolor{tablepale}
BM25: title&30.88&24.40&24.84&\cival{26.62}{6.44}{6.48}\\
BM25: title + abstract&58.21&32.14&42.69&\cival{44.07}{6.65}{6.55}\\
\rowcolor{tablepale}
BM25: query facet&50.45&29.22&46.30&\cival{41.82}{5.52}{5.90}\\
SPECTER2: full abstract&61.19&40.26&54.43&\cival{51.78}{6.76}{6.65}\\
\rowcolor{tablepale}
SPECTER2: query facet&58.66&40.36&50.33&\cival{49.61}{6.48}{6.16}\\
SPECTER2: both facets&58.55&40.11&52.99&\cival{50.39}{5.11}{5.43}\\
\rowcolor{tablepale}
Qwen3-Embedding-4B: full abstract&64.89&46.24&57.20&\cival{55.94}{6.55}{6.66}\\
Qwen3-Embedding-4B: query facet&64.60&45.81&60.86&\cival{56.94}{6.24}{6.00}\\
\rowcolor{tablepale}
Qwen3-Embedding-4B: both facets&65.05&45.40&60.74&\cival{56.90}{6.18}{6.20}\\
Qwen3-Embedding-8B: full abstract&63.84&43.56&59.21&\cival{55.37}{6.75}{6.83}\\
\rowcolor{tablepale}
Qwen3-Embedding-8B: query facet&67.06&43.54&61.18&\cival{57.07}{5.92}{5.92}\\
Qwen3-Embedding-8B: both facets&\textbf{68.44}&\textbf{46.86}&\textbf{61.62}&\cival{\mathbf{58.78}}{5.33}{5.49}\\
\bottomrule\end{tabularx}\end{table}

%% file: experiments/external_extra_v29.tex
\begin{table}[tbp]\centering\small
\caption{\textbf{Additional external retrieval endpoints.} Values are multiplied by 100. All rows retain the same 50 tasks and 6,242 non-self judged pairs as Tab.~\ref{tab:external-all-v29}.}\label{tab:facet-extra}
\setlength{\tabcolsep}{5pt}\renewcommand{\arraystretch}{1.16}
\begin{tabularx}{\linewidth}{Yrrr}\toprule
\rowcolor{tablehead}\textcolor{white}{\textbf{Configuration}}&\textcolor{white}{\textbf{NDCG@20}}&\textcolor{white}{\textbf{Precision@10}}&\textcolor{white}{\makecell{Linear-gain\\NDCG@20\%}}\\\midrule
\rowcolor{tablepale}
BM25: title&30.92&16.00&35.95\\
BM25: title + abstract&47.69&28.60&52.39\\
\rowcolor{tablepale}
BM25: query facet&45.65&27.40&51.11\\
SPECTER2: full abstract&55.70&34.60&61.59\\
\rowcolor{tablepale}
SPECTER2: query facet&54.26&31.80&61.36\\
SPECTER2: both facets&53.86&33.20&61.20\\
\rowcolor{tablepale}
Qwen3-Embedding-4B: full abstract&60.98&38.00&65.80\\
Qwen3-Embedding-4B: query facet&61.41&\textbf{39.00}&66.54\\
\rowcolor{tablepale}
Qwen3-Embedding-4B: both facets&61.21&38.80&66.30\\
Qwen3-Embedding-8B: full abstract&60.06&38.60&65.10\\
\rowcolor{tablepale}
Qwen3-Embedding-8B: query facet&62.02&38.00&66.75\\
Qwen3-Embedding-8B: both facets&\textbf{64.59}&36.60&\textbf{68.65}\\
\bottomrule\end{tabularx}\end{table}

%% file: experiments/trainable_graph_protocol_v32.tex
\section{Learning from the reusable author graph}
\label{app:learned-graph}

These experiments test whether cached text and source-bound author relations support learned retrieval: recovery of later authored work and reconstruction of masked authorship. Both use the anchored identity snapshot and modeling choices from an earlier inspected study. Authorship supplies relation supervision, rather than independent reading-utility labels; the product retains the bounded, encoder-free score in Sec.~\ref{sec:system}.

Chronological retrieval asks whether an earlier research record retrieves later authored work; static masking asks whether authorship can be reconstructed from the rest of the same library. Together they separate temporal transfer from within-library completion.

\subsection{Chronological and static populations}
The chronological study uses verified version-1 records and the source-date guard in App.~\ref{app:semantic}. Identities follow the fixed current mapping, so historical text windows do not reconstruct historically available identity information. Four training windows target December 2025 and January--March 2026, using earlier papers as context; they provide 3,452 query--window records and 3,774 author--paper relations. Validation targets April--June with context through March. Test targets July--October with context through June: 8,936 authors query 3,549 candidates with 11,637 observed positive relations. Test context contains 38,656 papers and 522,918 directed coauthor-message edges. The analysis includes one- and two-paper histories; requiring at least three context papers retains 3,804 authors. The extension reuses the earlier temporal cohort and its inspected baselines; subsequent model selection follows the fixed validation rule.

The static task uses the thirteen-month library without restricting papers to version 1. Administrative placeholders and both members of exact-abstract duplicate families are excluded. A fixed arXiv-ID hash assigns nominal 80\% training, 10\% validation and 10\% test partitions. Four training folds rotate masked targets, with the other three folds providing context. Validation and test use all training papers as context; neither contributes context edges or training targets. Multiple versions retain one paper identity.

The static population has 104,908 author nodes and 146,393 distinct training author--paper relations across 99,963 query--fold records. Final context includes 118,324 papers, 76,820 with bound authors, and 935,262 directed message edges. Validation has 15,457 queries and 5,838 candidates; test has 15,694 queries, 5,797 candidates and 19,374 observed relations. Authors may recur across folds or share targets. Partial bylines retain resolved positions, and candidates without observed query labels remain in the retrieval pool.

\subsection{Semantic histories and differentiable message passing}
Let $\mathbf e_p$ be a normalized representation of a paper's title and abstract. For author $a$, the historical or unmasked context gives
\begin{equation}
\mathbf x_a=\PaperNormHistory,
\end{equation}
where $d_p$ is the fixed 180-day decay and $|\mathcal A_p|$ is the recorded byline size. Static decay is measured from 31 October 2026. Papers with more than twenty authors contribute available byline evidence to author context but create no coauthor-message edges. Neighbour weights are normalized per receiving author.

Following inductive neighbourhood aggregation~\citep{hamilton2017graphsage}, the models transform semantic features without trainable author-ID vectors. One- and two-layer variants learn message and neighbour transformations with residual connections, Gaussian error linear unit (GELU) activations and layer normalization (LayerNorm). Gated variants apply a learned sigmoid to projected sender and receiver features and renormalize incoming mass; both operations remain differentiable. The matched no-graph model uses two residual MLP blocks. Candidates receive a separate projection of their observed title and abstract; structured bylines and affiliations are withheld.

For query $a$ with observed positives $\mathcal P_a$ and the complete candidate set $\mathcal C$, training minimizes
\begin{equation}
\mathcal L_a=\log\sum_{p\in\mathcal C}\exp(s_{ap}/\tau)
-\log\sum_{p\in\mathcal P_a}\exp(s_{ap}/\tau),
\end{equation}
with temperature $\tau=0.07$. Each query contributes one mean-loss term. Known positives are pooled rather than treated as mutual negatives, while every candidate enters the denominator. The objective learns compatibility with observed authorship.

\subsection{Backbones, controls and selection}
Frozen SPECTER2-base and Qwen3-Embedding-4B/8B~\citep{singh2023scirepeval,zhang2025qwenembedding} use the same paper IDs and input views, with dimensions 768, 2,560 and 4,096. Each backbone encodes 312,399 views: title-only and title--abstract representations of 148,150 papers plus 16,099 external-benchmark views. Qwen uses left padding, the last non-padding token, vector normalization, a 512-token limit and bfloat16 computation. Only the graph encoder and retrieval projections are trained.

Seven learned-projection families comprise title and abstract MLPs, one- and two-layer mean aggregation, learned-gate counterparts, and a two-layer rewired-graph control. Endpoint swaps change graph structure while preserving degrees, semantic features and the edge-weight multiset, without self loops or duplicate pairs. Each family uses four initializations (1109--1112), two learning rates ($10^{-4}$ and $3\!\times\!10^{-4}$), latent width 256, dropout 0.1, AdamW weight decay $10^{-4}$, gradient clipping at 1, at most eighty epochs and validation patience twelve.

Two semantic-preservation controls concatenate $\sqrt{0.9}\,\mathbf x$ and $\sqrt{0.1}\,\mathbf z$, where $\mathbf z$ is the normalized learned representation. Their dot product is exactly 90\% original semantic cosine plus 10\% learned cosine. This mixture was specified before the original static test access and is shared by a residual MLP and residual two-layer graph model; neither validation nor test outcomes select its weight. We compare learned projections separately.

Validation MRR selects the checkpoint and learning rate per family and seed, breaking ties by earlier checkpoint and smaller learning rate. The family choice uses the mean over four selected seeds. Static test labels are withheld until all backbone and model selections are fixed. Nested hash-selected supervision subsets of 25\%, 50\% and 100\% vary the number of labeled query authors while keeping the complete context graph and candidate pool fixed. Inference-only controls remove all message edges or deterministically mask 30\% of message-edge endpoint units. Symmetric pairs are masked together before row renormalization. These controls require no refitting; all families and initializations remain in the supplement.

Training and testing use sixteen NVIDIA A800 80\,GB GPUs and reuse all three backbones' encodings. There are 288 graph/readout fits: sixty chronological and 228 static, including sixteen additional supervision-scale fits overall. The 100\% scale point reuses the primary fit for initialization 1109. Foundation encoders, including the 8B model, remain frozen.

\subsection{Metrics and dependent observations}
We report macro MRR, recall at 10, 50 and 100, and binary NDCG at 10 and 50, breaking score ties by candidate-column order. We report sample SD across four fitted seeds. Marginal readouts retain non-catalogue authors, training-query membership, and histories of one, two, three--four, five--nine and at least ten papers.

Shared targets couple author queries. The chronological query--positive-paper graph has 1,699 connected components, the largest containing 3.03\% of queries. Component-resampled differences are conditional on fitted models and observed labels; historical graph dependence can remain. A prespecified sensitivity excludes a query if any known positive target has more than twenty authors, retaining the candidate pool and rankings. It excludes no chronological query because those target bylines already meet this limit. The same rule precedes static test-label access.

\input{experiments/learned_details_v32}

%% file: experiments/learned_details_v32.tex
\subsection{Chronological retrieval and supervision coverage}
\input{experiments/learned_temporal_table_v32}
The chronological task provides a stricter transfer setting, with only 3,774 observed training relations across 3,452 query--window records. The validation-selected one-layer gated graph reaches \pmval{5.65}{0.29} MRR, compared with 13.88 for frozen abstract histories on the same candidate pool (Tab.~\ref{tab:learned-temporal}). Its mean difference from the abstract MLP is \cival{0.258}{0.154}{0.157} MRR points under the conditional outcome-component bootstrap. Thus the learned relation model can differ from its matched neural control while still losing substantial information supplied by the frozen semantic history. The static residual controls examine this distinction directly.

The retained optimization traces cover all 56 primary chronological fits and 2,620 completed epochs. Across the four selected gated runs, mean training loss falls from 6.017 at the first epoch to 3.327 at the last. All four continue lowering training loss after their best validation checkpoint while validation retrieval declines. Every test readout uses the saved validation-selected checkpoint.

\subsection{Static subgroups, structure and target-team sensitivity}
\input{experiments/learned_strata_table_v32}
The static task supplies more observed training relations while keeping paper-level splits fixed across backbones. Validation selects the one-layer mean model for SPECTER2 and the residual two-layer model for both Qwen backbones. Tab.~\ref{tab:learned-static} retains all tested families. Semantic preservation benefits the Qwen comparisons, but the SPECTER2 residual graph remains below its frozen abstract baseline. The retained controls therefore support a specific combination of representation and readout rather than a universal benefit from message passing.

Outside the archived seed catalogue (Tab.~\ref{tab:learned-strata}), 15,576 query authors obtain 17.85 MRR with the 8B residual graph, compared with 15.94 for frozen abstract histories. Authors absent from supervised query rows remain in the reported groups, although some occur as context-graph nodes. Group scores summarize retrieval difficulty for the observed targets in each stratum.

The static outcome graph has 2,195 query-bearing components; the largest contains 5,951 queries, or 37.92\% of the test population. Its query-size effective component count is 6.94. Two thousand component-bootstrap draws give conditional MRR differences of \cival{1.913}{0.263}{0.363} points over frozen abstracts and \cival{-0.014}{0.205}{0.110} over the matched residual MLP. The latter interval crosses zero. These intervals describe the coupled observed outcomes; historical context can introduce additional dependence beyond the target-paper components.

The predeclared target-team check excludes 598 complete queries with a known target containing more than twenty byline positions. It retains 15,096 queries and 18,456 positive relations against the original 5,797 candidates, with rankings and retained labels unchanged. Residual graph MRR is 18.10, compared with 16.16 for frozen abstracts and 18.10 for the residual MLP. The 598 excluded queries remain separately reported: graph and MLP MRR are 10.27 and 10.67. The full population remains the primary result.

\subsection{Reusable study artifacts}
The accompanying index retains all 148,150 arXiv identifiers and observed versions, source dates, categories, recorded licenses, text checksums and fixed paper partitions. The numerical supplement contains every fitted seed, method, declared subgroup, sensitivity and scale point, together with source and encoding hashes. Local prepared arrays retain sparse context, message and label matrices with candidate/query indices. The distributable package is an index and numerical supplement; it excludes raw abstracts, model weights, browser state and personal-name lookup. Integer identity clusters are not guaranteed anonymous, and an absent authorship link is not an adjudicated negative relevance label.

Corrected graph bindings leave the original text IDs, bytes and order unchanged, so all 312,399 encoded views per backbone can be reused. Reproduction requires matching source versions and text; a missing historical abstract is not filled with a newer version.

%% file: experiments/learned_temporal_table_v32.tex
\begin{table}[tbp]\centering\small
\caption{\textbf{Chronological retrieval with learned histories.} All methods use frozen SPECTER2 features, 8,936 query authors and the same 3,549 test candidates. Metrics are multiplied by 100; grey $\pm$ denotes sample SD over four fitted seeds. Gated GNN (1) was selected by validation MRR. Frozen histories have no fitted-seed dispersion.}\label{tab:learned-temporal}
\begin{tabularx}{\linewidth}{Yrrrrr}\toprule
\rowcolor{tablehead}\textcolor{white}{\textbf{Readout}}&\textcolor{white}{MRR}&\textcolor{white}{R@10}&\textcolor{white}{R@50}&\textcolor{white}{R@100}&\textcolor{white}{NDCG@50}\\\midrule
\rowcolor{tablepale}
Frozen title&\nopm{8.34}&\nopm{12.23}&\nopm{21.67}&\nopm{27.69}&\nopm{10.53}\\
Frozen abstract&\nopm{\mathbf{13.88}}&\nopm{\mathbf{19.69}}&\nopm{\mathbf{32.53}}&\nopm{\mathbf{40.42}}&\nopm{\mathbf{17.02}}\\
\midrule
\rowcolor{tablepale}
Title MLP&\pmval{2.52}{0.28}&\pmval{4.31}{0.39}&\pmval{12.14}{1.22}&\pmval{18.00}{1.64}&\pmval{4.10}{0.43}\\
Abstract MLP&\pmval{5.39}{0.29}&\pmval{9.42}{0.41}&\pmval{21.93}{1.45}&\pmval{30.37}{2.03}&\pmval{8.26}{0.48}\\
\rowcolor{tablepale}
Mean GNN (1)&\pmval{5.65}{0.29}&\pmval{9.80}{0.54}&\pmval{23.07}{1.04}&\pmval{31.58}{1.39}&\pmval{8.67}{0.38}\\
Mean GNN (2)&\pmval{5.60}{0.18}&\pmval{9.98}{0.56}&\pmval{23.16}{1.12}&\pmval{31.48}{1.39}&\pmval{8.65}{0.31}\\
\rowcolor{tablepale}
Gated GNN (1)&\pmval{5.65}{0.29}&\pmval{9.80}{0.54}&\pmval{23.07}{1.04}&\pmval{31.58}{1.39}&\pmval{8.67}{0.38}\\
Gated GNN (2)&\pmval{5.66}{0.29}&\pmval{10.07}{0.25}&\pmval{23.43}{0.33}&\pmval{32.42}{0.39}&\pmval{8.73}{0.27}\\
\rowcolor{tablepale}
Rewired GNN (2)&\pmval{5.60}{0.25}&\pmval{9.83}{0.37}&\pmval{23.40}{0.59}&\pmval{32.31}{0.45}&\pmval{8.69}{0.30}\\
\bottomrule\end{tabularx}\end{table}

%% file: experiments/learned_strata_table_v32.tex
\begin{table}[tbp]\centering\small
\caption{\textbf{Static 8B readouts across declared author groups.} MRR is multiplied by 100, with sample SD over four fitted seeds. The blocks show alternative partitions of the same test population by catalogue membership, supervised-query status and history size. ``Other query authors'' were absent from supervised training-query rows, but can occur as context-graph nodes.}\label{tab:learned-strata}
\begin{tabularx}{\linewidth}{Yrrrr}\toprule
\rowcolor{tablehead}\textcolor{white}{\textbf{Author group}}&\textcolor{white}{Queries}&\textcolor{white}{Frozen abstract}&\textcolor{white}{Residual MLP}&\textcolor{white}{Residual GNN}\\\midrule
\rowcolor{tablepale}
Catalogue seeds&118&8.82&\pmval{\mathbf{12.52}}{0.58}&\pmval{11.67}{0.60}\\
Outside catalogue&15,576&15.94&\pmval{\mathbf{17.86}}{0.15}&\pmval{17.85}{0.07}\\
\midrule
\rowcolor{tablepale}
Supervised query authors&10,484&12.15&\pmval{\mathbf{14.07}}{0.10}&\pmval{\mathbf{14.07}}{0.07}\\
Other query authors&5,210&23.42&\pmval{\mathbf{25.36}}{0.26}&\pmval{25.33}{0.09}\\
\midrule
\rowcolor{tablepale}
One context paper&3,560&24.36&\pmval{\mathbf{26.31}}{0.30}&\pmval{26.25}{0.15}\\
Two context papers&2,722&19.06&\pmval{20.44}{0.07}&\pmval{\mathbf{20.57}}{0.02}\\
\rowcolor{tablepale}
Three--four&3,341&14.92&\pmval{16.85}{0.07}&\pmval{\mathbf{16.98}}{0.08}\\
Five--nine&3,507&12.04&\pmval{\mathbf{14.15}}{0.23}&\pmval{14.09}{0.13}\\
\rowcolor{tablepale}
At least ten&2,564&7.30&\pmval{\mathbf{9.53}}{0.13}&\pmval{9.30}{0.06}\\
\bottomrule\end{tabularx}\end{table}

%% file: experiments/data_value_appendix_v30.tex
\FloatBarrier
\Needspace{12\baselineskip}
\section{Semantic readouts and source annotations}
\label{app:data-value-v30}
\subsection{Expert retrieval protocol and complete endpoints}
The CSFCube comparison uses its expert relevance grades~\citep{mysore2021csfcube}, all 50 query--facet tasks and 34 query documents. Excluding two self-pairs leaves 6,242 candidate pairs per condition. Frozen Qwen3-Embedding-8B similarity~\citep{zhang2025qwenembedding} is compared with a fixed Qwen3.8-27B relevance scorer under matched text conditions. Prompts, conditions and token caps are fixed before inference, and the scorer receives no grades. This fixed-readout comparison extends the earlier benchmark and embedding analysis.

The relevance score is the log-sum-exp of \texttt{Yes}/\texttt{yes} logits minus that of \texttt{No}/\texttt{no}. Title-only uses both titles; full uses both title--abstract texts; query-facet combines the supplied sentence-facet query with the full candidate, falling back to the full query when its facet view is empty. Each side is capped at 512 tokens. Descending scores rank candidates, with ascending string paper ID breaking ties. Scores are not calibrated quality probabilities.

NDCG@10 and NDCG@20 use gain $2^g-1$ for expert grade $g$; P@10 thresholds at $g\geq2$. The deeper-pool endpoint uses linear gain at $\max(1,\lfloor0.2n\rfloor)$, where $n$ excludes self-pairs. Metrics average fifty task values (Tab.~\ref{tab:data-value-expert-v30}). For 2,000 bootstrap draws (seed 1108), query documents are resampled with all their facets together, yielding intervals conditional on the fixed pools, scorers and study choices.

\begin{table}[tbp]\centering\small
\caption{\textbf{Primary-model and frozen-reference endpoints.} Values are multiplied by 100. NDCG@10 reports a symmetric envelope of its conditional 95\% cluster interval; the other columns are task-macro means. Qwen3.8-27B is a relevance scorer and Qwen3-Embedding-8B uses frozen embedding similarity. All rows use the same judged pools.}\label{tab:data-value-expert-v30}
\setlength{\tabcolsep}{4.0pt}\renewcommand{\arraystretch}{1.14}
\begin{tabularx}{\linewidth}{Yrrrr}\toprule
\rowcolor{tablehead}\textcolor{white}{\textbf{Readout / input}}&\textcolor{white}{\textbf{NDCG@10}}&\textcolor{white}{\textbf{NDCG@20}}&\textcolor{white}{\textbf{P@10}}&\textcolor{white}{\makecell{Linear NDCG\\at 20\%}}\\\midrule
\rowcolor{tablepale}
27B relevance: title&\cival{41.22}{5.84}{5.74}&47.62&28.40&54.17\\
27B relevance: full&\cival{59.80}{5.77}{6.11}&63.92&\textbf{42.80}&68.88\\
\rowcolor{tablepale}
27B relevance: query facet&\cival{\mathbf{61.51}}{4.88}{5.30}&\textbf{65.24}&42.60&\textbf{70.18}\\
8B embedding: full&\cival{55.37}{6.75}{6.83}&60.06&38.60&65.10\\
\rowcolor{tablepale}
8B embedding: query facet&\cival{57.07}{5.92}{5.92}&62.02&38.00&66.75\\
BM25: title&\cival{26.62}{6.44}{6.48}&30.92&16.00&35.95\\
\bottomrule\end{tabularx}\end{table}

Query-facet scorer--embedding differences, in metric points ($\times100$), are \cival{4.45}{4.12}{4.62} for NDCG@10, \cival{3.22}{3.29}{3.59} for NDCG@20, \cival{4.60}{3.62}{4.04} for P@10 and \cival{3.43}{3.10}{3.41} for linear NDCG@20\%-pool. The title--abstract NDCG@10 difference is \cival{4.43}{5.59}{5.53}. Grey $\pm$ values enclose the conditional 95\% intervals. These comparisons jointly vary model architecture, size and readout. The numerical supplement provides all four metrics and matched contrasts, with task and facet results.

Mean probability mass on the four label tokens is 0.998981, 0.997717 and 0.997789 for title, full and query-facet input; minima are 0.992036, 0.980505 and 0.988242. No pair falls below 0.95. No query is clipped; in each full-candidate condition, 53 of 6,242 pairs, covering 33 documents, exceed the 512-token cap. Token mass measures readout concentration, not calibration.

\subsection{Stratified model-derived topic annotations}
The three sampling strata distinguish topical support among lexical hits, relevant content exposed only by abstracts, and semantic evidence missed by both lexical views. With unequal sampling, separate denominators distinguish coverage added by abstracts from coverage identified by semantic annotation.

The local sample contains 2,560 topic--paper rows on 2,529 versioned papers (Tab.~\ref{tab:data-value-topic-v30}). Each of sixteen topics contributes 40 title matches, 80 abstract-only matches and 40 no-lexical-match papers selected by fixed hash order from 148,117 eligible papers per topic. Qwen3.8-27B receives the abstract and topic definition but not the stratum. It assigns \texttt{primary}, \texttt{secondary}, \texttt{mention} or \texttt{none}, with an exact supporting quote required for primary/secondary. These are model annotations, not expert quality labels.

All outputs pass whole-response JSON validation, which rejects duplicate keys and nonstandard constants and requires exactly typed \texttt{relation} and \texttt{evidence} fields. Of 1,267 primary/secondary decisions, 1,262 include a nonempty literal substring of the supplied abstract and five do not; no Unicode or whitespace normalization is applied. All inputs fit the 1,024-token cap.

\begin{table}[tbp]\centering\small
\caption{\textbf{Sample composition and source-quote acceptance.} Primary/secondary columns count the model's raw relation choices. The final column counts only choices with a literal source quote. Each percentage describes the sample under its stated allocation.}\label{tab:data-value-topic-v30}
\setlength{\tabcolsep}{4pt}\renewcommand{\arraystretch}{1.13}
\begin{tabularx}{\linewidth}{Yrrrrrr}\toprule
\rowcolor{tablehead}\textcolor{white}{\textbf{Stratum}}&\textcolor{white}{\textbf{$n$}}&\textcolor{white}{\textbf{Primary}}&\textcolor{white}{\textbf{Secondary}}&\textcolor{white}{\textbf{Mention}}&\textcolor{white}{\textbf{None}}&\textcolor{white}{\textbf{Supported}}\\\midrule
\rowcolor{tablepale}
Title match&640&365&109&15&151&472/640\\
Abstract-only&1280&400&356&98&426&753/1280\\
\rowcolor{tablepale}
No lexical match&640&20&17&15&588&37/640\\
\bottomrule\end{tabularx}\end{table}

The stratified numerical results report all forty-eight topic--stratum cells and their population denominators. The 40/80/40 allocation oversamples lexical matches, so pooled unweighted rates do not estimate library prevalence. The five quotation failures remain included. Literal matching verifies source occurrence; the relation labels are the models' semantic judgments.

\paragraph{Portable annotation index.}
The portable annotation index contains all 2,560 rows, including invalid and unsupported judgments, with versioned IDs, topic/relation labels, strata, validity flags, evidence spans and SHA-256 hashes of abstracts and evidence. Spans give every literal occurrence as zero-based, half-open Unicode offsets into the exact abstract. The index excludes source text, titles, names, browser state and weights. Annotations are model-derived; CSFCube retains its CC BY-NC 4.0 notice, and source articles and models retain their respective licenses. No blanket redistribution license is assigned. The spans support checking evidence against separately obtained articles without redistributing their text.

\input{experiments/data_value_comparator_v30}

The numerical supplement reports the earlier authorship-projection transfer comparison, whose fitted projections did not improve the primary title--abstract facet-retrieval endpoint.

%% file: experiments/data_value_comparator_v30.tex
\subsection{Matched fixed-model comparison and semantic-label agreement}
\label{app:data-value-comparator-v30}
The Qwen3-32B comparator~\citep{qwen3_32b2025} was fixed before inspecting the primary strong-model outcome metrics. It uses the same 21,286 input IDs and strings, prompts, text conditions, document caps and decoding settings to produce 18,726 pair scores and 2,560 topic annotations. No model, prompt, threshold or condition was selected from these outcomes.

\begin{table}[tbp]\centering\small
\caption{\textbf{Complete fixed Qwen3-32B readouts.} The same fifty tasks, 34 query papers and 6,242 non-self pairs underlie every condition. Values are multiplied by 100. NDCG@10 reports symmetric envelopes of conditional 95\% cluster intervals; other columns are task-macro means.}\label{tab:data-value-comparator-v30}
\setlength{\tabcolsep}{4pt}\renewcommand{\arraystretch}{1.14}
\begin{tabularx}{\linewidth}{Yrrrr}\toprule
\rowcolor{tablehead}\textcolor{white}{\textbf{Qwen3-32B input}}&\textcolor{white}{\textbf{NDCG@10}}&\textcolor{white}{\textbf{NDCG@20}}&\textcolor{white}{\textbf{P@10}}&\textcolor{white}{\makecell{Linear NDCG\\at 20\%}}\\\midrule
\rowcolor{tablepale}
Title&\cival{41.13}{5.85}{5.95}&47.44&27.20&52.60\\
Full&\cival{\mathbf{52.97}}{5.89}{6.27}&\textbf{58.26}&\textbf{37.00}&\textbf{63.55}\\
\rowcolor{tablepale}
Query Facet&\cival{52.25}{5.13}{5.25}&56.95&36.40&61.90\\
\bottomrule\end{tabularx}\end{table}

\begin{table}[tbp]\centering\small
\caption{\textbf{Matched Qwen3-32B minus Qwen3.8-27B differences.} All values are metric points ($\times100$), with conditional 95\% query-document cluster deviations. Negative values favour the fixed 27B readout on these pools; title-only differences remain uncertain. The comparison changes the fixed model/readout, not the input IDs or judged candidate pools.}\label{tab:data-value-matched-v30}
\setlength{\tabcolsep}{3.5pt}\renewcommand{\arraystretch}{1.14}
\begin{tabularx}{\linewidth}{Yrrrr}\toprule
\rowcolor{tablehead}\textcolor{white}{\textbf{Input}}&\textcolor{white}{\textbf{NDCG@10}}&\textcolor{white}{\textbf{NDCG@20}}&\textcolor{white}{\textbf{P@10}}&\textcolor{white}{\makecell{Linear NDCG\\at 20\%}}\\\midrule
\rowcolor{tablepale}
Title&\cival{-0.09}{3.45}{4.16}&\cival{-0.17}{3.28}{3.51}&\cival{-1.20}{3.52}{4.00}&\cival{-1.57}{3.50}{3.70}\\
Full&\cival{-6.84}{3.64}{3.23}&\cival{-5.66}{3.65}{3.40}&\cival{-5.80}{2.98}{2.80}&\cival{-5.32}{3.03}{2.89}\\
\rowcolor{tablepale}
Query Facet&\cival{-9.26}{3.25}{3.36}&\cival{-8.29}{3.10}{3.29}&\cival{-6.20}{3.03}{2.75}&\cival{-8.28}{2.99}{2.92}\\
\bottomrule\end{tabularx}\end{table}

Relative to the matching 8B embedding reference (Tabs.~\ref{tab:data-value-comparator-v30} and~\ref{tab:data-value-matched-v30}), Qwen3-32B has NDCG@10 differences of \cival{-2.41}{6.74}{6.34} points for titles plus abstracts and \cival{-4.82}{4.88}{4.90} for query-facet input. Adding abstracts within this comparator gives \cival{11.84}{5.86}{6.54} points; changing the full abstract to the supplied query facet gives \cival{-0.72}{3.71}{3.74} points. All four metrics and all declared comparisons remain in the numerical supplement. This model dependence constrains claims about downstream reuse; it is not an evaluation of the encoder-free product scoring rule.

The comparator's four label tokens have mean probability mass 0.999998, 0.999989 and 0.999992 in title/full/query-facet conditions. No query is clipped; each full-candidate condition clips 55 pair rows from 34 distinct candidate documents, compared with 53 rows from 33 documents for the primary model. The underlying strings and caps match, but model tokenizers differ.

\paragraph{Payload extraction and semantic agreement.}
The primary model returns bare JSON; Qwen3-32B places each JSON object inside a canonical Markdown fence. A separate payload analysis removes that wrapper before applying the same duplicate-key, field-type and literal-quote checks. All 2,560 extracted objects satisfy the schema. Comparator labels are primary 901, secondary 329, mention 134 and none 1,196; 1,133 of 1,230 primary/secondary choices contain literal source evidence, leaving 97 unavailable required quotes. Original serialization and extracted-payload results are recorded separately. Neither model's topic abstract is truncated.

The uniquely parsed payload relation labels agree on 1,918 of 2,560 rows (74.9\%; Tab.~\ref{tab:data-value-label-agreement-v30}), with nominal Cohen's $\kappa=0.616$. Quote-supported primary/secondary proposals overlap on 998 rows; 264 are supported only by Qwen3.8-27B, 135 only by Qwen3-32B, and 1,163 by neither. This binary proposal flag agrees on 2,161 rows (84.4\%); the positive-set Jaccard overlap is 0.714. These are descriptive agreements on the matched, unequally stratified sample, not expert accuracy or population prevalence.

\begin{table}[tbp]\centering\small
\caption{\textbf{Matched topic-relation labels.} Rows are Qwen3.8-27B choices and columns are Qwen3-32B choices on the same 2,560 topic--paper rows. The 32B choices come from the extracted JSON payload within its Markdown fence. Bold marks matching labels.}\label{tab:data-value-label-agreement-v30}
\setlength{\tabcolsep}{8pt}\renewcommand{\arraystretch}{1.14}
\begin{tabularx}{\linewidth}{Yrrrr}\toprule
\rowcolor{tablehead}\textcolor{white}{\textbf{27B $\backslash$ 32B}}&\textcolor{white}{\textbf{Primary}}&\textcolor{white}{\textbf{Secondary}}&\textcolor{white}{\textbf{Mention}}&\textcolor{white}{\textbf{None}}\\\midrule
\rowcolor{tablepale}
Primary&\textbf{675}&47&8&55\\
Secondary&185&\textbf{176}&49&72\\
\rowcolor{tablepale}
Mention&10&32&\textbf{42}&44\\
None&31&74&35&\textbf{1025}\\
\bottomrule\end{tabularx}\end{table}

Comparator quote-supported sampled counts are 406/640 title matches, 682/1,280 abstract-only matches and 45/640 no-lexical-match rows. All 48 topic--stratum cells and every paired row remain available. A second text-free index pairs the same 2,560 source rows and records both models' relations, format flags, exact quote spans and hashes. The original primary-only index remains unchanged. Format validity, quote occurrence and relation agreement are separate fields.

%% file: experiments/reading_choice_joint_protocol_v32.tex
\input{experiments/reading_choice_qwen_protocol_v32}

\begingroup
\subsection{Matched Jev and Qwen model readouts}
\label{app:reading-choice-joint-v32}
\input{experiments/reading_choice_joint_arm_table_v32}
The same reading lists are evaluated by two model instruments (Tab.~\ref{tab:reading-choice-joint-arms-v32}). Jev returns fractional relevance (0--3), methodological-specificity (0--2) and reusable-reference (0--2) scores, with an insufficient-abstract option. Returned distributions and confidence fields are retained but are not calibrated against independent labels here. Substantive relevance requires a modal category of at least two; ties take the lower category. Jev reference-positive decisions require a modal category of at least one, while Qwen uses a separate 0--3 reference rubric. Reference scores are not pooled across these scales.

All 10,240 Jev items have one valid assessment per item. They use 13,541,716 input and 1,003,520 output tokens; at the recorded tariff of USD\,0.042 per million input tokens and free output, the estimated cost is USD\,0.57. Median request latency is 962\,ms and the 95th percentile is 1,399\,ms; Qwen batch timings are reported separately from per-request API latency.

At fifty reads, Jev's full-context minus fixed-penalty-baseline difference is \cival{-0.019}{0.038}{0.040} for relevance and \cival{0.006}{0.011}{0.011} for reusable reference. Against the fixed author reference, the differences are \cival{0.000}{0.028}{0.029} and \cival{0.007}{0.008}{0.009}, respectively. None of the sixteen per-direction relevance differences against either arm has a 95\% interval that excludes zero, so at fifty reads the context arms of this additive design left judged topical relevance unchanged within the precision of the pool (Tab.~\ref{tab:reading-choice-joint-arms-v32}). At 10, 20 and 50 reads, the six arms' mean grades stay between 2.57 and 2.72 for relevance and between 1.88 and 1.90 for reusable reference; the lowest relevance means, 2.57 at ten reads, belong to the institution-context and full-context arms. Conditional intervals use 2,000 paper-ID cluster draws (seed 1111), sharing each paper's weight across directions and arms. Qwen bounds instead describe missingness. The supplement reports all fifteen arm pairs, sixteen directions and three budgets with their intervals, preserving the primary and adjacent contrasts.

The 1,199-item mandatory union supplies 1,133 matched relevance labels: exact agreement is 69.8\% and quadratic-weighted $\kappa=0.576$. The 1,024-item random crosscheck has 983 usable pairs, 67.4\% agreement and $\kappa=0.728$; inclusion weighting to the 10,240-item pool gives 67.8\% and $\kappa=0.735$. These describe agreement on common relevance categories. Because papers matching none of the sixteen directions are outside the source frame, the estimates do not describe full-library precision.

\paragraph{Full-pool assessment and sampling coverage.}
A subsequent assessment applies the same Qwen checkpoint, prompt and validation rules to the remaining 8,273 items. The original 1,967 responses and labels are reused unchanged, preserving the primary top-fifty results and intervals. This extends descriptive coverage to the full pool without redefining the primary sample.

All 10,240 Qwen responses parse as strict JSON, but 261 violate schema/rubric rules and 134 fail literal-evidence checks, leaving 9,845 usable judgments. The 395 unavailable items include the original 96 failures. Of 24,502 nonempty quotations, including those in schema-invalid outputs, 24,340 occur exactly in their abstracts. This validates source occurrence only.

Among 9,845 usable Jev--Qwen pairs, four-category agreement is 6,761/9,845 (68.7\%), quadratic-weighted $\kappa=0.757$, and agreement on substantive relevance versus other categories is 8,726/9,845 (88.6\%). With all 395 unavailable labels, exact agreement is bounded by 66.0\% and 69.9\% over 10,240 items. The largest disagreement is 1,517 items rated three by Jev and two by Qwen; all confusion cells and directions are reported.

Population sizes and inclusion probabilities are recorded for all ninety-six sampling cells, with 640 items per direction. Inverse-inclusion weighting gives available-pair agreement of 75.6\% and quadratic-weighted $\kappa=0.642$ over the supported frame: 929,264 direction--paper pairs involving 58,079 papers that match at least one direction. Far nonmatches have zero inclusion probability, so these model-agreement estimates do not measure quality or precision across the 148,117-paper library.

\par
\endgroup

%% file: experiments/reading_choice_qwen_protocol_v32.tex
\subsection{Fixed-budget model readouts and unavailable labels}
\label{app:reading-choice-qwen-v32}
The candidate frame contains 148,150 papers first submitted from October 2025 through October 2026; excluding one administrative record and 32 exact duplicates leaves 148,117. Native matching yields 88,619 direction--paper pairs from 58,079 papers across sixteen directions. All arms share each direction's eligible population, without padding lists with unrelated papers.

Six additive arms start with content/frontier scores and recency, solo-author and venue penalties (A), then add catalogue authors (B), leave-author-out personal weights (C), institutional context (D), and network-author evidence for full context (E). The fixed-threshold full-context control (F) changes only the author reference. These penalties are computed before arm construction and reused unchanged across arms, including catalogue-dependent aging and single-author exemptions. The baseline is therefore labelled \emph{Fixed-penalty baseline}; the contrasts isolate additions from explicit contextual score terms conditional on those penalties. Primary contrasts are E--A and E--F; adjacent secondary contrasts are B--A, C--B, D--C and E--D. All arms and directions are reported.

The 10,240-item pool has 640 items per direction: 400 native matches covering the six top-50 lists, and 240 nearby nonmatches. Nearby nonmatches provide a contrasting input stratum. Fixed-hash strata retain their population sizes and inclusion probabilities. An initial crosscheck sampled 1,024 items, or 64 per direction across six strata. Before Qwen evaluation, 943 additional mandatory-list items completed a 1,967-item union spanning 1,713 papers. Selection used existing ranks rather than Jev judgments, although Jev evaluation had begun; the initial crosscheck remains separately identifiable.

Qwen3.8-27B uses greedy decoding with thinking disabled and a fixed quote-based rubric. Primary, secondary, mention and none map to relevance 3, 2, 1 and 0; reference cues use a separate 0--3 grade. Required quotations must occur literally in the abstract. Of 1,967 responses, 1,871 pass, 64 violate schema/rubric rules and 32 fail literal-evidence checks. The mandatory union has 1,133 usable judgments among 1,199 items; all 66 unavailable labels remain in the denominator. Literal quotation checks establish source occurrence; relation labels are supplied by the model.

\paragraph{Shared-item cancellation.}
Let $q$ identify a paper--topic item, $0\leq y_q\leq M$ its criterion value, and $\mathcal L_{ad}$ arm $a$'s $B$-item list in direction $d$, with budget $B=50$. For the equally weighted macro contrast over $D=16$ directions,
\begin{equation}
 c_q=\frac{1}{D}\sum_{d=1}^{D}\left(\frac{\mathbf{1}\{q\in\mathcal L_{ad}\}}{B}-\frac{\mathbf{1}\{q\in\mathcal L_{bd}\}}{B}\right),
 \qquad \Delta_{a,b}=\sum_q c_qy_q .
\end{equation}
Write $K$ and $U$ for known and unavailable effective labels and $S=\sum_{q\in K}c_qy_q$. The complete range consistent with the unavailable labels is
\begin{equation}
 L=S+M\!\sum_{q\in U:c_q<0}c_q,\qquad H=S+M\!\sum_{q\in U:c_q>0}c_q .
\end{equation}
Here $L$ and $H$ are the lower and upper limits.
A shared unavailable item has the same unknown value in both arms and cancels when $c_q=0$; identity is paper--topic, not paper alone. Rational coefficients avoid cancellation error without imputing labels. This removes 23 shared unavailable terms for E--A and 30 for E--F. Joint and marginal-arm ranges are both reported. Analytic checks cover identical lists, shared and arm-specific unknowns, and unequal weights. The fixed reading-budget denominator is preserved while uncertainty in separate arm means is distinguished from uncertainty in their paired difference.

The missingness ranges enumerate possible contrasts over unavailable labels (Tab.~\ref{tab:qwen-reading-bounds-v32}). Conditional bootstrap intervals use 2,000 shared paper-ID cluster draws (seed 1111) only when the required effective outcomes are observed. Cross-model agreement concerns common relevance categories because the reference-cue rubrics differ.

\input{experiments/reading_choice_qwen_bounds_table_v32}

%% file: experiments/reading_choice_qwen_bounds_table_v32.tex
\begin{table}[tbp]\centering\small
\caption{\textbf{Paired reading-list differences with shared unavailable items cancelled.} Qwen3.8-27B, top-50 in each of sixteen directions. Bounds give arithmetic ranges over unavailable labels in the fixed item set; interval endpoints round outwards; the identified last-row differences round to three decimals. The two primary and four adjacent secondary comparisons are all retained.}\label{tab:qwen-reading-bounds-v32}
\setlength{\tabcolsep}{5pt}\renewcommand{\arraystretch}{1.16}
\begin{tabularx}{\linewidth}{Yrrrr}\toprule
\rowcolor{tablehead}\textcolor{white}{\textbf{Contrast}}&\textcolor{white}{\makecell{Relevance $\Delta$\\Lower}}&\textcolor{white}{\makecell{Relevance $\Delta$\\Upper}}&\textcolor{white}{\makecell{Reference $\Delta$\\Lower}}&\textcolor{white}{\makecell{Reference $\Delta$\\Upper}}\\\midrule
\rowcolor{tablepale}
Full context $-$ fixed-penalty baseline&$-0.079$&$0.068$&$-0.069$&$0.078$\\
Full context $-$ fixed reference&$-0.043$&$0.048$&$-0.043$&$0.048$\\
\rowcolor{tablepale}
Add author catalogue&$-0.030$&$0.030$&$-0.024$&$0.037$\\
Add topic-fit authors&$-0.022$&$0.005$&$-0.018$&$0.009$\\
\rowcolor{tablepale}
Add institution context&$-0.047$&$0.044$&$-0.050$&$0.040$\\
Add network authors&$0.004$&$0.004$&$0.008$&$0.008$\\
\bottomrule\end{tabularx}
\par\smallskip\raggedright Each arm has 800 direction--paper memberships. Shared unavailable items cancel at equal weights: 23 for full context versus the fixed-penalty baseline and 30 for full context versus the fixed author-reference arm. Original wider marginal bounds remain in the numerical supplement.
\end{table}

%% file: experiments/reading_choice_joint_arm_table_v32.tex
\begin{table}[tbp]\centering\small
\caption{\textbf{Six-arm readouts at the same top-50 budget.} Each arm has 800 memberships across sixteen directions. Jev: mean $\pm$ enclosing 95\% half-width; Qwen: range centre $\pm$ enclosing half-width for missing-label bounds. Bold marks the highest Jev means. Relevance uses 0--3; reference cues use 0--2 for Jev and 0--3 for Qwen.}\label{tab:reading-choice-joint-arms-v32}
\setlength{\tabcolsep}{5pt}\renewcommand{\arraystretch}{1.16}
\input{experiments/reading_choice_joint_arm_table_body_v32}
\end{table}

%% file: experiments/reading_choice_joint_arm_table_body_v32.tex
\begin{tabularx}{\linewidth}{Yrrrr}
\toprule
\rowcolor{tablehead}\textcolor{white}{\textbf{Arm}} & \textcolor{white}{\textbf{Relevance}} & \textcolor{white}{\textbf{Reference cue}} & \textcolor{white}{\makecell{Substantive\\(\%)}} & \textcolor{white}{\textbf{Available / $n$}}\\\midrule
\rowcolor{knowlight}\multicolumn{5}{l}{\textbf{Jev: mean $\pm$ 95\% half-width}}\\
Fixed-penalty baseline & \cival{\mathbf{2.655}}{0.038}{0.036} & \cival{1.881}{0.013}{0.012} & \cival{93.8}{1.8}{1.6} & 800 / 800\\
Author catalogue & \cival{2.651}{0.039}{0.036} & \cival{1.884}{0.012}{0.011} & \cival{\mathbf{93.9}}{1.8}{1.6} & 800 / 800\\
Topic-fit authors & \cival{2.646}{0.038}{0.038} & \cival{1.884}{0.012}{0.011} & \cival{93.6}{1.8}{1.6} & 800 / 800\\
Institution context & \cival{2.633}{0.042}{0.040} & \cival{1.886}{0.012}{0.011} & \cival{93.1}{1.9}{1.7} & 800 / 800\\
Full context & \cival{2.636}{0.040}{0.039} & \cival{\mathbf{1.887}}{0.011}{0.011} & \cival{93.2}{1.8}{1.6} & 800 / 800\\
Fixed author reference & \cival{2.636}{0.040}{0.040} & \cival{1.879}{0.014}{0.014} & \cival{92.6}{2.0}{1.7} & 800 / 800\\
\midrule
\rowcolor{knowlight}\multicolumn{5}{l}{\textbf{Qwen3.8-27B: range centre $\pm$ half-width}}\\
Fixed-penalty baseline & \pmval{2.341}{0.079} & \pmval{2.391}{0.079} & \pmval{86.8}{2.7} & 758 / 800\\
Author catalogue & \pmval{2.342}{0.072} & \pmval{2.398}{0.072} & \pmval{86.5}{2.4} & 762 / 800\\
Topic-fit authors & \pmval{2.333}{0.077} & \pmval{2.393}{0.077} & \pmval{86.2}{2.6} & 759 / 800\\
Institution context & \pmval{2.332}{0.081} & \pmval{2.388}{0.081} & \pmval{85.8}{2.7} & 757 / 800\\
Full context & \pmval{2.336}{0.081} & \pmval{2.396}{0.081} & \pmval{86.1}{2.8} & 757 / 800\\
Fixed author reference & \pmval{2.333}{0.077} & \pmval{2.393}{0.077} & \pmval{86.2}{2.6} & 759 / 800\\
\bottomrule
\end{tabularx}

%% file: sections/semantic_portrait_appendix_v30.tex
% Additive appendix fragment; the manuscript owner chooses its insertion point.
\subsection{Semantic and lexical views of the cached corpus}
\label{app:semantic-portrait-v30}
\begin{figure}[tbp]
\centering
\includegraphics[width=\linewidth]{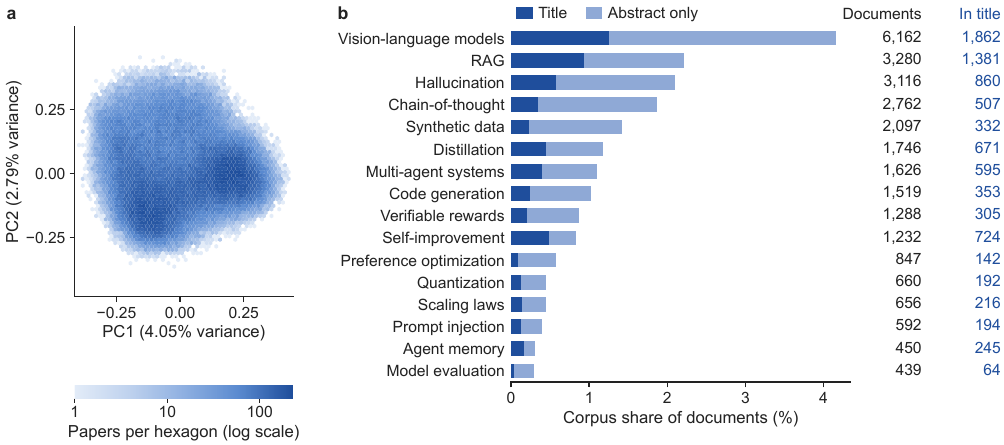}
\caption{\textbf{Complementary views of 148,117 acquired papers.} Both panels use the October 2025--October 2026 cohort after the fixed administrative and exact-duplicate exclusions. (a) Hexbin density (papers per hexagon, log scale) of normalized Qwen3-Embedding-8B representations in the first two components of principal component analysis (PCA), which together explain 6.8407\% of centered variance. (b) Document frequencies for sixteen fixed research-phrase groups as corpus shares: dark segments count title mentions, and light segments add documents matched only in cached abstracts. The columns give documents with a title or abstract mention and those with a title mention. Groups may overlap. The axes describe embedding variation, while the phrase counts provide literal content evidence.}
\label{fig:semantic-portrait-v30}
\end{figure}

Beyond individual reading decisions, a versioned library supports reproducible descriptions of the literature it has acquired. Fig.~\ref{fig:semantic-portrait-v30} summarizes the same eligible collection through frozen semantic representations and directly inspectable phrase mentions; both views reuse the cached records and representations.

%% file: sections/lexical_appendix_v31.tex
\subsection{Complete technical expressions and changes in their use}
We extract complete technical expressions from all 148,117 eligible abstracts, preserving internal connectors and long-form--abbreviation definitions. Candidates have one to six components, extended to twelve for explicit definitions; source identifiers and offsets accompany distinct-abstract counts.

\begin{figure}[tbp]
\centering
\includegraphics[width=\linewidth]{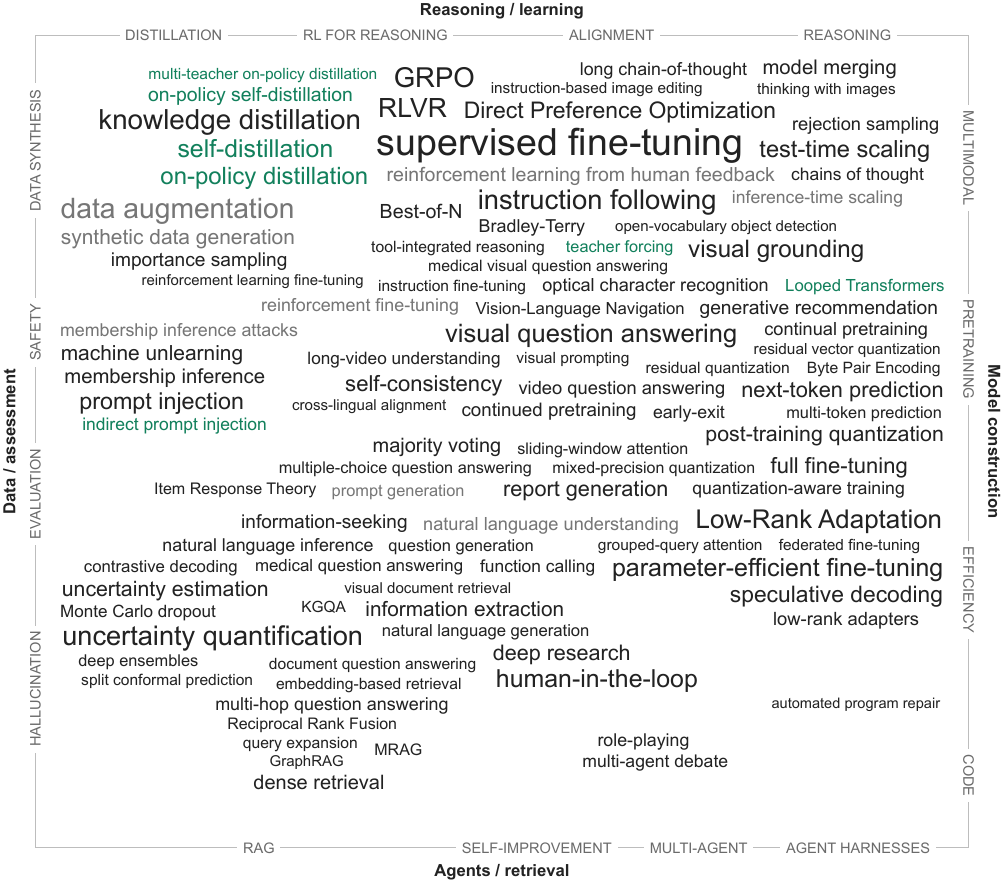}
\caption{\textbf{Methods, mechanisms and tasks in the acquired LLM literature, placed by the reading directions that use them.} The 100 leading expressions by original detail-score rank. Each side of the frame is one direction family (bold) and the grey capitals are the sixteen directions; position is approximate (see text). Type size grows with the square root of distinct abstracts (35 to 2,626; the deduplicated union for RLVR). Green marks expressions whose standardized prevalence at least doubled ($p_{t,L}\geq2p_{t,E}$, defined under Changes in use), grey those that at least halved. RLVR, GRPO, MRAG and KGQA are spelled out in Tab.~\ref{tab:exploratory-phrase-counts-v31} and GraphRAG in the text.}\label{fig:exploratory-phrase-cloud-v31}
\end{figure}

The detail score adapts the C-value nesting discount~\citep{frantzi2000cvalue}. Hyphens, spaces and underscores separate components; \emph{Best-of-N} has three components, two content-bearing. Let $d_t$ be distinct-abstract support, $\ell_t=\max(1,n_t^{\mathrm{content}})$ the count after stopword and reporting-word filters, and $\PaperCopyMath{D835DC510304005F007B2283D835DC61007D}{\bar{d}_{\supset t}}$ mean support over retained longer expressions containing the phrase (zero if absent):
\[
C_t=\log_2(1+\ell_t)\max(0,d_t-\PaperCopyMath{D835DC510304005F007B2283D835DC61007D}{\bar{d}_{\supset t}}),\qquad
D_t=\log(1+C_t)(1+J_t).
\]
Here $J_t$ compares the phrase's primary-category distribution, smoothed with fifty background documents, with the corpus distribution by normalized Jensen--Shannon divergence. Display candidates require 25 documents, three defining documents and half their support in the sixteen-direction LLM frame. Role filtering keeps 144 of 197 representatives. Grouping variants of reinforcement learning with verifiable rewards (RLVR), graph-based retrieval-augmented generation (GraphRAG), membership-inference attacks and majority voting gives 139 groups. Fig.~\ref{fig:exploratory-phrase-cloud-v31} maps the leading 100 by highest original group score; Tab.~\ref{tab:exploratory-phrase-counts-v31} reports twelve detailed forms. RLVR's exact union covers 873 abstracts and 714 defining documents. Scores and exclusions are available.

\paragraph{Reading the map.}
In Fig.~\ref{fig:exploratory-phrase-cloud-v31} the families and directions of Tab.~\ref{tab:llm-scope} are ordered so that neighbours share vocabulary. An expression starts at the mean of the direction positions, weighted by the square of its prevalence within each direction's abstracts and stretched to fill the frame; its label then takes the nearest free place, large labels first (median displacement 26~pt, 90th percentile 88~pt, on a 482 by 424~pt canvas). All 51 expressions whose largest direction share is at least 0.4 end nearest to that direction or to one beside it; the layout was chosen among drafts by eye, and type size and colour do not depend on it. The map is descriptive: directions are assigned by keyword matching, and 19 of the 100 expressions equal, contain or are contained in a search term of their largest direction, so they lie near it by construction.

\paragraph{Changes in use.}
The early and late windows are October--December 2025 (28,009 abstracts) and August--October 2026 (41,019). Within primary-category groups, prevalence $(d_{tc,w}+0.5)/(N_{c,w}+1)$ is averaged using fixed full-corpus weights; categories below 1,000 documents are pooled. Writing these prevalences as $p_{t,E}$ and $p_{t,L}$, we order positive changes by
\[
U_t=\max\!\left(0,\log_2\frac{p_{t,L}}{p_{t,E}}\right)
\log(1+d_{t,L})\frac{d_{t,L}}{d_{t,L}+d_0},
\]
with support constant $d_0=20$ and at least ten late-period documents. Early/late counts are 5/380 for on-policy distillation and 0/152 for on-policy self-distillation, comparing first-submission cohorts of the cached versions.

\begin{table}[tbp]
\input{sections/technical_terms_table_v37}
\end{table}

%% file: sections/technical_terms_table_v37.tex
\begingroup
\centering\small
\captionof{table}{\textbf{Detailed method and task terminology.} Twelve leading display groups with at least four word components. $n$ counts distinct abstracts; $n_{\mathrm{def}}$ counts explicit definitions. RLVR uses deduplicated union counts; other values retain the representative expression's support.}
\label{tab:exploratory-phrase-counts-v31}
\setlength{\tabcolsep}{6pt}\renewcommand{\arraystretch}{1.10}
\begin{tabularx}{\linewidth}{Xrr}\toprule
\rowcolor{tablehead}\textcolor{white}{\textbf{Method or task expression}} & \textcolor{white}{\textbf{$n$}} & \textcolor{white}{\textbf{$n_{\mathrm{def}}$}}\\\midrule
\rowcolor{tablepale}
Reinforcement Learning with Verifiable Rewards (RLVR) & 873 & 714 \\
Group Relative Policy Optimization (GRPO) & 990 & 724 \\
\rowcolor{tablepale}
parameter-efficient fine-tuning (PEFT) & 700 & 313 \\
human-in-the-loop (HITL) & 692 & 57 \\
\rowcolor{tablepale}
reinforcement learning from human feedback (RLHF) & 245 & 194 \\
multi-hop question answering (MHQA) & 158 & 6 \\
\rowcolor{tablepale}
on-policy self-distillation (OPSD) & 239 & 119 \\
long chain-of-thought (Long-CoT) & 151 & 10 \\
\rowcolor{tablepale}
multiple-choice question answering (MCQA) & 85 & 33 \\
multimodal retrieval-augmented generation (MRAG) & 76 & 29 \\
\rowcolor{tablepale}
medical visual question answering (Med-VQA) & 70 & 22 \\
Knowledge Graph Question Answering (KGQA) & 59 & 38 \\
\bottomrule\end{tabularx}
\par\endgroup